# The Framework
# of a
# Design Process Language

by

Arnulf Hagen

# Abstract

There are two main purposes with the thesis; to develop a view of design in a concept formation framework and to develop a language that can be used to describe both the object of the design and the process of designing. The main benefit in using a language along the lines proposed in this thesis is that it would enable a far better use of automated tools in the design process.

The unknown object at the outset of the design work may be seen as an unknown concept that the designer is to define. Throughout the process, she gradually develops a description of this object by relating it to known concepts, or things and characteristics. The search towards a complete definition stops when the designer is satisfied that the design specification is complete enough to satisfy the requirements from it once built. The complete design specification is then a collection of propositions that all contribute towards defining the design object - a collection of sentences describing relationships between the object and known concepts. Also, the design process itself may be described by relating known concepts - by organizing known abilities into particular patterns of activation, or mobilization.

In view of the demands posed to a language to use in this concept formation process, the framework of a Design Process Language (DPL) is developed. The basis for the language are linguistic categories that act as classes of relations used to combine concepts. One of the main contributions from the DPL is that it contains relations used for describing process and object within the same general system, with some relations being process specific, others being object specific, and with the bulk being used both for process and object description. Another important outcome is the distinction of modal relations, or relations describing futurity, possibility, willingness, hypothetical events, and the like. The design process almost always includes aspects such as these, and it is thus necessary for a language facilitating design process description to support such relationships to be constructed.

The DPL is argued to be a foundation whereupon to build a language that can be used for enabling computers to be more useful - act more intelligently - in the design process.

# Summary

The purpose with the thesis is two-fold; to investigate <u>what characterizes the knowledge</u> <u>used and produced in a design process</u> and to develop the <u>framework of a language that can be used to express this knowledge.</u> The discussion of these two issues cannot be separated - without knowing the language with which to express the knowledge it can hardly be expressed what the knowledge is, and without knowing what knowledge to describe the language can hardly be developed.

I start by discussing literature in Chapter 2 with two main intentions. One is to disclose whether design may be viewed as a general intelligent behavior, especially the kind of intelligent behavior that humans display in problem solving situations. Upon comparing literature in design with the theory of intelligence as proposed by Sternberg, I find that there are several common points. All major processes that Sternberg, as a proponent of the Information Processing Theory, argues to be important in intelligent creatures are in some way traceable into a wide specter of works in the field of design research. This sets the stage for discussing design in the context of human behavior.

The second purpose of Chapter 2, and of most importance to the subsequent parts of the thesis, is to show that when researchers discuss design they really discuss how things relate. These relationships may concern how activities are ordered, how decision rules map between a condition and an action, how an object search space is organized, how objects in that search space are located, or how decisions are ordered and what is decided upon. However, the fundamental question is still how things should relate to each other to form a coherent whole so as to express ideas about how design may be described. Theories are built by discovering regularities, and these regularities are described through establishing relationships among known things - unknown things are described by relating them to known ones. I take the idea of the relationships further to see whether it can bring some understanding of what design is about or, at least, how a design process and a design object can be described.

In Chapter 3, a brief investigation of existing representational systems

- representational languages - is undertaken to see what they can offer in terms of representing knowledge about a design object and a design process. They all, in some form or another, use relations between concepts in order to represent knowledge. However, the representational languages are too expressively weak for the purpose of sufficiently representing knowledge that is relevant when both design process and design objects are to be described. This is particularly so when the ambition is to use one process or object model for several different purposes, some of which may not be known at the time when the model is first developed. I argue that to make such a model applicable also to new applications - new kinds of processes or extensions to the objects (new characteristics) - the object and the process must be expressible in the same language.

In Chapter 4 I start the main development in the thesis through viewing the design process as one aimed at forming concepts. To perform concept formation, a designer also needs to form concepts of plans and sequences of activities to use in producing new knowledge. The design process is described as a set of relationships among abilities that can be represented as concepts in the same language as the elements that describe the design object. Thus, like "Activity A follows activity B" is typically a process description and "Deck A is above deck B" is typically an object description, they are both relationships among concepts and both representable in the same kind of language. The relationships are termed <u>sentences</u> in the <u>Design Process Language</u>. In other words, it is argued that a designer tries to form new concepts by relating them to other concepts in such a way that the relevant information for assessing the goodness of the design object is at least in its basis related to knowledge of the real world (or at least to 'generally known' concepts).

I further develop an alternative view of design as formulating, proving or rejecting hypotheses. The designer may be viewed as trying to prove hypotheses about how the design object will behave in the future, this by designing her design object such that it has the best chances of displaying such behavior as expected or required. On the other hand, nothing that the designer is doing is necessarily 'objectively' true in the future - there is another stage in which the description is to be <u>materialized.</u> Seen from that angle, the designer constructs hypotheses for operators performing in the next stage (the "materialization" stage) to prove by materializing a product that has the best chances of satisfying the design object specification, or proving the hypotheses, that the designer has expressed. The designer may be seen as a mediator between a 'customer', who expresses expectations, and a 'constructor' who materializes them according to the designer's specification.

The commonly referenced division between structure and function of the design object is argued to be somewhat elusive in that it is difficult to see where the one ends and the other starts. It is shown that the specification of a design object is a mixture of expressions of structural and functional nature. The specification should be S-F complete at the end of the process, meaning that the aggregate of functional expectations and structural descriptions presented by the designer should be such that the specification can prove the expectations that the designer set out to meet - i.e., such that it can prove the initial hypothesis. In other words, all parts of the design object are either described structurally, in terms of dimensions and other 'tangible' information, or functionally, in terms of behaviors expected from specific parts. Like when a designer in an outline specification lay out the dimensions of length, beam, and draught of a vessel (structural), and describe the engine in terms of expected performance (functional).

In Chapter 5 I start to develop the framework of the Design Process Language (DPL). Requirements to the DPL are presented and the language is developed from this angle. The relations in the sentences are discussed as linguistic categories, notably as various verbal, prepositional, and conjunctive relations, as well as adjective and adverbial modifiers. Each of the categories are argued to play a particular role in describing different phenomena experienced in, and that are characteristics of, the design process.

I further argue that several of the concepts that are used to describe a product, such as *class, system, attribute,* and others, are useful, but not crucial, as seen in a concept-formation framework. Such concepts are like any other concepts - defined by and serving to define other concepts. A class is thus more a concept being related to concepts constituting class membership criteria, than it is an aggregate of concepts that are members of the class. To treat a 'class' as an individual concept there must be ways to reference the concept properly - I claim that this is closely related to the use of language.

For instance, in DPL the sentence "(View) The ship as a structure" is not a message classifying 'ship' as a structure; it is rather a relationship asking that attributes should be extracted that are relevant for viewing the ship as a 'structure' concept. This may imply that relationships between the 'ship' and its 'paint' and 'rudder' are considered irrelevant in the context, whereas the relationships between the 'ship' and the 'plates', 'girders', 'beams ', and 'columns' of its 'hull ' are focused. In other words, a 'class membership ' can be viewed to filter knowledge from a larger, underlying description of the world. In plain terms, the vessel is conceptualized as a structure - not classified as a structure.

In conclusion, the DPL is a language where various categories of relations are applied to construct sentences that control both the design process and the design object. The semantics of the relations that tie one concept to another are what make definition of those concepts possible. This opens for a flexible and dynamic object and process description, where gradual definition and re-definition of both is possible. It also opens for a more dynamic approach to design, where the semantics of concepts may be more related to context than otherwise - the context will at any time determine what kind of meaning a concept should be given. This is useful both in relation to design research and to make possible an automated design environment.

# Table of contents

# Chapter 4: Resolution through relations ...........   68

# Chapter 5: Developing relations ...................  109

# Chapter 6: Drawing lines . . . . . . . . . . . . . . . . . . . . . . . . . . . . .  162

# Chapter 7: Concluding remarks

# Chapter 1: Introduction

## Background

From the  time design research was started in earnest In the early 1960's, a tremendous amount of effort has been Invested Into describing, defining, and theorizing about design. The descriptions of design as performed by actors ranging from novice to experienced designers, performing innovative or routine design on all from bank tellers (Goel and Pirolli, 1990) to offshore installations (WIkstr0m, 1989), are numerous. A new definition of design occurs more frequently than might be expected In a thirty year old discipline, and  theories regarding design apparently display little convergence as to what are 'scientifically sound· theories.

A striking commonality between several of these descriptions, definitions, and theories, is a lack of attention to the language with which they are communicated. Either researchers (I) tend to use fundamental concepts as given, presuming that these concepts are universally understood in the same way as they themselves do, (ii) define these established concepts in a particular way, making it fit more readily into the context, or (iii) establish new terms, possibly in order to avoid any serious conflict with what other researchers associate with them. However, there Is little <u>explicit</u> consensus on language.

Throughout the history of design research, a  recurring theme has been whether design and design research are in fact sciences. In this paper, there will be no argument as to  whether they are or not, but it is useful to note what would be demanded if design actually were to be  viewed as a scientific discipline. Indeed, it is possible to  "... regard the process by which theories are derived in science as similar to  the process by which we produce a design." (Coyne et. al., page 9) Ziman says of scientific theories that "(they) appear as ordering principles that explain general classes of observational and experimental facts, Including the taxonomies, 'laws', causal chains and other empirical regularities that are discovered about such facts." (1984, page 28) He continues to say that "... every scientific specialty has an  established <u>repertoire</u> <u>of useful concepts</u> <u>and formalisms</u> out of which new theories can  normally be constructed." (page 30, underscored here.)

To follow up on this view, it appears that it is crucial that there exists a consistent language, including vocabulary elements, a classification scheme (or a taxonomy), and rules for combining the vocabulary elements, in order to treat design with scientific rigor. Of course, it may certainly be debated whether this line of thinking at all is appropriate - since the scientific approach demands that there exist 'regularities', one would have to presuppose that regularities indeed do exist in the way design is accomplished. There is, however, quite a number of indications that there are <u>some</u> common elements in the way design is performed, although it still seems hard to merge all the findings into coherent and substantial theories. (See Chapter 2)

---

"We live in an age of growth, in which every day more and more things come into our lives; and things, and all their parts, need names. So more and more words come in with them - new words, or new ways of exploiting, embellishing and combining the old ones; and in this way the balance is maintained. There is no sign that our onomastic (vocabulary) resources are drying up; indeed we are likely to run out of natural resources before we run out of things we make out of them. But there are signs that the things are becoming more resistant to being named. There is no natural way of referring to the various plastic objects that lie around the house, or the toys we give our children, or the furnirure we now have to assemble for ourselves. They no longer fit into taxonomies. We live in modules, sit on units and entertain ourselves with systems.

Behind these nameless objects is a technology and a science that produce them; and there, less visible to us, Is another realm of things that have to be named. Many of them are abstract things, the categories and concepts of a theory; and some of these also prove recalcitrant to ordinary onomastlc processes - they only come to be 'named' by some mathematical formula, like 'a function of the coordInates x and y', 'the integer over $psi_1$ and $psi_2$, and so on. But somehow they have to be enmeshed in the language; otherwise they are not brought under control." (Halliday, 1988, page 27)

**Box** I.i

---

There is another development in design that makes a language increasingly more important; The increasing reliance on the computer as an aid to design, and the steady and massive efforts to enable the computer to solve tasks that hitherto were performed by humans. The more the computer is to solve 'human' tasks, the more it must be made to "understand · the information it is to process". Consequently, the more expressive must the language be that is used to represent this information. The catch is that the computer, as we know it, needs more formality and less ambiguity in its language than humans do. It remains to be developed a language that is expressive, that is formal, and that minimizes the risk of misinterpretation.

This work intends to establish the framework of a language with which design process and the design object can be described, defined, or theorized about. I claim that such a language is at most absolutely necessary if scientific rigor is aimed at in the work, and at least useful if design is treated without any concern attributed into the issue of whether design and design research are scientific disciplines or not. I claim further that a language developed along the lines in this thesis will enable a more efficient introduction of computer aids into the design process.

## Purpose

The underlying assumption for this thesis is that descriptions, definitions, and theories about design are most effective when a coherent and rigorous language exists with which it is communicated. That is, the content of the expressions is most effectively expressed when the terms and concepts used for verbalization, and the relationships between these terms and concepts, are strictly described or defined. This paper has a twofold purpose; it intends to propose some views of design at the same time as the framework of a language with which it is possible to describe and define it is developed.

With *language* is here meant a set of clearly defined fundamental concepts and a means for classifying these concepts.[a] This corresponds partly to the structure of natural language, where there is a vocabulary and a classification of the elements in the vocabulary. In addition, any language has rules for combining vocabulary elements into sentences - this aspect will to some extent be discussed in the course of the paper, notably in Chapter 5, but will mostly be left undone. Arguably, such rules, or regularities, is the crucial role for design research to discover - the language in development here introduces some of the elements about which regularities should be discovered.

---

[a]  Later. the term *concept* is used differently from what is done here. At this point,
*concept* refers to what is, in Chapter 4 and Chapter 5, called *relations* in the language.

Three aspects have then become crucial in developing a design language: (i) A set of concepts (or vocabulary elements); (ii) a classification scheme, henceforth called taxonomy; and (iii) rules, henceforth called regularities, for combining the concepts when communicating findings regarding design.

The concrete demands on this language is that it is <u>useful</u> for describing the processes (or tasks) undertaken by a human designer when designing, as well as the inputs, the intermediate results, and the outputs of this design process. As will be argued at a later point, the design process is here viewed as primarily an information production (or manipulation) process. This has the consequence that the resulting language is focused on describing properties of such information processes, as well as in describing the knowledge that these processes manipulate - that is, knowledge used by and generated from them. To the extent that rules are established here regarding how to combine vocabulary elements, these rules will focus on the relationship between various broad 'categories' of knowledge and not the particular instances as would be demanded if the design process indeed were to be described in full.

This leads to another major outcome of this research. In the development of any language one would necessary need to know what would be expressed with the language. So, one would need to discover which phenomena of importance that exist in the particular domain. Thus, the *process* of developing the language necessarily will include discussions of these phenomena, and the result becomes two-fold: A language, and a discussion of the phenomena described by the vocabulary elements in that language. The phenomena of particular interest in this work are the phenomena related to the information production in design.

One particular benefit from this approach is that it potentially provides a means for comparing more readily findings from different researchers with different viewpoints. This would be accomplished since characteristics of the engineering design process are described with a basis in clearly defined and described language elements. In other words, the terms used, and the phenomena they denote, are distinguished and to some extent classified.

---

[b]    With taxonomy is meant laws and principles of classification, here classification of
knowledge.

The resulting language is claimed to be neither the only appropriate nor necessarily a complete one for describing the design process. Likewise, it deals with one particular aspect of design; the (structurable) processing of knowledge, and is as such most efficiently used where it is feasible to view design as to some extent structurable. In other words, where intra-personal communication, intuition and other such 'intangible' aspects are concerned, the language is not developed to the extent that such knowledge can be described, although the language may be used for describing at what places in the design process where such phenomena 'intervene·.

Concluding, there are two particular areas where the work may promote design research: (i) In developing the framework of a language to use in discussing the design process; and (ii) in discussing basic activities and phenomena (the terms) from which a design process description is constructed. Some additional benefits from the latter claim is that design process modelling becomes easier. The approach taken here is to develop these terms such that they are capable of expressing the knowledge used and produced in a design process.

# Delimiting design

The more there is subscription to a definition, the more general this definition tends to be, and the less useful it becomes for the purpose of discussing design at a detailed level. Too general a definition tends to include in the category design activities that would seem inappropriate to throw into it. The risk is that design becomes like the psychologist Spearman said (in 1927) in response to an ongoing discussion regarding the concept *intelligence;* "... a mere vocal sound, a word with so many meanings that finally it (has) none." (Quoted in Guilford, 1967, page 12)

This constitutes a fundamental problem in even starting research into design. At a minimum, the area of concern must be indicated. Of course, researchers could subscribe to one definition and categorize accordingly. Using one definition as an example, from Rawson in 1979, illustrates the problem with this. He proposed that "design is a creative, iterative process serving a bounded objective." (Rawson, 1979) It may be easily agreed that design may

<u>be</u> both creative, Iterative, and bounded, but is it as easily agreed that it <u>always will be?</u>

The easy way out is to take the view that if a process is creative, iterative and serves a bounded objective it IS design. (Of course, the terms involved in the definition must be defined first.) In this case, the work of an artist is certainly creative, probably iterative, and will in some form serve a bounded objective, so <u>art work is design</u> according to the definition. Finding one's way through a labyrinth will probably also be design, as will baking bread without a recipe. To paraphrase Spearman; the word 'design' may get so many meanings that finally it gets none.

A more fruitful approach may be to say that if a process is creative, Iterative, and bounded, it <u>may be</u> a design process. Researchers may then want to delimit further by saying, for instance, that it must produce something tangible (this still doesn't exclude any of the examples above). They may say that the product should serve a function (as art work, finding one's way, and the bread all do provided 'function' is defined right). The problem now is that there is an infinite number of additions to the original definition that might be made, based upon what <u>feels like should be called design</u> - i.e., the definitions will be subjective. In the definition process, critics could come out with something like: "OK, so you subscribe to the view that design is creative. What about the shipyard you visited last month? Did they always do something creative in designing a ship? Or maybe that wasn't designing - maybe It was X? And, anyway, how do you define 'creative'?" A <u>creative</u> critic could generate more reservations than there exist additions and modifications to the definition of design.

Thus, a proper starting point might be to state something like "Design is characterized by X. Find X!" Although the present work does not aspire towards describing the design process as much as to develop the language with which it can be described, there will be a development of a particular view of design characteristics. Therefore, it is Important to establish the <u>context</u> in which the language is to be used. In other words, the first task Is to mark off a segment of the world where the findings will have validity - i.e., to state the <u>fundamental premises.</u>

The findings here deal with a particular situation, which is termed a case of a design process, and intends to identify some fundamental characteristics of such a process and to use these as a basis upon which a language can be formed, The premises presented here are intended to identify - not define - this situation, and the term *design* will be used as a reference to the design process situation. A description of (characteristics of) the design process will implicitly be developed throughout this paper.

One premise in this thesis is that design is considered as including *the production of knowledge,* That is, the designer accesses (if willing) more knowledge at the end of the process than at the start. Implied here is of course that *production implies process.* Further, the knowledge produced is intended to be implemented as a real world object which is viewed here as (Hagen, 1990, page 12): ",.. a combination of structures, procedures, and instructions meant to achieve a predefined purpose."

Another premise - a central one as such - to state here is that *design is something that may be accomplished by humans.* This does not say that all humans are capable of design, anything about the quality of the result or the process, or even that design necessarily must be performed by a human. It is merely to indicate that there 'is something' with humans that make them (or at least some of them) capable of designing.

As for the process of developing the framework of the language, I adopt the view that "(the) principal role of any design process is to convert information that characterizes the needs and requirements for a product into knowledge about the product itself," (Mistree et al., 1988, page 44) With this fundamental premise, the language will be developed with a basis in the knowledge used and produced during the design process. The vocabulary elements in the language will then focus on describing (or denoting) elements of knowledge and will be defined in terms of the role this knowledge plays in the design process.

# Viewpoint

As indicated earlier, the viewpoint chosen here focusses on the important 'ingredients' 'knowledge' and 'humans', If humans possess qualities that make them able to use and produce the knowledge required in the engineering design process, it would seem useful to take a closer look at these qualities. Likewise, since knowledge is viewed as the crucial result of the engineering design process, it would be useful to look at the nature of this knowledge.

There is frequently a separation in the literature between information and knowledge. One may take the view that information is embedded in all that is perceivable, and that this information becomes knowledge when someone has performed interventions upon the information in ways that are explained later. Thus, knowledge cannot in this view be permanently stored, while information indeed can be stored. Strictly speaking, there exists then no permanent *knowledge base.* I will not, however, distinguish between information and knowledge in this thesis, since it would seem like a philosophical question that is far beyond the scope to discuss at what point information becomes knowledge.

Here will further be taken an angle that is useful in this investigation; the qualities that make humans able to design are also viewed as knowledge. Indeed, Guilford states: "It is possible that It would be fruitful to regard our personal plans of action, our strategies, and our tactics, all of which have to do with what we do and how we do it, as forms of behavioral information .., Our repertoire of plans ... would constitute systems of behavioral information. With this conception, skills, psychomotor as well as intellectual, can be brought into the realm of information." (Guilford, 1967, page 238) In other words, some knowledge (information) is utilized to generate other knowledge. The physical matter (the structures) and the energy that able this knowledge to exist or be produced is disregarded.

It is useful now to look towards, e.g., temporal reasoning, where a situation is often called a *state,* and where any change in the situation implies a shift from one state to another. I will view a state as defined by its knowledge content - within one state the knowledge content remains unchanged. In this view, a change of states occurs whenever additional knowledge of the world is gained. Strictly speaking, the <u>act of living</u> is thus in this interpretation a continual change of states. However, in the present discussion on design it is more useful to view a transformation from one state to the next as triggered by the <u>introduction of new knowledge</u> that is determined to be <u>relevant</u> in the design process. Here, <u>all new knowledge acquired and stored</u> will by definition be viewed as relevant. A new state is then reached after each intervention in the process, and knowing that a second has passed will not necessitate a shift from one state to the next, provided time (in short increments like seconds) is not termed a relevant variable. It would anyway be extremely difficult (and probably pointless) to develop a language of concepts that had the flexibility to describe the information processes occurring when the design process is distracted by a tweeting bird outside the window or interrupted by a trivial telephone call.

In some sense, then, design can be viewed as a state-transformation process, as formulated by Willem: "In designing, a transformation occurs. The design problem information is transformed in the mind of the designer into design concepts - potential means of solving the problem - some of which ultimately become design solutions." (Willem, 1988, page 224.)

It may now be appropriate to take look at what a design process actually 'is'. In general, there seems to be agreement that it may be seen as a particular kind of problem-solving activity, often referred to as the act of solving *ill-structured problems.* (Simon, 1973; Newell and Simon, 1972) Such problem-solving activities are again seen as quite demanding cognitive tasks, requiring the extensive display of intelligent behavior. (Guilford, 1967; Sternberg, 1985) Thus, there is a natural opening to the field of *cognitive science,* in particular cognitive psychology - an opening that will be exploited in the pursuit of a language of concepts. This point of view will be revisited and elaborated later on in the thesis, specifically in Chapter 2. Here it suffices to recognize that design is viewed to constitute a particular kind of problem-solving situation.

Above was defined that a shift in states occurs when knowledge that is termed <u>relevant</u> for the design task is introduced. This notion of relevancy (as a relative concept) could cause problems, in that the design engineer does receive inputs all the time - inputs that do not carry labels regarding the relevancy of such information to the situation. However, the same engineer will want to filter out what is not related to the task at hand. "The individual (in a problem (solving) situation) is in a situation where the more or less automatic processes in a way come to a halt. The person is forced to concentrate towards various elements to a far more extensive degree than otherwise (in non-problem (solving) situations)" (Raaheim, 1983, page 49, in Norwegian) The concentration on the design problem becomes particularly intense at the cost of attention to incidental phenomena, reinforcing this tendency, and it is only natural that some aspect of relevancy comes into play.

Having identified engineering design by humans as a cognitive activity, there remains the issue of actually being able to describe it - the key concern of this thesis. As stated earlier, any phenomenon can be described only if there exists a language with which it can be described sufficiently for the purpose. While such languages are, for the most parts, quite well established for the particular disciplines in which engineering design is performed, it Is hardly established at all for the task of describing the design process. Thus, the *design object* can often be described quite well whereas the *design process* is more difficult to describe.

Most of the terms used in such a description are borrowed from other fields, as 'iteration', 'recursion·', 'creativity', 'convergence', 'divergence', 'analysis', 'synthesis', 'abstraction', 'modelling', and so forth. Indeed, it is difficult to find a single term in the literature that is developed uniquely for the sake of describing the design process. This would seem to indicate one of two things; either there are in fact no unique characteristics (for which one needs to coin terms) of the design process, or design research is too immature to have developed (or agreed upon) a unique language with which such a process can be described.

## Some motivations

It has repeatedly been proposed above that knowledge production is the ·essential contribution the design engineer makes to the creation of the design object. It has also been argued that the language of concepts used within the field of studying the design process seems grossly unfit to be for a detailed description or prescription of it. At this point, it may be appropriate to elaborate somewhat more on why it is useful to develop and refine a language also for describing the engineering design process. In this, there will be two lines of arguing; The scientific point of view and the practical point of view.

One benefit is that an expressive, while rigorous, language for the design process would make it easier to translate between findings from research into different fields of design. Establishing basic concepts in the language makes possible a definition of composite or abstract terms in the context of a Design Process Language - the DPL.

The utility of the present work from a more practical point of view is related to the viewpoint as described above. By being based on the roles that knowledge play, particularly on the knowledge that makes a design engineer able to produce other knowledge, the vocabulary elements bring with them descriptions of fundamental processes and activities (or tasks) required for these roles to actually be fulfilled.

The computer is already heavily organized into components, each having unique roles in the information processing. The software used in computers is built up in the same way - various unique commands trigger the execution of certain tasks. An efficient use of the computer in the engineering design process would, of course, be one that delegated to the computer the tasks that the computer could take care of best, and left to the human design engineer the tasks she could do best. With the language of concepts, and the corresponding description of vocabulary elements, developed here, can be constructed a model of the delegation of tasks between the human engineer and the external tools used, such as the computer.

There are some additional motivations behind the DPL as goes the application of computers in design. One is that the exchange of information between the computer and the human user is important. As for now, there are generally two different languages, where either the computer translates its internal representation of information into an understandable language for the user, or the user adopts to the computer's way of communicating. The DPL is based on natural language (English) which, arguably, makes such communication easier and more intuitively ·correct'. Another motivation is that as computers are used in more and more fields, the need for representing 'historic' knowledge of the design process becomes increasingly more important in order to 'learn' from past successes or failures. Such history may be accounted for in a language based on a natural language. A third motivation lies in the steadily increasing expectations from 'product models', particularly model that are to contain information in such a way that it can be set to use in all kinds of processes in all stages of life of the modelled product. As the aspirations go far, so must the language used to meet the aspirations.

## Structure of the Thesis

Chapter 2 establishes the arena through a discussion of literature, both within and outside the design research arena. The chapter develops a particular view of design, seen from the perspective of theories in psychology regarding intelligent behavior. In the chapter is argued that the main invariant in different works is that they all talk about relationships - whether describing the design process or the design object.

Chapter 3 summarizes some existing means and theories of how to represent and manipulate information. In addition, the distinction between *content theory* and *process model* is introduced and elaborated. Process is seen as a *mobilization* of knowledge in the content model - *abilities* are enacted. As in Chapter 2, it is argued that all the methods treated rely heavily on the *relation,* and that there are commonalities in the kinds of relations used. It is also argued that none of these 'languages' are sufficient for representing all the knowledge required to represent in a design process.

Chapter 4 introduces the view of design as the process of forming concepts. The design process is likened to the process of communication, where the object of communication is unknown. Design is further viewed as the process of testing hypotheses and as the process of proving hypotheses. Design is also viewed as the process of building sentences within a language where the relation plays a crucial role - "the knowledge is in the relations."

Chapter 5 is a development of the framework of a language intended to meet the needs put to it in the design process. The terms and the classifications are selected from English, where the main groups are verb relations, preposition relations, conjunction relations, and adverbs and adjectives. Most effort is spent in developing the verb relations, since these are claimed to be most central in representing knowledge, particularly that which pertains to the design process.

Chapter 6 is a short discussion of the virtues of the Design Process Language and the explicit relationship between the discussion on design in Chapter 4 and the discussion on the DPL in Chapter 5. It also contains a brief illustration of how the DPL may be conceived used in practice.

Chapter 7 contains a critique and presents the conclusions to draw from my work. It ends with suggesting important areas for further work.

# Chapter 2: Literature -    setting the stage

Several models have been developed to aid in achieving understanding of what the design process should or does look like. A discussion of various models might be pursued along several 'dimensions', for instance descriptive or prescriptive models, conceptual or detail design, automated or creative design, and so forth. I have chosen to discuss literature on design research in terms of two categories; that concerned with discussing the  design process *per se* and  that concerned with explicitly associating the design process with the design object. In addition, the idea of design as a  problem- solving activity requiring intelligent behavior is explored in the context of cognitive psychology. The platform of  the discussion is the Information Processing Theory (IPT), as formulated  by Sternberg.  After this, four different approaches to design research are discussed - those of Goel and Pirolli, Mistree et. al., Yoshikawa, and Coyne et. al.. Attempts are made to both relate the works to the IPT and to my own view of design as primarily using relationships to  build relationships.

## Sternberg -    an intelligent approach?

It may seem odd  to start the  discussion into relevant literature in the field of psychology. However, as will be  shown later, most researchers into design by humans either emphasize or assume that it is a  cognitive; intellectual activity. What would be more appropriate, then, than starting the discussion with theory from cognitive psychology? It would seem impossible to discuss a language without discussing what to describe with it. I will therefore distinguish some regularities and  relationships which are common to the psychologists (at  least information processing theorists) view of problem-solving and design researchers' view of design. This section contains a discussion of the psychologist Sternberg's work so as to form a reference to which other work explicitly  treating  design  can be  compared.

Sternberg builds his theory around a so-called *component/al analysis* of human abilities, focussing on three classes of components: *"Meta-components* are higher-order executive processes used in planning, monitoring, and decision making in task performance. *Performance components* are processes used in the execution of a task. *Knowledge-acquisition components* are processes used in learning new information." (1985, page 99) Note the interpretation of components as being processes rather than detectable structures or particular abilities. A component is "an elementary information process that operates upon internal representations of objects or symbols." (ibid., page 97) Newell and Simon (1972) use the same interpretation, and are as careful as Sternberg in noting that 'elementary' is a relative term in this context - any component may be broken into more elementary ones in an appropriate language; " .., there are many ways in which adequate sets of elementary processes can be chosen (in conjunction with corresponding schemes of composition)." Thus, any classification is pragmatic in the sense that the classification is one that is appropriate in the context rather than correct.

The discussion here will focus upon the meta-components of human intelligence. Sternberg identified seven such components, which he also terms *executive processes:* (1) Decisions as to just what the problem is that needs to be solved; (2) selection of lower-order components; (3) selection of one or more representations or organizations for information; (4) selection of a strategy for combining lower-order components; (5) decisions regarding allocation of attentional resources; (6) solution monitoring; and (7) sensing external feedback. For curiosity, it is interesting to compare this division of problem-solving components to the framework for problem solving by Polya as illustrated in Box 2.i.

*Decisions as to just what the problem is that needs to be solved* is the first part of what is often termed *problem structuring,* A precondition for solving a problem is certainly knowledge about what this problem is. While the problem may be very easy to identify, it may also be initially quite intangible. As the original structure of the problem becomes insufficient, inconvenient, unachievable, or simply wrong, the problem must be restructured, suggesting that this component stay active at all stages in the problem-solving, or design, activity,

The importance of *problem structuring* is frequently stressed in design literature. One of the earliest design researchers, Asimow, said that "before an attempt is made to find possible solutions for the means of satisfying the need, the design problem should be identified and formulated." (1962, page 20) Of a newer date, Willem says that "as the input ... the design problem information is a key factor in design.... Each new bit of information allows (the designer) to evaluate and possibly eliminate some of these ... ideas." (Willem, page 224) This is a kind of view that applies problem structuring for "shooting out' potential solutions as more information on the problem is applied. In fact, Yoshikawa suggests the same position in his work discussed later in this chapter. While this stepwise application of problem information is a somewhat radical, or dominant, view of the role of the problem information, others conceive it more as a 'guide' to the designer. Shahin says that "the designer should have a general idea about the problem.... Definition of (it) is what the designer is actually guided by, while solving the problem." (Shahin, 202-203)

Understanding the problem
    What is the unknown? What are the data? What is the condition? (Component l)
    Is it possible to satisfy the condition? ... (Component 1)
    Draw a figure. Introduce suitable notation. (Component 3)
    Separate the various parts of the condition. (Components 1 and 3)
Devising a plan
    Have you seen (the problem) before? ... (Component l)
    Do you know a related problem? ... (Component l)
    Look at the unknown! And try to think of a familiar
        problem having the same or similar unknown ...(Component l)
    Could you restate the problem? Could you restate it
        still differently? Go back to definitions ... (Component l)
    Could you imagine a more accessible related problem? A
        more general problem? A more special problem? An analogous problem?
        Could you solve part of the problem? Keep only a part of the condition;
        drop the other part; how far is the unknown then determined, how can it
        vary? Could you derive something useful from the data? (Components 2
        and4)
Carrying out the plan (Component 4)
    ... check each step ... (Component 6)
Looking back
    Can you check the result? ... (Component 6)
    Can you derive the results differently?.Can you see it
        at a glance? (Components 4 and 6)
    Can you use the result, or the method, tor some other
        problem?· (from Baron, 1987)

As is seen, there are clear indications of the presence of at least five of Sternberg's seven components. Component five ("attentional resources·) and seven ("external feedback") are not referred here, but it may be expected that this is due to the particular situation of mathematical problem-solving, where time and the external environment normally are of inferior importanee in the problem domain.

**Box 2.i:**    A framework for mathematical problem-solving from Baron, 1987. (Originally presented by Polya.)

<u>Simon</u> has still another view of the design problem. "Efforts to solve the problem must be preceded by efforts to understand it." (1981, page 111) He continues to say that "... solving a problem simply means representing it so as t,omake the solution transparent. If the problem solving could indeed be organized along these terms, the issue of representation would indeed become central." (ibid., page 153) Simon makes this statement even stronger other places, implying that problem solving essentially is an act of transforming ill-defined problems into well-defined ones. "There is merit to the claim that much problem-solving effort is directed at structuring problems, and only a fraction of it at solving problems once they are structured." (Simon, 1973, page 187) <u>Wikstr0m,</u> through comparison between observations and theoretical information, hypothesized that "an unclear project (or problem) description causes increased, and unexpected, consumption of resources." (page 75, in Swedish)

*Selection of lower-order components* is the process of choosing the appropriate lower-order (non-executive) processes, or the interventions, that are expected to cause the attainment of a final state from an initial state.

The notion of different order of the processes taking place during design is referred to by several researchers. <u>Christians and Venselaar</u> distinguished between *strategic* knowledge, what may be called executive processes, and *situational* and *procedural* knowledge, which are components of lower order. <u>Coyne et. al</u>. (1991) talk of different languages, where *language of plans* is the language describing the way lower-order components are chosen (or how the executive processes are enacted) and the *language of action* describe the way the non-executive processes are applied. (A more extensive discussion of Coyne et. al. is pursued in chapter 4.) Indeed, <u>Simon</u> lets this component become central in one of his definitions of design (1981, page 129): "Anyone designs who devises courses of actions aimed at changing existing situations (initial state) into preferred ones (final state)." Selecting lower-order components is clearly a matter of "devising courses of actions."

*Selection of one or more representations or organizations for information* is an obvious process in design. Sternberg notes that "the choice of representation or organization can facilitate or impede the efficacy with which the component operates." (1985, page 100)

Goel and Pirolli showed in research work discussed later in this chapter that designers utilize both internal and external artificial symbol systems. (Goel and Pirolli, 1990) The notion of using models in relation to design span from simply modelling the design object (Kimbler, Watford, and Davis; Sage; Fitzhorn), modelling the task or problem domain (Simon), modelling the design process (Hubka; Mistree et. al.), to even 'modelling the designer· (Yoshikawa; Maccallum). Maccallum views "the design system as a model of an intelligent mechanism" and aims "... to replace (sic!) this human centered activity with computer centered activity." (1990, page 59)  All these activities imply using some sort of information representation whose efficiency will influence the usefulness of the model in the design process. In fact some researchers, as Fitzhorn and Coyne et. al., introduce language explicitly; "realizable shape is a language" (Fitzhorn, page 151) and "design descriptions are like sentences in natural language." (Coyne et. al., page 28) Others, such as Alexander (see *A Pattern Language),* have attempted to form complete languages of design in particular disciplines (Alexander focused on architecture) based on an identification of syntactic elements and to some extent the grammar with which to combine them.

Actually, the efficiency of the fundamental problem structuring component referred to earlier presupposes the choice of an efficient information representation system. Simon states that "Solving a problem simply means representing it so as to make the solution transparent. If the problem solving could indeed be organized along these terms, the issue of representation would indeed become central." (1981, page 153)

*Selection of a strategy for combining lower-order components* mirrors the obvious fact that not only the identity of relevant lower-order components is important, but also their sequencing in the process of implementation. The most visible "proof' of its importance is found in research into computer aided design, where much work is performed into constructing algorithms for an efficient sequencing of various processes (Simon; Newell and Simon)

.It is more than likely that the process of choosing the non-executive components will go hand-in-hand with the process of selecting a strategy for combining them. In any case, it seems clear that the strategy may not be possible to lay out *6 priori,* so that there will be a constant judgement as to what non-executive components are to be chosen, and thus how they are to be combined into a strategy. This is equivalent to arranging the application of various tools in a (relevant) toolkit to ensure, or maximize the probability, that the problem is solved.

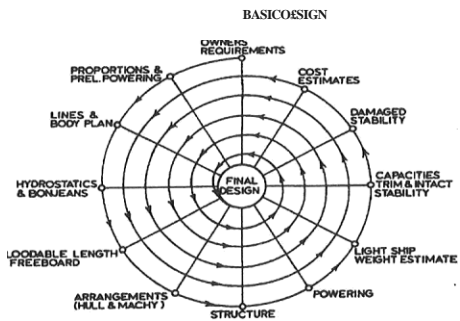

Figure2.l:      The "design spiral" taken from Dillon. 1969.

A frequently cited kind of model of the design process in the marine field is the *design spiral,* first presented by Evans in 1959. The spiral is illustrated in Figure 2.1, this particular one adopted from Dillon (page 15, 1969). The model, from the domain of ship design, identifies main activities and reflects the idea of the process being sequential, recursive, iterative, and converging. What is most interesting at this point, though, is that the sequencing of the activities is given in the model. While the choice of language, particular tools, and definition of problem are left unattended to by the model and rendered to the designer, the model commits the designer to choose a particular strategy for applying the tools. The relatively high frequency of this kind of prescriptive models (Evans; Gilfillan; Rawson; Ehrenspiel) could seem to reflect the importance attributed to this component by design researchers. Figure 2.2 is another illustration, borrowed from Asimow, underlining the persistence of the view that arranging the components (processes) is key in determining the quality of the design process.

*Decisions regarding allocation of attentional resources* refers to the resources vested into the problem-solving process itself rather than the obvious allocation of resources in the design object. It is a component because the solution to any task, and design in particular, is obviously subject to bounds in the temporal, monetary, or other resources to be spent on the task.

While there is little research into how designers actually use this component, some researchers identify the issue as important. Erichsen says that "design may be considered the task of coordinating the spending or resources to reach a specified goal." (1989, page 179, original underscore) Simon elaborates; "the design process itself involves management of the resources of the designer, so that his efforts will not be dissipated unnecessarily in following lines.of inquiry that prove fruitless." (1988, page 144) The designer thus needs to prioritize among tasks - to focus efforts into areas where the pay- back from these efforts is greatest. Westerberg et. al. says that "most likely, designers make many decisions based on intuition because of time constraints .. ,," (1990, page 114) implying this dilemma of the designer of "allocating the attentional resources." Some researchers have tried to formalize the issue of allocating temporal resources. Wikstr0m discusses this issue through comparing theory and empirical data. He says that "as the size of the design project increases, the percentage of time allocated to the first phases (viability and feasibility) is reduced." (page 51, in Swedish)

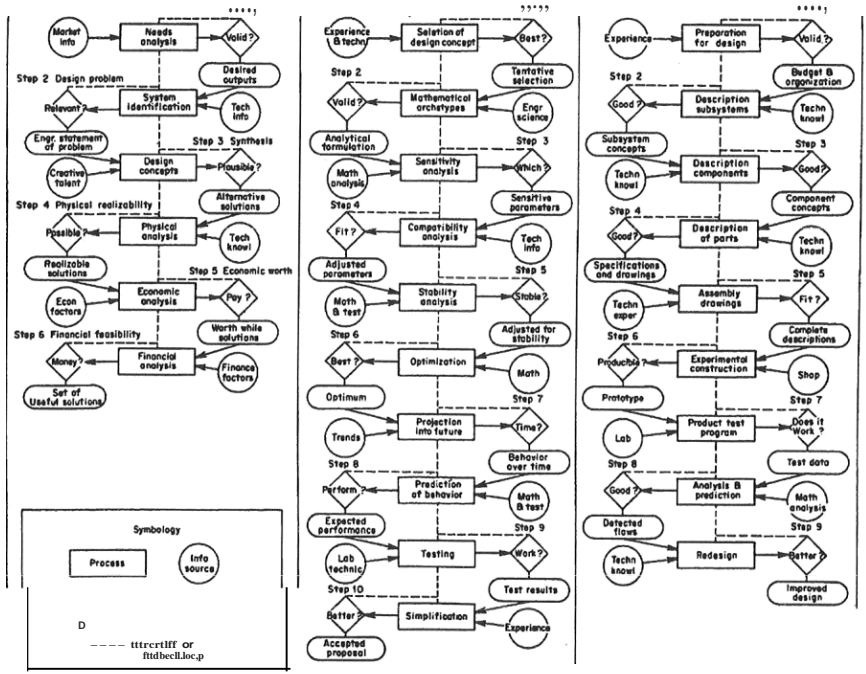

Figure 2.2:        Asimow, 1962: Morphology of design.

*Solution monitoring* is the component that helps problem solvers "keep track of what they have already done, what they are currently doing, and what they still need to do." (Sternberg, 1985, page 105) The importance of this component in design is obvious when the designer may be seen to have an outer environment (like a customer) that demands results at particular points in time. Large project scheduling and management is exactly aimed to take care of such solution monitoring. Its importance is less obvious when the designer operates with no explicit exterior demands. Goel and Pirolli showed that designers use "personalized stopping rules" to determine when the problem is solved to a satisfactory degree. These are clearly rules that are involved in the solution monitoring process. In addition, any kind of performance modelling of the designed object would seem to imply that the results are used for "monitoring the solution." Simon (1981) introduced the notion of the *satisficing* designer rather than *optimizing* designer. The solution monitoring process is thus one that terminates when the designer has optimized her satisfaction -  in other words, when she is content.

Asimow's  model in Figure 2.2 further  illustrates the importance of the  solution monitoring process. Most of the diamond-shaped icons in the figure signify tests of the solution ("Best?,"  "Valid?," "Producible?," "Does it work?," and  so forth) as  the  design process evolves. The 'design history' aspect is dealt with by Smithers et. al. who say that "the record of the concurrent and serial searches, leaps of intuition, analyses, syntheses, simulations, prototypes,  decisions, and judgements, etc .. ,, represents important knowledge........It represents a record of  what parts of the design space  were explored, how, and  why decisions were made." (page 82) Such historical information is important in design both to make re-design more efficient and to make final design documentation (or specification) more complete.

Other researchers highlight the process of retaining knowledge of past occurrences as crucial in learning to approach new design problems more efficiently, Maccallum says that "experience of past problems and their solutions ... forms one of the most powerful and important sources of solutions to design problems." (page 62) Gero and Rosenman say primarily the same: "(Designers) accumulate a library of vocabulary elements within their domain ,.. each such element having with it associations to other elements and their relevance to other situations." (1990, page 66)

*Sensing external feedback* is the component through which an individual adjusts or learns from observing how the results of the problem-solving process are received in the outer environment. M'Pherson formalizes this as an integral part of systems design:

> "The Monitoring System - the information retrieval and feedback system (-) ...is composed of two parts:
> (i)  that specifically designed and inserted to provide information feedback to the management, designers and operational controllers;
> (ii) the general information and data gathering instruments … that contribute to the general information ... on the ... system."
>
> (1980, page 552)

MacCallum says that "designers operate within an environment and with a purpose which necessitates communication and interaction. ... In order to design, any system must interact with its environment." (page 59) Although this latter statement is very strong, it is at least clear that the best (most confident) information regarding the success of the design activity will come from observing the behavior of the implemented design in its environment. This would seem to be crucial in the adaptation process of the designer. Indeed, Simon describes the design system as an adaptive one and concludes from this premise that it is self-evident that feedback must reach the designer at some stage (possible all stages) of the design process.

It ought to be clear that the 'list' of components, or processes, discussed thus far by no means can be proved sufficient in a design context, nor that they necessarily constitute distinguishable processes. It ought to be equally clear, however, that Sternberg's seven components have relevance to design situations, and that the abilities required to perform such processes necessarily must be present during a design situation. Goel and Pirolli (1990) set out to investigate the relevance of the information processing theory to design, and found good correspondence between IPT and observed design activity. Newell and Simon (1972) used the same foundation, albeit from a different angle, and provided strong indications of the usefulness of IPT.

The remainder of this thesis is written from the perspective that design involves processes that can generally be discussed in terms as is done here, and that the seven components form a sufficient basis upon which to base further discussion. The literature discussed in the following, particularly those explicitly concerned with the design process, will therefore to some extent relate to this 'componential analysis',

## Goel and Pirolli -   An information processing approach

Goel and Pirolli (1990) investigated the behavior of three designers in three different tasks. They analyzed the findings in the context of information processing theory (IPT). The primary task of their research was to investigate whether there existed some generic characteristics of design. Their specific focus was on the way designers treat the problem space, or the information processing space (IPS). Before pursuing, it is useful to note Newell and Simon's definition, or rather description, of this IPS:

"1.    There is a set of elements, called *symbols.*

2.  A *symbol structure* consists of a set of *tokens* (equivalently, *instances* or *occurrences)* of symbols connected by a set of *relations.*

3.  A *memory* is a component of an IPS capable of storing and retaining symbol structures.

4.  An *information process* is a process that has symbol structures for (some of) its inputs or outputs.

5.  A *processor* is a component of an IPS consisting of:

    (a)      a (fixed) set of *elementary information processes* (eip's); a

    (b)      *short-term memory* (STM) that holds the input and output symbol structures of the eip's;

    (c)      an *interpreter* that determines the sequence of eip's to be executed by the IPS as a function of the symbol structures in STM.

6.  A symbol structure *designates* (equivalently, *references* or *points to)* an object if there exist information processes that admit the symbol structure as input and either:

    (a)      affect the object; or

        (b)       produce, as output, symbol structures that depend on the object.

7. A symbol structure is a *program* if (a) the object it designates is an information process and (b) the interpreter, if given the program, can execute the designated process. (literally this should read, "if given an input that designates the program.")

8. A symbol is *primitive* if its designation (or its creation) is fixed by the elementary information processes or by the external environment of the IPS."

(Newell and Simon, 1972, pp. 20-21.)

Goel and Pirolli studied various designers at work in concrete design tasks, and claimed from their findings eight characteristics of the establishment of the problem space, or eight generic aspects of design: (Goel and Pirolli, page 23)

(i)     The existence of many degrees of freedom entails extensive problem structuring.

The authors note that the design problem generally is under-defined, so that a massive structuring effort is needed. This structuring activity is recursive, in that the initial structure may be based upon lacking information - information that may become available as the problem-solving process unfolds. They found that tightening the problem definition throughout the process is important so that the designer is able to close in on specific solutions.

(ii)    Delayed or limited feedback from the environment entails extensive performance modeling,

The designer produces descriptions of a product meant to exist in the future rather than the product itself. Therefore, Goel and Pirolli say, it is impossible to know for certain how the product, or the artifact, indeed will behave before it is constructed. However, to construct the artifact from the specification means a committal of resources based on a specification that may be the result from wrong decisions, and any correcting measures in response to errors amounts to either (a) altering the existing artifact or (b) to learn and improve until the next design task is to be undertaken. In response to this, the authors state, the designer models the artifact to assess expected performance. This apparently reduces the risk of delivering a flawed specification for implementation or construction.

(iii)    The relative goodness of the answers entails use of personalized stopping rules.

As other researchers (Simon; Coyne and Gero), the authors recognize that there are normally "no right or wrong answers in design situations, only better or worse ones." (Goel and Pirolli, page 28). This is what Simon intimated by saying that the designer satisfices rather than optimizes. The set of decision rules thus are not unambiguous, which is to say that no set of decision rules will guarantee an objectively optimal artifact to be constructed. Goel and Pirolli observed that the rules determining what is perceived (by the designer) a satisfactory design are individual, or personalized.

(iv)    Nested evaluation loops in a limited commitment mode control strategy (LCMCS).[a]

The authors observed that the designer has conflicting objectives. She tends to 'freeze· the artifact specification while still wanting the freedom to alter it throughout the process. The investigated subjects handled this in two ways. (a) They worked through the entire artifact, completing the specification, before testing it; or (b) by evaluating artifact components as they are generated. LCMCS is a term for describing the designer's reaction to this situation, or their negotiation of "... tension between keeping options open for as long as possible and making commitments." Goel and Pirolli make the strong claim that this strategy "... is necessary due to the sequential nature of symbolic processing ..." (Ibid.) -  there is a limit to how many balls a person can keep in the air at one time.

(v)    The necessity for the designer to specify an artifact conflicts with the LCMCS.

---

[a]    LCMCS is a result of the combination of two approaches to evaluation of the design product. One of these approaches implies to start with a kernel idea, work through until complete specifications are complete, and then evaluate. The other implies the evaluation of artifact components as their specifications are generated. (Ibid., pp. 28-29)

In simple terms, Goel and Pirolli observed a serious conflict: The designer wants to fix solutions so as to more easily focus on other aspects of the problem, but wants to have the option to change the artifact as more information is gathered. "Designers are adept at negotiating this tension between keeping options open for as long as possible and making commitments." (page 28) Thus, there is a constant assessment as to whether it is worth the effort to 'unfix' solutions, altering them, and risking a subsequent effect on other design decisions made in the course of design. For instance, changing major dimensions may cause a complete redesign of the entire artifact, and may not be performed if the cost of such re-design exceeds the benefits from the change.

(vi)    The size   and    complexity   of  design problems    require   solution decomposition (with subsequent danger of 'leaky' modules).

In large (complex) problems it is impossible to keep track of all aspects of the solution. Designers thus decompose the problem (and solution) into modules in order to handle few problems at a time. Since there invariably will be inter-relationships between the modules, they needed a strategy for handling this situation. Two such strategies were identified: (a) Some modules were put on 'hold' while the attention was focused on solving others and  (b)  functional level assumptions were made to model those relationships.

(vii)   The necessity of abstraction hierarchies to relate goal and artifact.

While the decomposition described above results in so-called abstraction hierarchies, the authors noted a different kind of abstraction hierarchy. The goals or objectives tend to be expressed in quite abstract terms, for instance as functional requirements, while the artifact is generally expressed in concrete terms, as structural configuration. This causes a 'mismatch ' in terminology that the designer must resolve. To do this, she must in some way create a 'meeting point' of the two kinds of descriptions -  a meeting point that is closely related to the use of language.

(viii)    The use of artificial symbol systems.

In the solution decomposition and in the goal/structure resolution process the designer uses artificial symbol systems for representation. These symbol systems operate at different levels of abstraction to capture the different 'levels' of the goal formulation and structure specification.

Goel and Pirolli performs a very useful analysis of particularities in design by human designers. They identify some characteristics which may be called tactical concerns, such as those related to goal-decomposition and performance modeling, as well as more strategic characteristics, such as use of language and decision-making in the process. It is Interesting to note that their findings correspond well with Sternberg's information processing theory of mental abilities. Three of Sternberg's meta-components can be directly related to the results of Goel and Pirolli; (1) Decisions as to just what the problem is that needs to be solved (I -  "extensive problem structuring"); (2) selection of one or more representations for information (viii -  "use of artificial symbol systems"); and (3) solution monitoring (iii - "personalized stopping rules" and  ii - "extensive performance modelling").

Sternberg's  four other  meta-components are: (4) sensitivity to  external feedback;  (5) selection of lower-order components ('sub-ordinate' skills and tools); (6) selection of a strategy for combining lower-order components; and (7) decisions regarding allocation of attentional resources. Although these components are not directly distinguished by  Goel and Pirolli, their ·existence' is not  in conflict with Goel and Pirolli's findings. Indeed, (5) and (6) are implied under (i) "problem structuring," as an  organization of available knowledge. The difference  is that Sternberg Identifies these as so important that they deserve special treatment, separate from the problem structuring activity.

The implication from Goel and Pirolli Is that characteristics of observed design activity corresponds neatly with the more general theories concerning human cognitive abilities. The authors also cast light over some complex interactions occurring in design; the conflict between 'fixing ' solutions to make the rest of the task easier, while keeping solutions 'unfixed' to update them later on in the process. The research dealt with super-ordinate skills, and  did  therefore not delve into the tactical concerns, like for instance: How do designers choose representation language? How are tools chosen in the design process? How are different kinds of knowledge applied in particular situations? How does the designer map between an internal representation to an external representation? And, perhaps most important; How actually is language used in mapping between different levels of abstraction? Neither did the research focus on developing a theory for distinguishing the various abilities that the designer necessarily must possess - rather it identified and systematized the result achieved from enacting such abilities. These are relevant questions to be answered in the process towards developing a comprehensive theory of design.

## Mistree et. al. - "A Decision-Based Design paradigm"

In their paper *Decision-Based Design: A Contemporary Paradigm for Ship Design,* (1990) Mistree and others present a way of viewing the design process, claiming it to constitute a paradigmatic shift. Underlying their work is the fundamental premise that "... designers are and will continue to be involved in two primary activities, namely, *processing symbols* and *making decisions."* (page 565) Their concrete goal is to design a (computer-oriented) system through which models of different design processes can be created and used by the designer as guidance in producing artifacts. Their particular frame of reference is that "the principal role of a designer, in Decision-Based Design, is to make decisions ......In Decision-Based Design, decisions serve as markers to identify the progression of a design from initiation to implementation to termination." (page 572) Logically following this view is that, for instance, analytical tasks and information retrieval become in some sense subordinated to the decisions made to perform these tasks. Thus, where the 'typical' model is task-oriented without necessarily distinguishing decisions from other activities, Mistree et. al. place the decision central in the picture of the design process. When compared to the information processing theory of problem-solving by Sternberg, this view conspires well. Five out of Sternberg's seven meta-components deal explicitly with decisions and decision-focused activity (selection).

Mistree et. al. define the design process as "... a process of converting information that characterizes the needs and requirements for a product into knowledge about (?) a product." (page 573) In Decision-Based Design the process of the designer can be made more efficient and effective in two ways: (1) By increasing the speed with which design iterations are performed and (2) by reducing the number of such iterations. "An increment in the iteration speed can be achieved if at least some parts of a design process are modelled. To achieve a reduction in the number of iterations not only a model of the design process but also information that can be used to determine how the process can be improved is needed." The authors proceed to say that "without a model of the design process that can be represented and manipulated on a computer, it is impossible (!) to provide guidance that is suitable for improving the efficiency and effectiveness " (ibid.)

The design process is viewed as a set of decision blocks, where the relationships among these blocks are "not ordered and hence not directed." This is termed a heterarchical representation of the design process. Designing starts when "the first step is taken to extract a hierarchy from a heterarchy, that is, when the dominant (first) node ... is chosen." In other words, efficiency and effectiveness can be achieved through structuring the decisions involved in the design process from "... a pattern or structure (which is) re-entrant or recursive" into one that is ordered and directed. (Smith, 1992, page 49) There is thus a clear focus on *component selection and ordering,* in Sternberg's words, or upon the *language of plans* as described by Coyne and Gero later on in this chapter. Mistree et. al. are intending to give the computer the 'responsibility for these central activities.

The decision rules themselves are modelled and optimization is performed as a consequence of the overall structure (ordering and selection of decision tasks) of the *meta-model* and the *decision rules* underlying the various decision tasks. The decisions are taken as a consequence of these rules and the particular resolution methods, in this case Goal Programming. The decisions the designer normally makes are therefore to a large extent automated while based on (hitherto) internal rules that include Goel and Pirolli's *personalized stopping rules* in the process.

---

[b]    This is optimization of the decisions, based on a model including the decision rules and analysis based on Goal Programming. This is to be distinguished from optimization of the design judged from 'objective· criteria - Mistree et. al. are in agreement with Simon's view that designer's satlsfice (optimize their satisfaction) rather than optimize.

Two other conclusions from Goel and Pirolli are directly supportive of Mistree et. al; the observation that designers apply *solution decomposition* (treat problems separately) and subsequently use *abstraction hierarchies* (assess feasibility of or represent solutions at different levels of abstraction). This is particularly apparent in Mistree et. al.'s definition of *meta-design:* "A meta- level process for designing systems that includes partitioning ('decomposing ') the system for function, partitioning the design process into a set of decisions and planning the sequence in which these decisions will be made." (Mistree et. al., page 577)

For modelling the design process Mistree et. al. have developed a DSP palette of icons which are reproduced in Figure 2.3. In short, *phases, tasks,* and *events* designate gradually "smaller pieces'. "Events occur within a phase ... (and) tasks and decisions occur within events." Tasks may or may not contain other tasks and decisions. There are three kinds of decisions (expanded to four in later work - a 'heuristic decision' was added (see Smith, 1992)); *compromise,* which implies that there is a trade-off, *preliminary selection,* which. implies that a set of alternatives result, and *selection,* which implies that only a single candidate emerges. *Information* may be seen as a stock, or a data-base, from and to which activities extract and supply. *System* and *system variables* designate information pertaining to the design object, *Goals/Bounds/Constraints* belong to the problem information, and *analytical relationship* would seem to imply some sort of analysis performed to supply new information.

In lieu of the objective of Mistree et. al., which is to provide an efficient means of structuring a design process by means of the decisions, treating the system being designed as rather parametric, the DSP technique seems like an efficient means of attacking the problem. It holds good promise for enabling the designer herself model her own process, and to make interventions into the process to update the model as experiences are gained from failures and successes with previous models. Furthermore, it is an appealing thought that the process is optimized as seen from the perspective of the designer (from her own rules and sets of values) rather than from the design product according to some well-defined 'objective' criteria. This makes possible the introduction of "softer" values than those that monetary and performance standards express. In this respect it is an enlightening view of the design process.

---

[c]     Decision Support Modelling Technique.

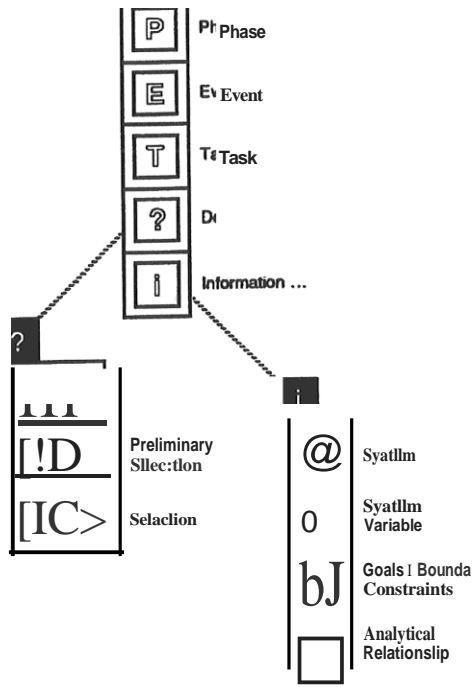

Figure 2.3:      DSPT palette from Mistree et. al.

In terms of the Design Process Language, however, the <u>information entities</u> introduced in the DSPT palette are not nearly sufficient to capture the knowledge needed in the design process. There Is no explicit mention of important entities such as *classes* or system parts (which themselves may or may not be systems). It is not necessarily so that it is feasible to define something as a system once and for all. A T-beam will normally not be treated as a system in ship design, although it certainly may be seen as a system of interacting molecules or as a system consisting of web and flange. It may be necessary to reason about 'abstract' concepts, not only about numerically valued parameters.

Another insufficiency is that the <u>system variables</u> in DSPT designate what is often referred to as a *property.* However, a statement (property) such as 'length of 125 meters' In "the ship has a length of 125 meters" gives more than one 'category· of information. It says that the ship <u>has a length</u>, that this length <u>has a value</u> 125 meters, and that the length <u>has a dimension</u> in meters. It is often appropriate, or even necessary, that the *attribute* (length) Is treated separately from its *value* (125 meters), and maybe even to view 'l 25 meters' as a composite itself. (See, for instance, Coyne et. al., 1990, pp. 92-93) Indeed, such a separation is crucial In the use of <u>frames</u> (Aamodt, page 154) or <u>prototypes</u> (Gero and Rosenman, 1990) common in computer-aided design systems.

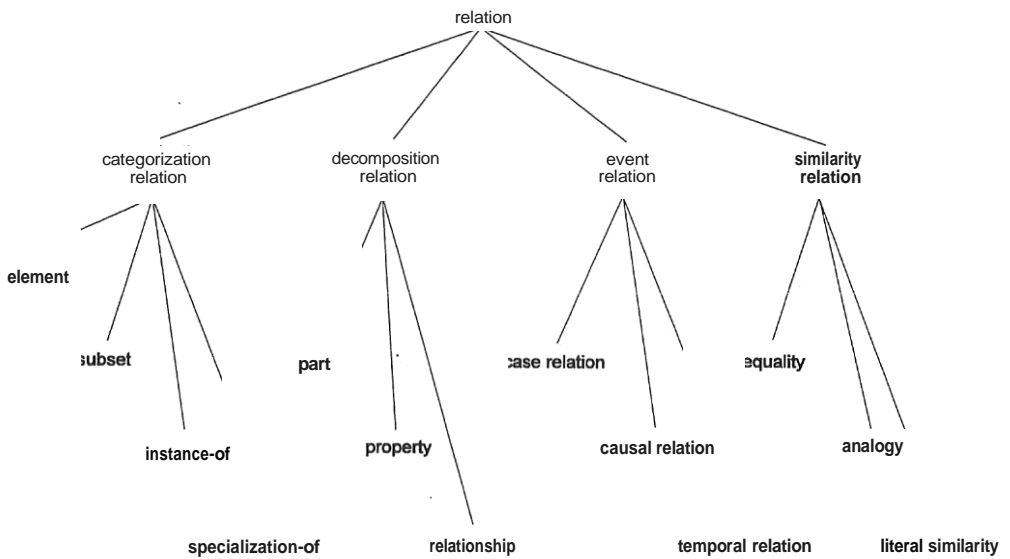

Figure 2.4:        "A taxonomy of conceptual relations,·adopted from Aamodt, 1991.

However, the most significant drawback with the DSPT terminology, seen in the context of the DPL is the sketchy treatment of the relationship. Only one kind of relationship is introduced, namely the *analytical relationship.* Viewed In contrast to the philosophy underlying Figure 2.4, for instance, it seems as though Mistree et. al. underestimate the importance of the concept of relations. As will become apparent towards the end of this chapter, the Design Process Language is founded upon the view that the relation is the most important concept in the language with which to describe the design process. I would claim that the DPL is the framework of a language with which the concepts In the DSPT palette can be described - I argue It to be a more basic language consisting of more fundamental concepts.

What separates the DSPT from other work, and presumably is the reason the authors claim It to constitute a paradigmatic shift, is that DSPT treat the decisions as milestones. The mapping of the design process is centered around the stages at which decisions are made. In most other work on design process modelling such mapping is centered around any task, whether analysis or decision.

---

[d]     One of the authors present in later work two new relationships in the DSPT palette; the
        conditional and the limiting relationship.  (Smith, page 56, 1992)

My own work maps the design process as a sequence of states where any new information introduced into the designer's memory (or domain) constitute a shift in states, similarly to what Hubka (1983, page 188) says: "Every design process may be structured into more or less complex partial processes, phases, and detailed design steps with the help of a general procedural model. The resulting procedural elements are also processes, within which the states of information are changed." This is similar to what Newell and Simon say, where the milestones are any processes that can be described in the language.

For a detailed analysis of the process, it is anyway difficult to distinguish between a decision and an analysis as done by Mistree et. al. - several analytical tasks, particularly when a computer program is applied, implies that decisions are used in the course of, or as a means of, analysis - it is not always possible to draw a sharp line between the two tasks'.[e] Thus, a meta-model of the design process would depend upon the kinds of tools applied for analysis, and the freedom to choose a kind of analyses suitable for the context would be limited since the decisions and other tasks are separated in the model *a priori* to the start of the process.

## Yoshikawa - A mapping approach

Yoshikawa (1979, 1981) views the design process as a sort of mathematical search in an ideal world where all solutions exist - the problem in the design process is to find these solutions. It must be noted that Yoshikawa primarily strives towards establishing a theory of design that can be useful in terms of efficiently applying the computer in the design process.

Yoshikawa's theory can be summarized as follows: There is a set of object $S$ which contains all objects which existed, does exist, or will exist some time in the future. These objects are completely defined by attributes and corresponding values. The objects have particular (or concrete) functions which are found In the set of functions, **F.** Thus, any object (that has a function) maps onto the set **F.** If all objects have a function, then $S \cap F \neq \emptyset$. He distinguishes between an <u>abstract function,</u> which denotes a function when this is not related to a particular object, and a <u>concrete function,</u> when it refers to the function of a particular object. (Thus, a functional description in a goal statement is abstract, provided it does not refer to a particular solution, and a functional description in an artifact specification is concrete.) The set of abstract functions is termed **T.** Yoshikawa recognizes that for every concrete function, there will necessarily be an abstract function. (That is, if there is a "pen can write," there must necessarily exist an abstract function "some things can write"). Therefore, $F \in T$.

---

[e] At the very least, most programs would use implicit decisions in the form of IF .. , THEN ... statements.

Yoshikawa places the concept of *topology* central in his theory. He makes a quite complicated argument to prove that **T** (and all subsets thereof) is a topology on **S,** and uses this consideration in developing the theory. He says that "the set of abstract concepts is a topology of the set of entity concept," where entity concept is defined as "a concept which one has formed according to the actual experience of an entity." (Yoshikawa, 1979, page 76) These entities 'contain· all information on attributes and values, without these being 'declared' explicitly. An abstract concept "is derived by the classification of concepts of entity according to the meaning or the value of entity." With these definitions, it is difficult to see the difference between a topology and a classification scheme. Therefore, I choose to interpret a topology as a description of the classification system, such that a perfect topology implies a perfect taxonomy.[1]

What he terms the "mathematical design process" is one of applying abstract functional descriptions onto the set of concrete functional descriptions, separating out **F'** as a subset of **F.** The resulting set of concrete functions **F'** $\in$ **F** is mapped back to the set of objects **S,** a resulting subset **S•** $\in$ **S** of which contain plausible functional design solutions. If there is more than one object in **S',** a new abstract function is applied to reduce the size of **F'** and subsequently **S•.** This process is repeated until **S•** contains only one object or, arguably, until the problem statements are 'exhausted'.

---

[1] Taxonomy is defined as "the science, laws, and principles of classification" (American Heritage Dictionary)

Yoshikawa identifies four different contexts in which such a mapping may take place:

  " (a)   design in a perfect function space,
   (b)   design in a space with perfect topology and with some objects missing,
   (c)   design in a space with imperfect topology and without missing objects,
   (d)   design in a space with imperfect topology and with some objects missing."

(Yoshikawa, 1981)

Case (a) implies a situation in which all historic, existing, and future objects exist, and with a perfect mapping between the abstract function space and the object space (perfect taxonomy). Thus, establishing search-criteria onto **T** marks *off* all possible solutions in **S;** the "specification is a filter."

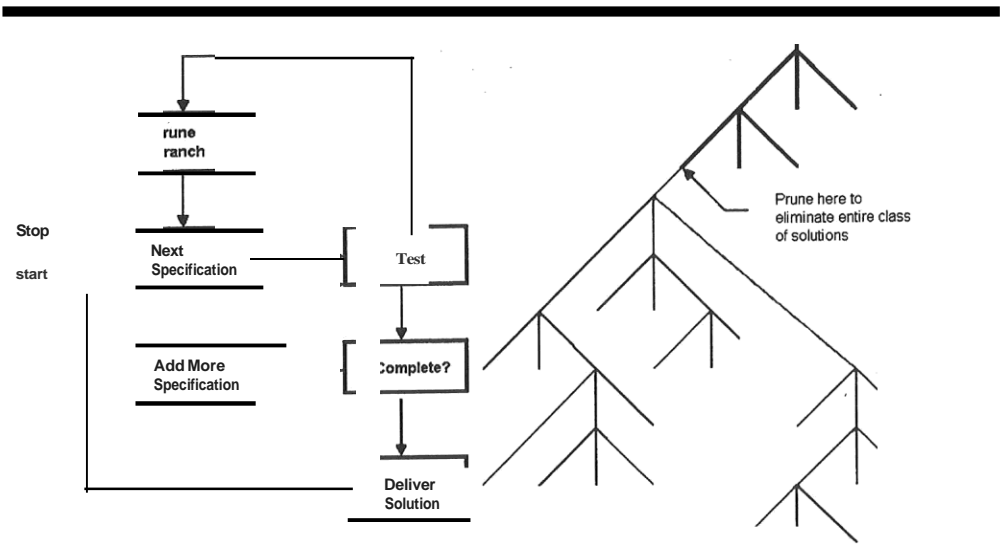

Figure 2.5:    Process illustration of an implementation of Yoshikawa's design theory. Adopted from Zeuthen, 1986.

Case (b), Yoshikawa says, "represents the essence of design in the actual sense, because design implies the creation of an object which does not exist in the actual world." Notwithstanding the particular definition proposed here, it is doubtful whether it is possible to claim the existence of a perfect topology in an actual design situation, since this would seem to demand that the designer knows all possible attributes - "perfect topology" - (although not necessarily knowing any objects possessing those attributes) and can map any specification In terms of these attributes. As my own work will claim, the designer may also need to 'design' attributes, or develop their meaning, In the process of creating an artifact description. For instance, there Is nothing prohibiting the designer from establishing 'girk' as a new function or attribute as long as it is defined properly in terms of known concepts: "To 'girk' is to be two places at once." The designer would have to represent 'girk' Into the system, rewriting the topology, before any mapping towards an entity that can 'girk' could take place. In view of this, case (d) would seem closer to actual design work.

Yoshikawa says later on in the 1981 paper that "the memory capacity of a designer is finite, so he can not memorize the complete set of objects and possibly all functional categorization or complete abstract function set (or a complete topology - my remark)" This would indeed imply that the realistic design scenario is case (d), although Yoshikawa presumes that the only problem is memory-related - not philosophical, or rather epistemological. In this imperfect situation, Yoshikawa discusses what may happen in a search for, or mapping towards, a design solution:

"(The designer) describes a specification t,

$t = t_1 \cap t_2 \cap ... \cap t_n$ (n : finite)

He will have the following cases:

a) He finds an object $s \in t$, (where) s is the design for the specification.

b) He finds a multiple solution, that is,

$s_1, s_2, ..., s_k \in t$.

Then, he must detail the specification by adding another abstract function:

$t' = t \cap t_{n+1}$,

and  check the objects in t'; If objects of more than one are still included in t', he repeats the process until only one object in t' ... , The new abstract function to be added to the initial specification must be created by the designer. This process needs the designer's skill. It should be noted here that the so-called design includes this 'designing· of specification.

  c) He finds no objects int. In other words, his memory is vacant for the given specification. Then, he starts narrow-sense designing. There are  three reasons why he finds no memory of object.

    (C1)    The intersection of the specified abstract functions is truly null set, due to logical contradiction of them.

    (C2)    His memory is really vacant correspondent to the specification.

    (C3)    He made a wrong mapping due to inaccurate structure of topology in his brain."

<div align="right">(Yoshikawa, 1981)</div>

It should be  clear from this that Yoshikawa views designing as a form of query into a database through a query (or search) condition based on any conceivable function or attribute. These functions or attributes  are all known. As soon as the query is performed, the set of matching objects contain all plausible solutions. If there is only one solution in the set, the  design process is completed. If there are multiple solutions, more conditions will be put  into the query to narrow the set.[9] If no objects appear, the query condition must be

---

[9]    The mapping suggested by Yoshikawa may be classified as a kind of discrete, combinatorial problem, similar to  that referred to by Zeuthen (1990, page 33): "Find in a finite, discrete space (analogous to F) a point (or a set of points) (analogous to $\{s1'\ s_{2'}\ "_{..}\ sn'\}$) satisfying the set of constraints  (analogous to t)" (Comments in square brackets added here.) To reduce the number of solutions, or $s_i$, various techniques are suggested, for instance integer

(linear) programming. This can  be shown by establishing a function:

$$G(s_i) = \prod_{j=1}^{n} H(t_j) \ \wedge \ (G(s_i) = 1 \ \Rightarrow s_i \in S^{I}\ ) \ where$$
$$H(t_j) = 1 \ if\ s_i\ satisfies\ specification\ j\ and\ 0\ otherwise$$

$G(s_i)$ is 1 if s, satisfies all  specifications tl' Application of an integer linear programming technique would seem an appropriate way to find the correct solution. This is a  quite well-known and non-controversial method to use in finding solutions in the kinds of situations discussed by  Yoshikawa.

relaxed (C1) or the designer must amend by either "selection of another (object) or by modification of the initial object."

Indeed, Zeuthen says of the Designer's Workbench (DWB), which is an implementation of Yoshikawa's theory, that"... the basic idea is to  prune (cut) the total space (tree) of solutions by means of heuristic knowledge." The process is illustrated through Figure 2.5 as adopted from Zeuthen.

The most significant contribution of Yoshikawa is that he introduces the concept of mapping (by means of attributes and functions) as a significant factor in the design process. Later in the paper, this  mapping is cast as a specific kind of <u>relationship</u> between the designer and  the designed object. It is also a  useful idea to visualize all concepts as existing, albeit not necessarily in the way that Yoshikawa does. This will be revisited later on. In addition, the idea of a structure relating functions (or functional attributes) and attributes to objects will be exploited. Yoshikawa does not, however, give any indication of what mechanisms, or abilities, to use in the mapping or in the modification of existing objects. Neither does he explicate how to <u>represent</u> these relations. Both of these are important in understanding design.

## Coyne and Gero -   "The linguistic paradigm"

Coyne and Gero develop in *Semantics and  the Organization of Knowledge in Design* a meta-language -   a design process language -   as a means for generating designs, or specifications of artifacts particularly as applied to the field of architecture. One of their main premises is that "design can  be discussed conveniently as operations within a language, (where) a system can be characterized as a language if it consists of a set of indivisible elements (an alphabet), a set of operations (such as union or difference), a vocabulary, and a grammar." (page 68) They say that one can talk of three languages in design; a language of form, a language of actions, and  a language of plans.

The *language of form* is essentially that which is applied to describe the design(ed) object. Its vocabulary is (in part) composed of the (physical) elements from which the object is built, and  the grammar is composed of rules that control the configuration of these elements, or their mutual relationships. Thus, in their domain of architecture a grammatical rule in this language could be the command: (1) "Place the kitchen next to the living room" and (2) "Put a door between the kitchen and the living room." It could also be a kind of production rule: "If this is a single-story house then place the kitchen next to the living room." *The kitchen, the living room,* and *door* are here vocabulary elements, as are presumably the other sentence segments, like *if, place* and *next to,* although Coyne and Gero are somewhat unclear on this point.

The *language of actions* is concerned with 'organizing' the language of form, or contains grammatical rules which control the selection and ordering (note the correspondence with Sternberg at page 17 - component selection and component ordering) of the rules governing the language of form. "The vocabulary ... consists of *planning tasks* such as ...:

*expand actions,*
*order actions,* etc." (page 80)

The vocabulary elements in the language of actions include the grammar of the language of form - the grammatical rules of the latter language are vocabulary elements to be related through the grammar of the prior language. Thus, a grammatical rule could be "First place the kitchen next to the living room and *then* put a door between the rooms" or formally "Enact Action (Command) 1 before Action (Command) 2." An *expansion* of the actions could cause: "In order to place the kitchen next to the living room *then* do the following:" Coyne and Gero say that the "artifact (final state) of the language consists of a sequence of actions (or rather commands, it seems) that provides the control for a language of form" (page 78) – the result arising from application of this language is "a statement about actions and their ordering." Coyne and Gero term this a *procedural network.* (See Figure 2.6.)

The *language of plans* is the language which is the primary focus of the two authors. They say that "knowledge about the selection and ordering of rules ('action' rules) constitute the grammar of actions in the same way that the rules in the language of form can be given names that enable them to be treated as vocabulary elements in the language of actions, it is possible to "give the rules in the language of action names and regard them as vocabulary elements in the meta-language called the *language of plans."*

---

Or action or (equest.

(page 80) The authors term this the language with which the *scheduling* of different actions is performed. The grammar consists of rules that help determine which plan (or sequence of actions) to pursue. The difference between a sentence in the language of action and that in the language of plans is that there may be several sentences in the prior language which describe sequences of actions that will bring about the goal. A sentence in the language of plans determine exactly which sequence of actions to choose, and at what time to effectuate various sequences. It is "essentially a method of controlling production systems." (page 72)

In essence, the two authors term the knowledge captured through this highest-order language *domain-independent* knowledge.

> "Rules for manipulating tasks could be based on general strategies for problem solving, such as
>
>   1. Consolidation
>   2. Diversification
>   3. Convergence
>   4. Divergence
>
> ... The control strategy for this system could be one in which only a single scheduling rule is adopted, or there may be an overriding strategy that states, for example,
>
> *Attempt to consolidate, and converge attention (whichever is applicable at any given state): if, after a particular period of time, no satisfactory end state is reached, then attempt to diversify."*

> (Coyne et. al., page 81)

Incidentally, the rule above would seem to be similar to Sternberg's *solution monitoring* component.

.In short, the system of languages developed by Coyne and Gero can be described as a three-levelled one in which the grammar of the lower-order language becomes the vocabulary elements of the language at the next level. Each language thus in some sense controls the construction (or at least the 'application) of sentences in the lower-order one. "From some initial representation of a design (the 'initial state), therefore, the grammar rules are applied to bring about changes .... It is possible to generate designs that conform to a syntactic design grammar." (Coyne and Gero, page 70) Underlying this view is that it "is possible to talk about mappings between actions in the same way that we talk about mappings between physical attributes." (Ibid., page 71) In the context of the present thesis, it is possible to replace the term *mapping* with *relation* and retain the general philosophy - design can then be viewed as the *act of building relations.*

The similarities between the approach of Coyne and Gero and my own approach are several. The view that design can be treated as operations within a language is similar. As are the notions of knowledge at multiple levels, and that this layering can be captured in some language (or languages), and the importance of *control* in the design process.[1] The most important difference is in the particular view of language. I will, in the DPL try to develop the framework for a language that is independent of the control-level at which the knowledge captured through the language 'exists. ' In this view, the approach of Coyne and Gero will be argued to represent a somewhat artificial segregation of knowledge.

Another important difference is in the actual structure of the language. Coyne and Gero primarily focusses on both vocabulary elements and grammar as consisting of rules. The approach pursued through the DPL may be seen to be a more 'linguistic' one. Some of the vocabulary elements are: Kinds of relations, for instance verbal ("I *have* a car"), conjunctive ("I have a car *and* a boat"), and prepositional ("The car is *in* the garage"); classes of 'objects' as operated upon by the relations ("The *(system)* car *(relation)* has an *(attribute)* color *(value)* red"), and modal relations ("I *should* have a car"). This implies also that a discussion of the *composition* of the vocabulary elements for Coyne and Gero's language of form is performed - while they treat these elements as "dimensionless spaces" with no particular meaning attached, grammatical rules are applied in the DPL to assign such 'meaning' to the elements. The rules in DPL thus operate at a level beyond that of the two authors.

---

[1] In later work, Coyne and Gero (Coyne et. al., 1991, page 37) separates knowledge from control, which counters the view in the 1986 work of meta-languages as crucial in representing control. I will disagree with the view that separating the two is appropriate - control may be viewed as being a particular <u>aspect</u> of 'any' knowledge rather than being a particular <u>kind of</u> knowledge.

Before concretizing some of the differences in viewpoint, it is useful to extract one example from the paper of the two authors. At page 78 they give an example of sentences constructed as *grammar* in the language of form:

*put A east_of B*
*put A east_of Band north_of* C

These are treated as *vocabulary* in the language of action, the grammatical rules in which will expand and sequence the sentences above. *To put A east_of B then do* If we view the result from the successful application of these sentences, the following sentences result:

*A is east_of B*
*A is east_of Band north_of* C,

respectively. Intuitively it would seem that these are sentences in the language of form, without this being too clear from Coyne and Gero. However, establishing concrete different rules for the two cases (grammatical rules in the language of form) would seem to elude the fact that the *control* in these sentences is carried by the verbal relation *put* and *is* (the degree of transitive verbal relations partly captures the control-issue in the language). Wouldn't it then seem reasonable to let grammatical rules operate upon these relations rather than the sentence as a whole?

In Figure 2.6, Coyne and Gero exemplifies an application of the language of actions. The particular tasks described, painting a ceiling and painting the ladder (state 1), are expanded (state 2) and ordered (state 3) - one should not paint the ladder before painting the ceiling using that ladder. The expansion and ordering are performed through *grammatical rules* in the language of actions, while the particular actions *(paint the ceiling, paint the ladder)* to be ordered and expanded constitute the *vocabulary.* This vocabulary consequently constitute the *grammatical rules* of the language of form. Language of plans is not explicitly used in their example, but it could presumably consist of grammar saying something to the extent:

Grammatical rule (Involving Sternberg's problem structuring meta-component):

> *The particular context is to paint the ceiling. In this context there is a choice among the following plans: A, B, or C. Apply grammatical rule II to schedule actions (combine vocabulary elements).*

Grammatical rule II (Involving Sternberg's component-selection (or ordering) meta-components):

> *If the task is to paint the ceiling, there are some sets of actions (vocabulary elements) that will help achieve this: Plan A - 7) get paint, 2) get ladder, and 3) apply paint to ceiling: Plan B – get a pneumatic painting*

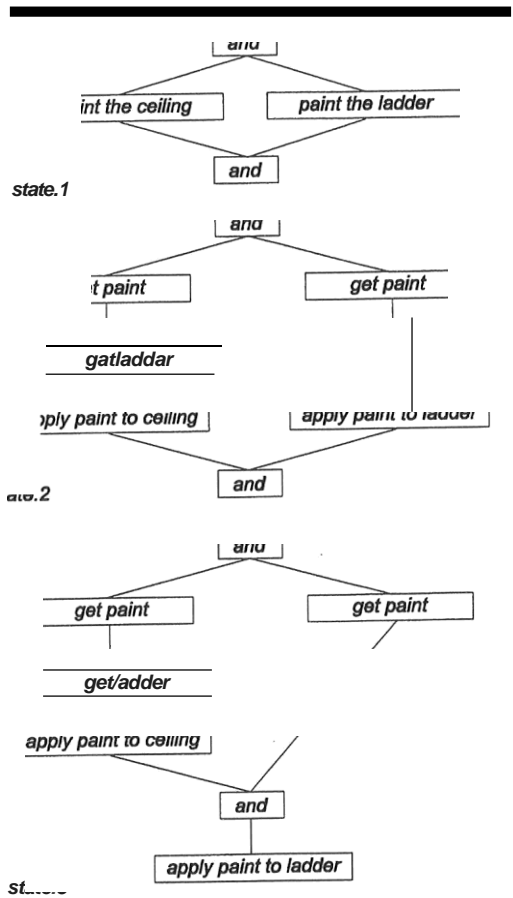

Figure 2.6:     Example from Coyne and Gero
- language of actions

*machine, 2) get paint, and 3) spray paint to the ceiling from the floor: and Plan C - 7) get paint and 2) throw the paint upwards to the ceiling. Plan (or strategy or schedule) A should be selected in this particular context.*

Parts of this situation (that of painting the ceiling) could be described in the DPL like:

1. *"should 'paint the ceiling'"*ₐ               Known

2. *"must 'apply paint to ceiling' to 'paint the ceiling'"'*     Known

3. *"must 'have paint' to 'apply paint to ceiling'"*		Known

4. *"must 'have ladder' to 'apply paint to ceiling'",*		Known

5. *"should 'apply paint to ceiling '"*		Match I and 2

6. *"must 'have paint'"*		Match 3 and 5

7. *"must·'have ladder'"*		Match 4 and 5

8. *"have paint"*		Known

9. *"have not ladder"*		Known

I O. *"must 'get ladder' to 'have*		Match 7 and 9

11. *"'get ladder' to 'have ladder'"*		Eliminate *must* from I0

12. *"get ladder"*		Match 7 and 11

13. *"apply paint to ceiling"*		Eliminate *should* from 5

14. *"applies paint to the ceiling"*		Match 2 and 13 and  general, grammatical knowledge about changing infinitive to present)

15. *"applied paint to the ceiling"*		General, grammatical knowledge about changing present to past

This is by no means a complete example. For one, the choice of actions would certainly depend on the context, for instance height of ceiling above the floor, height of the painter, and so forth. The grammatical rules needed for 'understanding' the various keywords are also numerous and complicated. However, the example ought to illustrate major differences between the approach chosen by Coyne and Gero and my own approach. As they themselves note, their own approach is primarily a means of producing and controlling production systems *(If* ... *Then* ... rules).

In the DPL rules would be applied primarily to control *the application* of the grammar, like: "If there is a *should* in the sentence, then this indicates (...)" or even "An *if* in a sentence indicates (...)." Rules are tied to keywords, or groups of keywords (primarily relations). In the above example, *should* is a relation indicating a desired or necessary state or action (or actually sentence or proposition) and *must* is a relation indicating a required state or action. For instance, *"should 'have ladder'"* signals a desire to reach a (future) state where the sentence "have ladder" exists. Both of these words belong to a group signaling things to be done - in DPL called *modal verb relations* (also including *will, can,* etc.). Other groups of words are the preposition relations (to), conjunction relations (and, *or),* and verb relations *(have, get).*

A rule could say that "if *must have* and *not have* then get."J More (grammatical) rules could expand *'must have·* to find what is the consequence if (possessive verb relation) *have* is preceded by (modal verb relation) *must.* The <u>tense</u> of the verbal relations might be captured in rules like "If a sentence involving the infinitive tense of a transitive verbal phrase is <u>established</u> then a new sentence is constructed where the verbal relation is replaced with its present tense" ("Apply paint," as a command, is immediately followed by "Applies paint") and "If a relationship involving the present tense of a transitive verbal phrase is <u>removed</u> then a new proposition is constructed where the verbal relation is replaced with its past tense." ("Applies paint" is replaced with "Applied paint" upon 'terminating' the activity) This is only sketchy indication of the difference between Design Process Language and the 'linguistic paradigm' of Coyne and Gero. The principles underlying the DPL are developed in more detail in Chapter 4 and Chapter 5.

## Chapter summary

I had two main intentions with discussing the literature in this chapter. One of these was to disclose whether design may be viewed as general intelligent behavior, in specific the kind of intelligent behavior that humans display in problem solving situations. Upon comparing literature in design with the theory of intelligence as proposed by Sternberg, I found that there are several common points. All major processes that Sternberg, as a proponent of the Information Processing Theory, argued to be important in intelligent creatures

---

The idea that some relations *(if, and)* are used to establish grammatical rules for relations at another level ("if an *??* then ...') should not constitute difficulties. This is what occurs in natural language, where concepts and relations are defined by means of already defined (or known) concepts and relations.

were in some way traceable into a wide specter of works in the field of design research. This set the stage for discussing design in the context of human behavior.

The second intention, of most importance to the subsequent discussion in this thesis, was to show that when researchers discuss design they really discuss how things relate. Whether these relationships concern how activities are ordered, how decision rules map between a condition and an action, how an object search space is organized, how objects in that search space are located, or how decisions are ordered and what is decided upon, the fundamental question is still how things relate to each other to form a coherent whole that expresses ideas about how design may be described. The theories are built by discovering regularities, and these regularities are described through establishing relationships among known things - unknown concepts are described by relating them to known concepts. I will in the remainder of this thesis take the Idea of the relations further to see whether it can bring some understanding of what design is about or, at least, how a design process and a design object can be described.

# Chapter 3: Formal knowledge representation

The idea that <u>relations</u> are central in representing knowledge has been pursued by several researchers in several disciplines. In Chapter 2 it was shown that the general notion of relation could be traced into a wide spectrum of literature in design research and into all areas of design; from those focused on the design object to those focused on the design process.

The areas of study where <u>symbolic representation of knowledge</u> has been treated most rigorously are those areas focused on developing computer applications, particularly within *artificial intelligence.* An obvious reason for this is that the computer traditionally has required the knowledge to be <u>explicit</u> in order to process it - no knowledge can be assumed to exist unless it is in some way directly represented symbolically or can be drawn from information that is represented symbolically. Researchers who focus on enabling computers to 'behave' in domains of knowledge manipulation, are bound to "take knowledge seriously." It is therefore useful to review briefly the way knowledge is indeed treated in these fields.

## Main-stream conceptions of relations

This section discusses in brief *predicate calculus, production rule systems, connectionist models, entity-relationship models,* and *object-oriented programming.* The purpose is not to provide a comprehensive and exhaustive discussion of all strengths and weaknesses of the techniques - the primary focus is on the use of relations between concepts in some form or another, and later on to show that relations are essentially what constitutes the common denominator in these languages. In the next section the various kinds of languages are discussed in terms of strengths and weaknesses as far as design goes.

### Predicate calculus

In *artificial intelligence* one of the more predominant and earliest languages used for representing and manipulating knowledge was *predicate calculus.* Predicate calculus is essentially a kind of language intended for applying formal rules of logic to make inferences in a well-structured domain.    A
sentence is in predicate calculus called a *proposition.*

"There are two types of *symbols* in predicate calculus: variables and
constants. Constants are further subdivided into object constants,
function constants, and relation constants   Variables are used to
express properties of objects in the universe of discourse[a] without
explicitly naming them. An *object constant* is used to name specific
element of a universe of discourse     A *function constant* is used to
designate a function on members of the universe of discourse.   A
*relation constant* is used to name a relation on the universe of
discourse. Every relation constant is either a mathematical operator
or a sequence of alphabetic characters or digits. ... The following
symbols are examples.

| | | |
|---|---|---|
| Odd | Parent | Above |
| Even | Relative | Between |
| Prime | Neighbor | Nearby |

In addition, every *n-ary* function constant can also be used as an
*(n+*1)-ary relation constant[b]"

(Genesereth and Nilsson, pp. 14-15, footnotes added here)

Predicate calculus is a quite pragmatic way of arranging the language. For
instance, the reason for *Parent* to be cast as a relation constant instead of an
object constant is to ease the representation of and operations upon the fact
that "Anne Lise is the mother of Bill," which turns out in this world to be a
proposition like *Mother (*Anne Lise, Bill), where both objects and relation
constants are easily identified. One problem arises, however, when we state
the fact "Anne Lise is a mother." This can be accomplished in two ways; either

---

[a]    "The set of objects about which knowledge is being expressed is often called a *universe
of discourse.·* (Genesereth and Nilsson. page 10)

[b]    *n* refers to the number of 'arguments' in the relation -   the relation *Above* would be a
binary (n=2) constant, for instance Above ( A, B ). However, In a different language
*Above* might even be unary, like in *Is(*A *Above ( B))* -   A is above B! *Between* could be
cast as a ternary constant, like in *Between(*A *B, C)* -   the grammar (rules of
interpretation) determines which is between which.

(i) as ∃ x Mother (Anne Lise, x), where x is a variable, or (ii) through establishing two relation constants, *Mother-of* and *Mother,* where the former is binary *(Mother-of* (Anne Lise, Bill)) and the latter is unary *(Mother*(Anne Lise)).

The problem with (i) above is that the proposition contains unnecessary information - the general understanding of the concept *mother* implies that there is someone to which the person is a mother.[c] The language becomes still more tedious when we state "The mother of Bill is the neighbor of Bill." Presumably this would read something like ∃ x Mother( x, Bill) ∧ *Neighbor ( x, Bill).* The problem with (ii) above is that two definitions, or 'grammatical' rules, are necessary, and the fact is eluded that the concept of *mother* has characteristics that are constant regardless of the context in which the concept appears. Thus, it should be possible to use the same definition, or concept, of mother whether we state that "Anne Lise is a mother," "Anne Lise is the mother of Bill," or "Anne Lise is the mother of Bill and Mary." Incidentally, *Mother-of* is a combination of concepts, where *of* is a prepositional relation.

This inefficiency of language is obvious in the following segment of a Prolog code, cited from McCarthy (1986, page 261):

> *"sterile (X) :- sealed (X), not alive-bacterium (X).*
> *alive-bacterium (X) :- in (Y, X ), bacterium (Y), alive (Y)"*

(Note that":-" stands for 'if' - *reverse implication(<=))*

In the cite above, the term *alive* appears both alone - as a separate function constant - and in conjunction with the term *bacterium* which is also a separate function constant. They are joined into the composite *alive-bacterium* because it becomes tedious in this language to use the elementary terms (like in line two above) every time a reference is needed. There is no way, in predicate calculus, to merge two function constants, such as *alive* and *bacterium,* in a way like it is done in natural language. In daily speaking the knowledge contained in the cited propositions could be delivered in the form: *"X is* sterile *if* it (or X) *is* sealed *and* no alive bacterium *is in* it (or X)."

---

[c]    Anyway, in general language the statements that "Anne Lise is a mother· and "Anne Lise is a mother of someone" might not be considered identical.

## Production rules

In other approaches to artificial intelligence, *production rule systems* are frequently used in the knowledge representation task. In this language, knowledge is represented as condition-action rules, or *"If A Then B"* kinds of rules. If the condition A is true then the consequent B follows. The three languages of Coyne et. al., discussed in Chapter 2, were essentially constructed from such production rules. It is interesting to note what Holland et. al. say of this kind of language: "Rules are (the) building blocks; but for efficient operation a processing system must have its rules organized in relation to each other. *Implicit* organization arises from patterns of conditions and actions: if the action of rule 1 satisfies the condition of rule 2, then the firing of rule 1 will tend to lead to the firing of rule 2. (Coyne et. al.'s language of actions) *Explicit* organization can come through pointers that are used to link rules directly together. (Coyne et. al.'s language of plans)" (Holland et. al, 1989, page 17, comments in brackets added here.) The relations are again seen to be central.

In advanced systems, there are several kinds of rules where each rule has a strength in performance (which is a function of confidence and relevance, among other measures). Competition, or selection, among rules in such production systems is based on their relative strength - a kind of "survival of the fittest" strategy, Production systems are somewhat similar to stimulus-response *(s-r)* theories in psychology and Test-Operate-Test-Exit modules *(TOTEs)* in machine behavior and, incidentally, also to a large number of models suggested to describe or prescribe the design process. (See, for instance, Figure 2.2. Asimow implicitly introduced several TOTEs - analyze and evaluate sequences - as elements in the design process.)

## Connectionist models

*Neural networks* can generally be described as follows: All concepts are viewed as nodes in a network, related in some way through connections. Most connections are of zero value (that is, unrelated concepts) and some have value close to one (closely related concepts). In this model, some conceive a problem-solving task to start with presentation of a problem which activates various nodes in the network, the activation of nodes (or concepts) depending on the strength with which they are associated to the problem. In short, concepts are defined by means of their connection strength to other concepts, and solutions to problems are the concepts that get the highest 'score.' "In this sense, 'the knowledge is in the connections,' as the connection theorists like to put it, rather than in static and monolithic representations of concepts." (Holland et. al., pp. 25-26)

A close relative of the neural networks in psychological theory are *semantic network models,* where memory is seen as consisting of connected nodes (concepts). In the simplest of these models, the connections between concepts are assumed to be of different 'lengths'. The distance between two nodes represents the degree of association between the concepts that these nodes represent. (Klatzky, 1980) In the more complex of them, the relationship between them is not only one of simple correlation, but each concept is associated also to the role it plays with respect to other concepts, as agent, object, goal, and so forth. (Sowa, 1984) The results from activating the 'simple' network models is an association among concepts, whereas an activation of the more complex models are more similar to complete sentences. Incidentally, these 'connectionist' models are in some way congruent with Yoshikawa's view of the world as a highly organized set of concepts, where the closest fit to the problem statements (incremental constraints) is the design solution.

## Entity-relationship models

Outside the specific domain of artificial intelligence and expert systems, database modelers in particular and computer programmers in general are quite involved with the issue of knowledge representations. One of the more pervasive approach of the prior has been that of using *entity-relationship* (e-r) models. The thinking behind e-r models is actually not unrelated to that behind the connectionist models. All entities are defined by means of other entities to which they are explicitly related. At the 'bottom' of such a structure are *kernel entities,* meaning that they "can exist on their own without being related to other entities." (Tsichristzis and Lochovski, page 115) The knowledge is captured through the relations in both connectionist models and e-r models. In addition, the applied relationships are quite similar to the underlined predicates, or relation constants, and the entities to underlined object constants in first order predicate calculus.[d]

## Object oriented programming

*Object oriented programming* (OOP) has received much support among computer programmers and data modelers. It essentially is a means of structuring knowledge as class-subclass-instance-property relationships, where each subclass and instance inherits and holds some properties and methods from its ancestors while new properties and methods may be defined at the 'subclass' level. Each object 'holds' knowledge of the methods that may affect it, and operations are performed through instructions to the objects telling them "what to do with themselves." One of the major advantages to representing knowledge within OOP is the aspect of *inheritance* of properties from a general category to a less general category of concepts. Thus, *mother* and *father* both share some properties defined in a general class, for instance that of *persons,* while having specific properties defined in more specific classes (e.g., those of *female* and *male).* Multiple inheritance is allowed, implying that one concept may inherit properties from more than one (immediately) general category (or concept).

OOP also brings with it the possibility of *encapsulation,* leaving some knowledge regarding an object open for *public* scrutiny while other knowledge is *private,* in OOP terms. Such encapsulation tends to stabilize and provide security in a model, in the sense that only that information which is allowed to change may change.

---

[d]     Functions (or rather transactions and interventions) cannot be represented in the traditional e-r model. However, recent work has merged object-oriented programming and the e-r models to include methods (transactions) to be represented in e-r terminology and to allow for inheritance through, for instance, *HasSubclass* relationships. (See Dale, 1991)

# Why traditional languages are insufficient

Most of the knowledge organization or representation languages[e] described above have virtues in particular contexts or applications. Except from the connectionist models, they all have been implemented in some way and proven to be useful in processes involving transactions on knowledge. However, there are some features in all of them that make them insufficient for the present purposes; to enable design process and design object to be modelled in the same language.

The present section discusses each of these languages that are briefly described earlier, with a special emphasis on the problems with object-oriented programming languages for the reason that OOP is frequently considered in the context of knowledge representation for automated design and life-long product models.

Predicate calculus is in general based on formal logical rules for operations on objects, where the objects carry no meaning in and for themselves. For the most part, predicate calculus requires the problem to be well-structured, particularly in the sense that the initial condition should contain all the rules needed to reach the final state. In addition, the objects in the universe of discourse cannot themselves be functions or relations in first-order predicate calculus (FOPC). If logical operations are to be performed that change the 'meaning' of existing function constants or relation constants or that introduce new ones, this is higher-order predicate calculus which brings with it serious problems in terms of logical consistency and stability in knowledge manipulation. (Minsky, 1986). The efficient use of predicate calculus is therefore essentially limited to FOPC and to well-structured problem domains. Since design is shown to introduce functions and relations at several levels (Coyne et. al.'s language system included three such levels) predicate calculus is of limited use. In addition, since the design problem cannot generally be assumed to be well-structured predicate calculus seems unfeasible.

---

[e]    It is realized that probably none of the representation models discussed above would
satisfy the requirements as a complete language. The term language is used here in a
very loose sense, as a reference to the particular characteristic features each
'scheme· applies in representing and manipulating knowledge.

Predicate calculus thus does not seem to do the trick in establishing a language with which to describe an arbitrary universe of discourse and a process changing this universe. It is, however, useful to note that the function constants may be viewed as relation constants in predicate calculus, as in *above (A, B) (above* as relation constant) as compared to *is (A, above (B)) (above* as function constant). This conspires well with what was said earlier; interventions and transactions in the design process may be viewed as relations between that which performs and that which is performed - 'functions' may be viewed as relations.

Production systems have a great virtue in their effective control of operations in design, which several have shown of empirical importance in design (see discussion of Coyne et. al. in the previous chapter). The main problem, however, is that a problem-solving system based on *if-then* relationships requires two important features aside from an understanding of the semantics of the two terms *if* and *then* - it requires an 'understanding' of the antecedent (the condition) and an understanding of the consequent (the action). This issue is frequently eluded when these expert systems are discussed. There is often an implicit assumption that the condition and the action can be treated as statements that are considered true if their exact match is found in memory. In other words, a rule like *"If* 'length is larger than maximum length' *then* 'reduce length'" 'fires' on finding the string of characters (A) "length is larger than maximum length." (The sentence "reduce length" is then hopefully to be found as a condition in at least one other rule.) If the sentence (A) did not exist, on the other hand, the only way it could be found were through applying another rule, for instance *"If* 'Value of length is larger than value of maximum length' *then* 'length is larger than maximum length '."[f]

Since there is no semantics tied to other words than *if* and *then* (except from some 'logical operators' like *or, and, not,* and the like) and assuming that (A) is not stored as a string of characters, there must necessarily exist a means (rules) of transforming the stored information to a form matching the condition of a rule. That is, to assess the truth value of a condition. Should one then store all information in such a form that it matched all relevant conditions? It would then seem that when new information is added to memory one would have to investigate to which rules that information could be relevant and have to adjust the manner of storage. If semantics were tied to all relations, like *larger than,* that semantics could help assess whether the condition were true or not, without necessarily generating specific sentences to spell out that fact. With proper semantics, the knowledge could be implicit - without it, it has to be explicit.

---

[f]     In the end there must be some semantics tied to other relations, for instance to resolve
        the truth of relationship involving *larger than.*

In short, production systems are generally more logical than they are expressive. They are superb for mapping from an initial state to a final state as sequences of decisions. However, when it comes to expressing and interpreting facts of the world, those facts are ultimately stored as strings of characters, where no other words carry meaning than the few keywords in these 'languages'. Thus, the only way concepts are related is through the 'logical' relations as mentioned earlier. As viewed in the context of representing the design object and design process, as discussed in Chapter 2 and in remaining chapters, it is not a sufficient terminology. Production rules and the keywords used are still important, but as will be shown in Chapter 5 they constitute a particular application of grammar - not a complete language, and the terminology used is subset, important as it is, of a larger grammatical system. This terminology must be understood to make production rules function efficiently.

Connectionist models have great explanatory powers in their explication of the issue of knowledge as relationships. The idea that a set of nodes 'light up' as a consequence of a particular set of facts, where the highlighted nodes are 'relevant nodes', is an idea that seems intuitively appealing. However, consider these nodes as all words in a language, and the relations between these nodes as dimensionless 'values'. The relations in the language must then themselves be nodes in the network - sentences would then be formed as a constellation of concept nodes and relation nodes, where the strength of the connections decided which were related with which. However, the sequence of the words is certainly a major factor in determining what the sentence means. Assume that the nodes *vessel, large, is,* and *the,* form a 'strong quadruple'. The words can be combined into two meaningful sentences; a question - "is the vessel large?" - or into a statement of fact "the vessel is large." Which is appropriate and what determines what indeed are 'meaningful' sentences? The problem grows, of course, if there is need to represent complex facts as in the production systems.[g]

The problem is addressed in some interpretations of these models, where the role that concepts play in relations to other concepts is also included in the network. However, in this view it seems that what is stored is more <u>sentences</u> than disparate nodes in a network. Another problem is that there would be a host of concluding 'sentences' that all could seem to pertain to the solution, whereas all but a few such relationships might be irrelevant information.

The connectionist models thus do not provide a means of representing knowledge in a form that makes it suitable for use in practical matters. Most of them are anyway as yet theoretical constructs that face quite significant problems in terms of implementation.

Although <u>object-oriented</u> <u>programming</u> gives good tools for representing knowledge, in particular through class-subclass inheritance, and are highly efficient in computer programming, they do have some characteristics that introduces problems when used in design situations. Let me first discuss a situation where OOP is of obvious utility - that situation which exists in Yoshikawa's theory (see Chapter 2).

Assume that the design process can be treated as occurring in a situation where the topology, in Yoshikawa's words, is ideal. In this case all concepts (objects or entities) would belong to at least one class, and all imaginable classes would be represented in the topology. Assume further that the design process proceeds along the pattern roughly suggested by Yoshikawa;

(1)     Study the problem statement and pick out one aspect of it (one function that the D-object is to serve or one property that it is to hold).

---

[g]     It could be argued that if-then relationships are superfluous since the problem-solving is a consequence of the model itself. However, in most realistic cases the rules should be overt -  the designer may want to plan actions according to what information is necessary to meet conditions. Such knowledge might even come from outside the system if the concept in question does not exist in the model yet.

(2)     Find the most general class (or a set of most general classes[h]) of members that can serve this function or hold this property.

(3)     Repeat with the next function or property, and select from the set of most general classes the set of most general classes that can serve the new function or hold the new property.

(4)     When all the information in the knowledge statement is 'exhausted· then either (i) select one of the classes remaining in the set and assign values to the attributes that are unvalued or (ii) select an instance of these classes (where values are already assigned to attributes) or (iii) let one (or all) of the class definitions be the design solution (that is, assume such a class description to be a sufficiently detailed specification of the design object since the problem statement does not express more 'expectations' from it).

Is this problem-solving framework sufficient to produce designs? Intuitively it would seem to at least include aspects of the design, in particular those aspects that deal with *prototype selection and refinement.* However, a problem arises if any of the following conditions are present:

(a)     The topology is not ideal, which is to say that there is at least one property or function which the D-object is expected to hold or serve that may not be included in any class definition.

(b)     The solution to the problem demands a new class of objects (or concepts) to appear, where the D-object or the secondary concepts constitute a hitherto non-existing combination of properties and functions (or relationships among concepts).

(c)     The particular approach that the designer chooses to pursue is *bottom-up* (decompose problem and find solutions at a detailed level) rather than *top-down* (find a general solution and specify components that helps meet the overall expectations).

---

[h]     It is not necessarily so that there is only one general class whose subclasses/members satisfy the requirement. For instance. the functional expectation "Should have sanitary facilities" would bring out a host of classes, including '(most) buildings', '(some) trains', '(some) airplanes', and '(some) ships', rather than the more general 'man-made things' to which all these classes arguably could belong. If the next functional expectation is "Should be able to transport containers at the sea" one would probably end up with the class of '(some) ships'. To get a strict (or perfect) class-subclass hierarchy as more expectations are treated would demand that the problem statement were structured heavily according to the 'topology' such that a progressive application of functional expectations causes a monotonic reduction of one general class into one subclass or member of this class.

(d)     The designer gradually creates relationships to incrementally describe the D-object (as discussed in Chapter 4).

The situations (a) and (b) are related to the subjects of class and concept formation. OOP rests firmly upon the view that objects belong to *a priori* defined classes (the ancestors). In other words, OOP singles out particular classes for "special treatment· from which the 'children' are to inherit properties. Thus, a car may be classified as "Cars subclass-of Vehicles." More untraditional or unexpected classes, like in "Cars subclass-of Things_on_wheels" or "Cars subclass-of Killing_devices" would probably not be considered as represented in the class/subclass hierarchy, and a redefinition of the ·topology' would likely be needed for efficient representation. Such addition of classes is hazardous in OOP, particularly due to the danger of making a language that McDermott (1987, page 158) calls 'Computerdeutsch' - classes are defined that capture the aspect of the object that is suitable for the present purposes, but which becomes absurd when stretched to the theoretical limits. McDermott's argument, which essentially dealt with a critique of predicate calculus, could easily be extended to OOP as well. If a 'perfect' topology, in Yoshikawa's words, were to be constructed, there would arguably be an infinity of classes to which 'car' could be a subclass.

---

"Which classification is correct? Silly question! It depends on what you want to use it for.· (Minsky, 1986, page 91)

**Box3.i**

---

In the case of (c) above, in a bottom-up approach to design rather than top-down, it is difficult to see how class-subclass arrangements are complete enough to sufficiently support the design process. As elements are chosen, the design object gradually becomes defined. However, in this case it does not make sense to move upwards in a hierarchy of classes, since this would only bring forth a more general class of that element - not of the system to which that element were to be assigned. For instance, the concept *diesel engine* might be considered a member of the class 'engines' (as well as other classes), and little would be said regarding to what conceivable systems this diesel engine could be related (at least not in the traditional view of class/subclass hierarchies as one of general/less general relationships). The aid of an object-oriented language (in the understanding OOP) is essentially limited to either (1) *instantiation,* which implies identification of instances of *diesel engine,* or (2) *abstraction,* which implies to move "upwards' to the more general class 'engines' to find alternative (classes of) solutions. There is evidence that both abstraction and instantiation are applied in, and may be important characteristics of certain design situations (see, for instance, Goel and Pirolli). However, it is not likely that they are sufficient tools in most situations, particularly in those described under (a) and (b) above.

---

Example of a class-definition set-up in OOP:

```
Class Things_on_wheels
{

};

Class Vehicle
{

};

Class Killing_devices
{

};

Class Car: Vehicle, Things_on_wheels, Killing_devices
{

};
```

The same set-up in predicate calculus:

$$\forall \text{ Car (x) => Vehicle (x)}$$
$$\forall \text{ Car (x) => Thing\_on\_wheels (x)}$$
$$\forall \text{ Car (x) => Killing\_device (x)}$$

The parallel between OOP and predicate calculus should be reasonably clear.

**Box 3.it**

---

In case of (d) above the design object is described as relationships that increasingly relate the object to other known things and facts - that is, that gradually define the object. In that case, OOP loses even more of its thrust. The specification of the object does not follow from its class membership, but rather the class membership follows from the specification of the object. Identification of relevant relationship through referring to class definitions may be <u>tools</u> to find such relationships from which the design object specification is built. However, as mentioned in the preceding paragraph, this tool is only used pragmatically. In other words, rather than the class-subclass inheritance principle being the sole guiding one for converging towards a solution, it is <u>one of several means</u> to use in that process -  it is not  sufficient.

An additional problem with OOP is its counter-intuitive nature with respect to <u>operations</u> upon objects -  at least if the context is to  describe intelligent behavior as discussed under Sternberg in Chapter 2. In OOP every object holds knowledge regarding which methods and procedures may operate upon it. The enactment (or execution) of methods can only be accomplished if the object possesses 'knowledge ' that the specific method in question can access the object. A method that is not explicitly allowed this access is denied it. Therefore, there is a  built-in restriction on the components that can be selected for operations upon specific objects. This is a practical problem of three (related) reasons: (l) It restricts the search in the design process for innovative application of tools that existed, but were not conceived to use on the particular object, when that object was defined; (2) it restricts the object from being processed in other circumstances than what it was intended for at the outset; and (3) application of new tools demands that the object is explicitly defined to allow those tools to operate on it.

There is another problem of a more philosophical and  psychological nature. In Sternberg's and  most other psychologists view, and as is widespread among the AI community (see for instance Hayes-Roth, 1985; Coyne et. al., 1990; Davis, 1980), intelligent behavior is closely related to strict control of processes, where responsibility of actions is spread into a hierarchy or network of processes. In this view, each lower-level process is delegated control and executes still lower level processes to perform more concrete tasks. To assume that control over actions is only allowed if the objects that these actions are to affect are accessible does not seem very 'psychologically correct·. It would make it difficult to say that the processes control the outcome - formally it is the objects that filters the processes that control the outcome. Although this object-method view is useful and efficient in use in particular applications and contexts, there is a problem in basing a knowledge representation language on this philosophy of object-method

---

The problem that an object is prohibited from being processed in other circumstances than what it was intended for at the outset is particularly important in the context of *life-cycle product models*. The fundamental Idea behind this construct is that once a model is started to become developed, for instance in the design process, the knowledge gathered during development may be useful later on in the life of o product - the knowledge is to be recycled. For instance, a product model built in the design process should 'follow' the product into construction, manufacturing, operation, maintenance, and even dismantling, and be refined along its lifetime. If simulation is to be performed in the phase of operation, knowledge gained during the design process should be usable in the (simulation) model. To restrict the model to be subjected only to those manipulations explicitly thought of in the design process would seem to put heavy demands on the foresight of that modeler or programmer!

The issue of OOP and its problems in regards to life-cycle product models is not trivial. A concrete example from a Norwegian shipyard should be illustrative: The preliminary design process decomposed the vessel into traditional classes like hull, machinery, auxiliary equipment, and internal arrangement. In the detail design phase the decomposition was changed, into steel, piping, electrical systems, and woodwork. The construction was modular, in sections of 13 meters - a new decomposition. If a product model that were to follow into these three phases were rigidly based on the initial classification, it would have to be reconstructed for each new application. The number of classifications at this particular shipyard was actually moderate. Another Norwegian shipyard could after a thorough Investigation identify fourteen different systems of decomposition.

**Box3.iii**

---

relationships - it would simply restrict the discussion of control.

Another catch treated as objects, with classification according to common properties, decomposition, and subject to intervention from other methods (for instance through learning). (See Newell and Simon, 1972, for an excellent discussion of decomposability and classification of information processes.) The objects in OOP resemble in some is that methods cannot be objects, in the sense that objects operated upon by methods can be. There is a sense to which methods can semantically be way the *variables* and *object constants* in first order predicate calculus (what can be done to), and the methods resemble *function constants* (what can be done). In some sense OOP is an advanced alternative to first order predicate calculus, but, as predicate calculus, it becomes less potent in higher orders where also methods may be subjects of intervention, i.e., be treated as objects.

Concluding, it seems that OOP may be appropriate in certain design situations, particularly in any of the following:

(i)      *Top-down* approaches.
(ii)     When the D-object can be assumed to closely resemble existing objects.
(iii)    When all methods that are to be used in the process are *a priori* known.

In any of the following situations, OOP is not very efficient:

(i)      When the problem solving or design process is of such a nature that a
         concept is classified according to its properties, and not defined
         according to the properties of its class.
(ii)     When the structure (configuration) and property types are frequently
         changed (and subsequently also the class-membership of that object).
(iii)    When new methods are in development or when old methods are used
         in an innovative manner.
(iv)     When new classes are frequently defined.

In addition, in an environment where there is talk of control by an intelligent
mechanism, OOP introduces some significant problems in relation to the issue
of control.

## Process vs. Content -   Design Process vs. Design Object

Before moving on to remaining chapters it is necessary in this section to discuss
the distinction between what can be done and what can be done to. This is
a  segregation that has caused several problems; where does the boundary
between process and  object lie?

Traditionally, operations upon memory have been treated as separated from
the memory at which these operations have been directed. This is even seen
in research within psychology. Blumenthal notes (Blumenthal, 1977, page 161):

> "Modern investigators of memory include questions of the forms of
> storage of memory material, the separation of memory into different types
> or categories, and the distinction of different types of retrieval strategies.
> Earlier analyses of mind described ideas or memories as being associated,
> organized, stored, and retrieved. But when we ask who or what does this
> associating, organizing, storing, or retrieving, we are thrust back again to
> the organismic question of the *process* of cognition -  to questions about
> the user of the information rather than the information itself."

In traditional computer programming, there is a traditional segregation along the following lines: A combination of *programs* operate upon the *data* in the form of *variables, structures, files,* and so forth. The community has to a large extent taken for granted the distinction between *procedural knowledge,* or "implicit knowledge (which is stored) in the sequence of operations performed by the program," and *declarative knowledge,* which "is contained in declarations about the world. Such statements typically are stored in symbol structures that are accessed by the procedures that use this knowledge" (Genesereth and Nilsson, 1987, page 3)

Design researchers are also accustomed to the distinction between *content* and *process,* or more specifically design object and design process. Christiaans and Venselaar talk of four 'kinds of' knowledge - declarative, procedural, situational, and strategic. (Christiaans and Venselaar, 1987) Mistree et. al., as discussed in Chapter 2, distinguished the process model as one composed of *decisions* and *analytical relationships,* while the object description is reduced to one defining *systems* and *system variables.* The traditional interpretation of the popular term *product model* in the design context is a manifestation of the segregation of content and process in design, object and method in OOP terms, indicating a desire to construct a 'process free ' model of the design object that can be put to use by any conceivable process model, whether this is of the actual design process, or the later processes of manufacturing or operation.

In the context of artificial intelligence, McDermott systematically evaluates different approaches to creating a process-free model of the world, and concludes that "(in) most cases there is no way to develop a 'content theory' without a 'process model. ' ... A content theory is supposed to be a theory of what people know, how they 'carve ' up the world, what their 'ontology' is. A process model explains how they use this knowledge. A content theory is at Newell's 'knowledge level', supposedly independent of how the facts it expresses are to be manipulated. (McDermott, page 157, 1987) McDermott seems to give support to the approaches of, for instance, Mistree et. al. and Coyne et. al., and counter the view of, for instance, Yoshikawa. The prior researchers include in the models of the process also specific models of the information being processed. The latter separates the object space as something whose representation is grossly independent of the 'mapping' process that locates objects in this space.

The question is then what accounts for this interrelation between process model and content theory. One answer is that there obviously is a need for consistency between the two, in that the content must be accessible to the process in one way or the other. However, this only says that there must be a unifying language such that concepts in the content model are 'understood' in terms of the process model.[i]

Another answer might be that one cannot represent all the information that is available, since this would constitute an infinitely large content theory. Therefore, the model cannot support all conceivable processes. Of course, the mere use of the terms 'model' and 'theory' imply that these representations are selections or simplifications of the world - no models or theories are complete pictures of what they are to describe. Arguably, there will therefore always be processes that cannot be supported by the content theory simply for the reason that the knowledge needed for these processes to be undertaken is yet not included in the theory.

The above is not to say that new knowledge <u>cannot be</u> included in the content theory, or that 'new' or extended processes <u>cannot use</u> some of the knowledge that is represented in the 'old' content theory -  it is merely to say that the particular application of the theory -  the process model -  <u>was not complete</u> at the time the theory was established. Thus, the clue apparently is to use a language in which all content knowledge and all process knowledge <u>may be</u> included, and to use a language that enables content and process to be related. This makes possible a gradual expansion of both models without having to 'throw away' old representations. To rephrase McDermott: There is no way to develop a content theory without a process model unless the

---

<sub>h</sub>      When the terms *content* and *process* are used ln this thesis, it is not  to  be  interpreted as a subscription to the view that these indeed are segregative models. The terms are used to loosely separate between when something concerns activity and when it doesn't. It is argued earlier. and will be repeated later, that lt is difficult to draw sharp lines between that knowledge which pertains to process and  that which pertains to non-process.

content theory and the existing and future process models are represented, or rather representable, in the same language.

---

"It wouldn't be much use to have a theory in which knowledge is somehow stored away - without a corresponding theory of how later to put that knowledge at work. (Minsky, 1986, page 120)

**Box 3.iv**

---

In most researchers' view, as shown in Chapter 2, some processes run at a higher level than others - general problem-solving knowledge, represented for instance through Sternberg's seven components, is knowledge of a higher level than the procedures applied for search, calculation, symbol manipulation, and so forth. Newell and Simon stretched this even further by showing that processes can be decomposed into more elementary processes. (Newell and Simon, 1972) At the outset of the design process, or in the initial state, there does exist some process model that can 'cooperate' with the content theory. The point is that the process model gradually expands as component selection and ordering is undertaken, in Sternberg's theory, or as the processes are described in more detail, in Newell and Simon's theory. In other words, the process model develops as the process develops, ideally in such a way that the process causes the desired goal to be reached.

In this view, the content theory obviously supports the extension of the process model even if the extended process is not explicitly conceived or known at the time the content theory was established. Some (higher level) processes are represented initially in conjunction with the content theory, while other (lower level) processes are included in the process model as the problem-solving process evolves. " "Cognitive science ... aims to construct a ·working model' (initial process model and content theory) of a device for constructing working models (subsequent process models and content theory)." (Johnson-Laird, 1983, page 8, comments in brackets added here.) Similarly, the design process may essentially be viewed as the task of developing content theory and design process models.

This brings back the issue of natural language. Johnson-Laird takes the position that mental models do not have any procedural/declarative dimensions. He rather asserts that any mental model includes *procedural semantics,* where "... There appears to be no strong empirical consequences of claiming that a certain piece of information is stored in memory procedurally as opposed to declaratively. The choice of representation largely depends on what the system is supposed to do." (Johnson-Laird, page 248)

**Box3.v**

*grammar* plays a crucial role in making such a model 'meaningful'. "Mental models are constructed by a semantics that operates on superficial propositional representations of the premises. In general terms, this process consists in mapping strings of symbols into models. ... The intentions of a sentence can be built up compositionally from the intentions of its constituents in a way that depends only on their grammatical mode of combination." (Ibid., pp. 167-168) *Process* may then be seen as mainly an issue of applying knowledge through grammatical interpretation, while *content* is mainly an issue of storing knowledge as relationships among vocabulary elements in the language.

## Chapter summary

In this chapter I have undertaken a brief investigation of existing representational systems - representational languages - to see what they can offer in terms of representing knowledge about a design object and a design process. They all, in some form or another, use relations between concepts in order to represent knowledge. These representational languages are, however, too weak in expressive powers for the purpose of representing the knowledge that is relevant when the design process and the design object are to be described. This is particularly so when the ambition is to use one 'content theory' for several different purposes, some of which may not be known at the time when the theory is first 'started'.

I have argued that to make the content theory accessible to also new applications the content theory and the process model must be expressible in the same language.

In summary, I can say that no existing languages are completely appropriate for use in representing the design process because they either:

1)     Are not flexible enough in terms of process control.
2)     Are not broad enough to meet the particular needs in the design process.
3)     Do not give a language for developing content and process in the same framework

I claim that there is an underlying language which is more general and which is the ultimate 'source' of the artificial languages discussed in the preceding. I claim that this source is found among the natural languages.

# Chapter 4: Resolution through relations

The preceding chapter discussed a number of publications, both within the field of design and in the more fundamental field of psychology, all viewing the theme from different angles. The present chapter has two purposes: (1) To introduce design as using and constructing relationships in the process of forming a design concept, and (2) to prepare the grounds for discussing the Design Process Language developed in Chapter 5. The present chapter concludes with some axioms about what design may be considered to be, and some of these axioms will be used to lay out criteria by which the DPL should be judged and, more importantly, to specify what is to be expressed in the DPL.

It is important to recall that neither the DPL nor the axioms formulating design as a process of building relationships are algorithms for design. Nor is the aspiration to describe how design is or should be performed. It suggests a framework for a more integrated view of design, allowing the design process and the object of the process to be described in the same conceptual system. The purpose of the present chapter is thus to find invariants in design as seen from different perspectives, and the purpose of the Chapter 5 is to develop a language from the view of design as essentially a task of building and using relationships between concepts. DPL is to provide terminology and methodology that enables these relationships to be established.

This first part of this chapter discusses the general idea of knowledge as represented by relationships, or sentences in a language. Then it proceeds to discuss design as building such relationships and to discuss the premises for this process - the premises that condition the final result of the design activity. The chapter continues to broaden the perspective, arguing that the process of developing knowledge about the final result can essentially be represented in the same manner as the final result itself - i.e., as relationships between known concepts. Finally, some loose ends are collected and concludes with a set of axioms intended to capture invariants in design when viewed in the framework presented here.

| Term | Description |
|------|-------------|
| Token | Any string of characters |
| Symbol | A string of characters that denotes a distinct idea |
| Concept[1] | A reference to the meaning of a symbol |
| Primary concept | The aim of the concept formation activity - in a design-process equivalent to the D-object |
| D-object | The overall target of a design-process - the primary concept |
| Secondary concept | Concepts that are applied to attach meaning to the primary concept |
| Intrinsic relationship | A relationship between concepts ultimately related to the primary concepts that does not depend on the situation in which the primary concept is 'placed' |
| Extrinsic relationship | A relationship between concepts ultimately related to the primary concept that depends on the situation in which the primary concept is 'placed' |
| Generic memory | Memory of items of knowledge stored independently from the words that denote these items |
| Semantic memory[2] | Part of generic memory that concerns the meaning of words and concepts |
| Propositional memory[2] | Memory consisting of explicit relations between concepts |
| Design memory | Kind of propositional memory where all propositions constructed during the design task are stored |

Initially, it is necessary to distinguish *concept* from *relation*. This is done for the sake of convenience in discussion and will be modified at a later point. A concept may for now be viewed as something perceptible or concrete, while a relation may be seen to represent an abstract mapping between concepts. Thus, an *apple* and *the ground* are concepts, while *on* is a relation - we can observe both the *apple* and the *ground,* but we cannot observe the *on.*

Note that most of these terms are borrowed from psychology, but that the definitions do not capture the aspects required if this were a study in that area. They are used here, however, to ease the discussion and to provide a terminology that is easy to grasp.

**Table 4.1:** Some terminology and definitions used in this chapter

# Some fundamentals

Let me first focus on the issue of concept formation in a communication scenario. Assume that the world contains only four tokens - *the, ship, red,* and *is.* In addition, it contains the collection of tokens, or the sentence, "The ship is red." If the meaning of none of the tokens are known or if there are no means with which to interpret the sentence, the sentence itself carries no meaning. The sentence only carries meaning if two conditions are satisfied; at least some (possibly all but one) of the individual tokens must carry meaning (denote meaningful concepts) and there must be rules that say how they interact when placed together in a sentence.

Assume a situation where the persons A and B are communicating, where A possesses knowledge of the vessel and B doesn't. Three of the tokens denote, or symbolize, 'completely' defined phenomena to both A and B; *the, red* and *is. The* denotes a definite article, *red* denotes the value of a particular attribute (color), and *is* denotes a particular relation - the token *ship* does not yet denote anything to B. From the sentence that A transmits, B can then conclude that there is a kind of relationship between *the ship* and *red,* and that this relationship is symbolized by *is.* Immediately, there are three interpretations; either (1) *the ship* is identical to *red,* (2) *the ship* holds some property *red,* or (3) *the ship* is a kind of *red.* Which interpretation is chosen depends on the grammatical rules that apply. For the moment, assume that B chooses interpretation (2).

So what? It is now known to B that *the ship* is red, but this is all there is to it. If the task at this point is for B to simply categorize the token *ship* according to the <u>color</u> of the real phenomenon it denotes (an actual vessel), it may be sufficient knowledge. However, if the task for B is to enter into the real world and identify this particular *ship,* it is not sufficient (unless it is the only red thing in the world). B needs more knowledge that relates this ship to other known concepts, which is to say that <u>B needs to know more relationships between *ship* and known concepts.</u>

So, let me expand the universe of concepts known to both A and B to include *hull, machinery, propeller,* and so forth. In addition, both know another relation - the relation *have/has.* A could then communicate new sentences to B, iike "The ship has machinery." Assuming that the significance of numeric symbols were also known to both, A could establish that "The ship has 2 hulls," "The ship has 2 propellers." At some point, when A has constructed and transmitted enough relationships between *ship* and known concepts, it is possible for B to distinguish this particular ship (and possibly other ships) in the real world based upon the knowledge carried by the set of relationships (or by the concepts and the grammatical rules applicable to interpret the relations) that A has transmitted - B has formed a concept of 'ship '.

Of course, if A were to transmit a mere (unorganized) collection of the symbols introduced in those sentences this obviously wouldn't do the trick. Of importance to B is the way these symbols are combined, specifically that they are placed together in correspondence with a grammatical rule enabling B to interpret the sentences in the same manner as A The more sentences with which A relates *ship* to known concepts, the more defined the concept becomes to B. Thus, in order for A to define a new concept to B, both need (1) (elementary) concepts with which to define it, (2) a set of relations which is sufficient to develop the needed relationships, (3) sentences that combine known concepts and relations, and (4) (grammatical) rules which B can use to interpret the sentences correctly.[a] All of these must be common to both A and B, which is to say that they must use the same language and the same interpretation of that language (or at least of the segment of the language necessary for the communication to take place).

---

"Suppose that you chopped each picture into a heap of tiny pieces. It would be impossible to say which heap came from connected drawing and which came from disconnected drawings - simply because each heap would contain Identical assortments of picture fragments! ... It is therefore impossible to distinguish one heap from another by simply 'adding up the evidence, ' because all information about the *relations* between the bits of evidence has been lost· (Minsky, 1986, page 202)

**Box4.i**

---

So, if concepts are defined in terms of their relationships to other concepts; where does it all start? What is the point of reference? How do both A and B know what the symbol *red* signifies? The answer to the last question is probably: "Because both have experienced what phenomenon *red* denotes and learnt that *red* actually does denote it!" They assume an identical *frame of reference* for the concept *red.* The same probably applies to the relation *is.*

---

[a]     Where nothing else is mentioned, sentence and relationship will be used to denote the same phenomenon; a grammatically correct combination of relations and concepts.

Before continuing, let me state the following assumptions: A concept is either stored as *generic memory,* i.e., as a concept independent of the word that denotes the concept; as *semantic memory,* i.e., as a concept in generic memory with a word to denote it; or as *propositional memory,* i.e., with 'explicit' relationships to words symbolizing known concepts (see also Box 4.i, 4.ii). Put bluntly; either we need explicit sentences to attach meaning to a concept, or we don't. In the following, two assumptions are made for the sake of discussion; (1) a <u>known</u> concept is stored in propositional or semantic memory (there is a word for all concepts) and (2) definition of a <u>new</u> concept is undertaken as an enhancement of propositional memory (or through establishing new sentences relating the concept to other concepts.)

---

It should be realized that the issue of semantics and semantic/propositional memory is a difficult field undergoing extensive research, and too complex an issue to elaborate here. The reason behind the distinction between semantic and propositional memory is to distinguish between that which is implicitly known and that which must be represented explicitly. For instance, in most digital computers the binary digit 00001001 is related to the natural number 9 inherently, or as a consequence of the structure (hardware) of the computer, and may be interpreted as existing in semantic memory. C equais 9, on the other hand, would in most cases be represented explicitly - not as a consequence of the particular hardware, and may be interpreted as existing in propositional memory. Still in computer terminology, the mapping between 00001001 and 9 does not have to be represented symbolically, while the mapping between C and 9 does. See Dodd and White (1980) for a formal discussion of propositional/semantic memory and application of language.

---

With this in mind, return now to the sentence "The ship is red." Assume that the concept of 'ship' (or what characterizes a "ship") is now known to B as well as A.[b] Assume then that the task for A is to enable B to distinguish <u>this particular</u> *ship* from others. The 'redness' of the ship should be sufficiently established, so A pursues to say "The ship is a container ship," "The ship is a catamaran," and "The ship has name 'Anne Lise'," Depending upon the objective - whether B is to distinguish <u>this particular ship</u> from others (needs a specific definition, for instance by name) or is merely <u>generally interested in ships</u> (needs a general definition, for instance by kind) - A transmits sentences until she is certain that the concept of this particular *ship* is satisfactorily understood by B.

---

[b]    This actually means that if B spots an object she will know whether it is a vessel or not. (In other words, the listener accesses enough relationships to classify the object along the vessel/not vessel dimension.)

To make a preliminary conclusion in regards to design, recall the definition of an artifact in the design context: "A complete, procedural and declarative description, which when executed by an external agent results in the construction of an artifact." (Goel and Pirolli, 1989) In the present context, the designer must describe the artifact through so many relationships (sentences) that its (correct) construction is ensured.

Terming the concept that is the focus of the communication the *primary concept,* and the concepts used to define it *secondary concepts,* the following rather trivial statements may be made:

(1) The number and kinds of sentences communicated depends on how much about the primary concept (object of communication) the source perceives the target to know.

If most characteristics of the concept are known, the source needs only to transmit sentences distinguishing the current concept *(the ship)* from others, whereas the more undefined is the primary concept, the more sentences need be formulated to 'tighten· its definition.

(2) The number and kinds of sentences communicated depends on how much about the secondary concepts used in defining the primary concept are known by the target.

If the source utters a sentence like "The propellers have adjustable pitch" and the target does not know the meaning of 'adjustable pitch ', the prior might

---

Design is probably the only activity where the future is treated as present. Few would say "I am sleeping" before going to bed; or "I have a car" before buying it. Designers, on the other hand, invariably present something along the lines of 'The ship has length 125 meters' or "The ship has hull"· although the object in question does not exist. Do designers treat future as present? The conflict is clear if the relationship "the ship is red" is constructed and there are two 'possible worlds'; one in which the ship does not yet exist and the design process aims to design it, and another in which the ship exists and the design process aims to alter its color. In the prior case, one might get away with claiming that "The ship is red", whereas claiming this in the latter case might cause problems. Is it redat the outset, or is it to become red? (Presumably, one could use "The ship is red" and "The ship should/will/ought to/is to be green" to distinguish between present and future.) These questions will be dealt with later on in the chapter and in Chapter 5.

Box4.iii

have to define this concept through additional sentences like "'Adjustable pitch' means that the angle of the propeller blades can be varied." Thus, a concept may be defined through relationships to concepts that themselves are defined through relationships to still other concepts. In the end, some concepts are known through their relationship to the real world – purely semantic concepts.

(3)     The number and kinds of sentences communicated depends upon how the target is to  use the knowledge of the concepts being defined.

The kind and extent of information depends upon what is the  purpose with the communication. Obviously,  the need for information increases  with the number of uses to  which the information is to  be  put, and increases also with the depth needed -  how much definition a concept needs.

These three statements do not represent very radical thoughts, and  are indeed what researchers into speech activity have said for years; theme and context.

One of the fundamental premises stated in Chapter 1 was that designing essentially involves the production of knowledge. It then follows that designing is essentially viewed as a matter of producing a collection of relationships, and thus constitutes and enhancement of propositional memory, alternatively enhancement of what is here called *design memory* (see Table 4.1). (See also Aamodt,  page 149)

## Designing relationships, or relationships in design

The preceding section assumed a situation where the process was communication -  knowledge was to be transferred between a source A and a target B. A 'possessed ' knowledge about the ship prior to the commencement of the process, and  selected what knowledge to 'transmit' to enable B to  form a concept of this 'ship' - to <u>conceptualize</u> it.

Shifting the  situation somewhat, saying that also A is unknowing of the definition of the  primary concept, complicates matters. Assume that this situation is the initial state of a design process, where the purpose of the process  is to   define the primary concept through establishing enough relationships between it and secondary concepts. In the context of <u>design memory</u> the task is thus to Introduce relationships that previously didn't exist there - the construction of sentences is not to merely to map from memory that which Is known, but also to add to what Is known in memory.[c] So, before communicating to B the sentence "The vessel has a hull," A needs to determine that the vessel really does have a hull (or needs to construct the relationship) - A needs to <u>design</u> the vessel. It may be said that communicating a design object is to select relationships for transmission to another such that the design object is described adequately.

In the bypass It is worth mentioning that, strictly speaking, the sentence "The vessel has a hull" would be false until the vessel actually exists - the concept defined through any design process obviously does not exist prior to its definition and productlon.[d] However, where nothing else is noted, the form of presence refers to when the design is Implemented. The Issue of present and future is discussed in a later section of this chapter. See also Box 4.lli.

Let me return to the situation of the designer, the object of her *efforts* referred to as the *0-object.* All <u>relations</u> *(is, has,* and so forth) needed to establish the sentences describing the D-object are assumed known and the <u>secondary concepts</u> ('hull ', 'machinery', and so forth) needed to describe it are known - the task is thus to relate the D-object to these secondary concepts. Assume further that the designer has the <u>ability</u> to generate an appropriate description of the D-object (that is, she has components, tools, and knowledge of objective and constraints so that the D-object will be satisfactorily specified). The sentences constructed will then gradually describe the D-object until, upon termination of the process, the D-object is satisfactorily specified:

---

[c]   For now, assume that the memory in question is something that may be called *design memory,* where that knowledge which pertains to the design activity is explicitly represented - the design memory may be viewed as a notebook. This is to avoid dealing with the issue of whether the designer 'knows' something before remembering that she knows it! It is anyway necessary to assume an external or person-independent memory to exist in order to abstract the discussion away from the brain of the designer. and to discuss the activities of a human designer in the same language as design by an organization or by a computer.

[d]   There are exceptions to this. Designing a computer program is one activity where the result from the design process often exists when the process is finished.

"There is a  D-object"

"The D-object has a hull"

"The D-object has a machinery"

"The D-object has two propellers"

---

### A thought experiment

Assume that there are three  instances involved in the birth of a D-object; a Customer, a Designer, and a Builder. The Customer formulates expectations from the D-object, the Designer describes a D-object that can meet these expectations, and  the Builder constructs, or materializes, a physical object that conforms to this description. Let's see how the hypotheses-facts relationships turn out in this conception:

Source: Customer                    Target: Designer

Hypotheses: Implicit expectations -  desires and  needs                    Expect0tions
Process:      Problem description -  need assessment
Facts:        Explicit expectations -  Contract specification, objective Expectations met?

Source: Designer                    Target: Builder

Hypotheses: Explicit expectations -  Contract specification, objective  Expectations
Process:      Design
Facts:        Explicit specification, D-object                    Expectations met?

Source: Builder                    Target: Customer

Hypotheses: Explicit specification                    Expectations
Process:      Construction or implementation -  manufacturing
Facts:        Physical object                    Expectations  met?

In this scenario, the 'facts' resulting  from each 'process' become  'hypotheses' in the next one -  each process aims to transform the  description at hand to a result that best will fulfill the expectations stated in the description. As in this example, each phase in the design process might be  seen as processes that seeks to  transform hypotheses (expectations) into facts (specifications). The degree to which the process is successful depends on the degree to which the 'facts' fit the 'hypotheses'.

**Box 4.lv**

---

As in the communication scenario, and  what may not be too surprising, it is possible for the designer to represent this D-object as a set of sentences, where it gradually  becomes more defined as more sentences are constructed.  In some sense, the designer is in communication with herself.

Of course, the world isn't as simple as the example above might indicate. What is started is the first sketch of what is commonly referred to as a <u>system</u>, of which the 'hull', the 'machinery·, and the ·two propellers' are all considered <u>parts</u>. These parts are, however, not sufficiently defined concepts, so the designer needs to attach more meaning to them. The overall purpose is still to define the D-object, and the process of defining the secondary concepts may thus be viewed as a <u>means</u> to attach meaning to *D-object* - knowledge is to 'flux' to the D-object through these secondary concepts.[e]

So, the designer continues to create sentences to specify secondary concepts, in some sense <u>expanding</u> them:

> "There is a  D-object"
> "The D-object has a hull"
>> "The hull (of the D-object) has a frame"
>> "The hull (of the D-object) has plating"
>>> "The plating (of the body) has paint"
>>>> "The paint (of the plating) is red"
> "The D-object has a machinery"
> "The D-object has two propellers"
>> "The two propellers (of the D-object) are adjustable pitch propellers"

Now there is a  problem; the designer may initially exhibit <u>some</u> knowledge of the secondary concepts of 'hull', 'machinery', and 'propellers', (as 'known' in semantic memory) but this <u>semantic knowledge</u> is not  sufficient for this purpose -  while the concepts aren't entirely defined by the sentences above (which are placed in design memory), they aren't entirely known to the designer without these sentences, either. In fact, they are somewhere in-between being defined in semantic memory and design (which is propositional) memory, where some knowledge about the secondary concepts exists but  some doesn't prior to the design process. Let us for now call the existing (unstated) knowledge *implicit* knowledge and the new knowledge

---

[e]     Arguably, it wouldn't be necessary to apply these concepts. Had the designer chosen to do  this the tedious way, the D-object might (theoretically) have been defined as atoms related in particular ways. Thus, the  secondary concepts are indeed secondary!

| | |
|---|---|
| Hypothesis: | "The D-object floats" (Expectation: "The D-object should float'" |
| Fact: | "The D-object has maximum displacement equal 5000 tons"[1] |
| Fact: | "The D-object has maximum weight equal 4500 tons"[1] |

Mapping from hypothesis to D-object:

| | |
|---|---|
| Known relationship: | "If 'The D-object floats' (expectation) then 'Maximum displacement (D-object concept) is larger than maximum weight (D-object concept)'" |
| New relationships: | "If 'Maximum displacement Is larger than maximum weight· then '5000 tons (More concrete D-object concept) is larger than 4500 tons (D-object concept) " {Instantiating abstract concept with more concrete concept} and ' '5000 tons is larger than 4500 tons ' is "True". {Verifying against 'fact' base} |
| Proven hypothesis: | "The D-object floats"[2] |

Mapping from D-object to hypothesis:

| | |
|---|---|
| New relationships: | "If '5000 tons is larger than 4500 tons' then 'Maximum displacement Is larger than maximum weight'" and ' '5000 tons is larger than 4500 tons' is 'True·· {Verifying condition against 'fact' base} |
| Known relationship: | "If 'Maximum displacement is larger than maximum weight' then 'The D-object floats'" |
| Proven hypothesis: | "The D-object floats"[2] |

> This knowledge might be gathered by means of analysis. (It would probably be a consequence of more 'basic' knowledge, such as a geometrical model of the vessel.) Although the statements probably wouldn't be true, since they deal with expectations, they are treated as 'facts' in relation to the concrete hypothesis.

> The 'proof' is more the result of a preliminary test based upon forms of analysis (for instance. induction and deduction, respectively). The facts used for the ultimate test can only be gathered when the D-object is constructed, in that future which the hypothesis concerns.

**Box 4.v**

*explicit* knowledge.

So. the definitions of the secondary concepts above are based both on implicit and explicit knowledge, and the D-object is in the end defined entirely through explicit knowledge.[1] In other words, all knowledge about the D-object

---

f    One may interject that the D-object may partly be known semantically as well. For instance, If the D-object is a car there might be properties associated with the concept 'car' that are generally known without having to be explicated. However, in this case 'car· is not a primary concept! The D-object is a 'reserved· term to designate something that is unknown to the world before the commencement of the design process.

is stored as explicit relationships to other, secondary concepts. These secondary concepts may in turn be defined both in semantic memory and in design memory, or both through implicit and explicit relationships. The implicit/explicit distinction is elaborated further towards the end of the chapter.

Note that those secondary concepts that must be related to other secondary concepts by means of *has/is* relations to become completely known are not considered sufficiently defined in semantic memory, whereas concepts at the 'end' of a *has/is* sequence may be. It seems that the issue of concept definition can be linked closely to the notion of <u>control</u> in the grammar. In the case above the relation *has* directs control, where the target concept (to the right side) has less control than the source concept (to the left) - the target is 'possessed ' by the source. Suffices it to say here that the further 'to the right' in a sequence of has-relationships, the more fundamental or concrete is the concept.[9]

Said in another way, since the concept 'paint' normally (at least in this case) is not distinct enough, it must be described further. The concept 'red ', however, is assumed known - few designers would continue specifying what 'red ' means, unless they were intending to qualify it with nuances, brightness, and so forth. In fact, the designer in this case treats this particular concept as being <u>generally known</u>, or something that both she and others could distinguish simply by name - its definition is complete in <u>semantic memory.</u> The same is the case with the relations *(is, has)* whose meaning can be assumed given through general, grammatical rules. So, one might preliminarily hypothesize that the designer seeks to create sentences which relate items (concepts) that are under-defined to items (concepts) that are satisfactorily defined. Of course, if it were <u>implicit</u> that red paint was the only kind of paint existing, the sentence <u>explicating</u> that "The paint (of the plating) has color red" would probably be unnecessary, and the concept 'paint' would be over-defined (or its definition would be redundant) or superfluous.

---

[g] This seems to be a pattern when transitive verb relations are used to relate two concepts.

## Quality assurance and concept formation

Consider what happens in a design process where it is of ultimate importance to avoid misunderstandings of what the design result should look like when implemented, like is the case in the North Sea offshore industry or in the aerospace industries. One would expect even elementary concepts to be specified - little would be taken for granted. This in fact means that more secondary concepts are described through relationships, reducing the risk of misinterpretations.

It might be expected that as the cost of misinterpretation increases, the number of concepts introduced in the description of the D-object increases. If the cost of applying a wrong engine type increases, actions would likely be taken to define the engine in more detail, i.e., relating It to more secondary concepts. As a matter of fact, the detail design phase may be viewed to be an effort to specify secondary concepts that have remained under-defiAed after the completion of the preliminary design phase.

> 'Maybe it's a weakness of human designers that we leave up to chance to formulate concepts that we ourselves should have defined in order to get a result that corresponds to our expectations?' (Stian Erichsen, personal communication, 1993)

**Box 4.vi**

# Direction in designing relationships

At this point it is useful to stop for a moment and adjust the scenario before continuing to discuss design and relationships. Assume that all there is is a language with which to describe an object. What would happen if the designer choses secondary concepts and relations at random, puts them together in sentences, and tests the result after its implementation by observing which 'survive' and which don't? Assume that there is an environment - the World - of which the designer initially doesn't know anything. The designer would probably experience a catastrophe! Although the probability of acceptance is greater than zero, the undirected construction of sentences would most certainly lead to failures. Fortunately, this is as far away from human design as Is possible to come. As most researchers acknowledge, human design as opposed to Nature's design is a directed activity- a human designer exhibits purposeful behavlor.[h]

---

It is noted that Nature may have a purpose with her design, but it would hardly be called 'design'· in the same manner as we talk of design by humans.

So, the question arises: "By what is design directed?"   In general terms, one answer is that it is directed by what the designer wants her D-object to become in the real world, or what purpose she perceives the D-object to serve. It is also directed by the resources at disposal to the designer in both designing and implementing the result, the limitations and constraints put upon the solution and the process, personal preferences (internal value systems), the state of the environment at large in which the D-object is to be placed, and so forth. However, let me first focus on the purpose with the D-object once implemented.

Glynn says of designers that they "... *move* from the relatively disharmonious and unmanageable elements that constitute the problem, to the hypothesizing of solutions that aim to render them harmonious and manageable." (Glynn, 1985, page 123) In this view, design is similar to the process of generating scientific hypotheses. The final, that is, implemented and constructed, result can then be viewed as the fact in relation to which the design specification (set of hypotheses) is tested.

It is useful, in this analogy, to discuss in some detail the process of moving from problem to the solution. First cast the purpose as formulated by a set of sentences describing a desired future, or expectations from the D-object. Terming the sentences as hypotheses about the future, the scientific approach demands that these hypotheses are testable, and their verification or falsification results from the outcome of the tests.

Ideally, these tests are performed by observing the facts of the future which the hypotheses describe and see if these facts fit the hypotheses. In a design situation, the facts directly related to the D-object are under the designer's control, so the designer may indeed specify them -  that is, she has the luxury of designing facts that appear to best withstand the testing of the hypotheses. The description of the D-object then is a set of sentences, the implementation of which will be the expected best 'fit' to the hypotheses. Thus, a hypothesis {purpose} like "The D-object has {is to have} a capacity of 3500 containers" is obviously verified if the resulting D-object is shown to have a capacity of 3500 containers. However, as mentioned in Chapter 2, the designer can rarely implement or construct the D-object to perform the test - she must describe the D-object and treat this description and the deductions made from them as if this knowledge constitutes facts.

---

"To state a problem is to designate (1) a *test* for a class of symbol structures (solutions to the problem), and (2) a *generator* of symbol structures (potential solutions). To solve a problem is to generate a structure, using (2), that satisfies the test of (I)'. (Newell and Simon, 1975, page 301)

**Box 4.vii**

---

A problem arises to the designer now, since the hypotheses normally are statements concerned with how the D-object <u>functions</u> in the future, or the <u>role</u> it will play - not how it is <u>shaped</u> (or how to meet demands regarding function). The latter is, of course, what the designer is to describe.' Therefore, in order to <u>verify</u> the design of a D-object the designer needs to either (i) interpret its (structural) specification in terms of the concepts used in the hypotheses {purpose, or expectations}, (ii) to interpret the hypotheses in terms of the concepts used in the (structural) specification describing the D-object, or (iii) both. Either way, there must be some means to relate the D-object specification to the hypotheses - that is, the designer must construct sentences describing these relationships. In some sense, this is analogous to what was earlier referred to as "extensive performance modelling" - it is a means to assess to what degree the resulting object will satisfy requirements.

Thus, it is not sufficient for a language to enable the construction of sentences relating the D-object to its individual make-up (D-object specification). It is necessary for this language also to enable the construction of sentences relating the specification of the D-object to the expectations put upon it, or <u>the premises of</u> the D-object (the hypotheses), This can be performed in two ways. Either the designer (I) ·translates' the 'hypotheses· to 'facts', which is to say that the (functional) goal statement is transformed into a (structural) description (point (i) above), or (2) translates the 'facts' into 'hypotheses', which is to say that structure is transformed into function (point (ii) above). In

---

Some researchers claim this to be the essence in design; to move from a functional or abstract description to a technical or concrete description. The design process thus becomes a mapping between a function space and an object space, or between a problem space and a solution space (see earlier discussion on Goel and Pirolli and Yoshikawa, e.g.).

any case, there must be some means of translating in order for the designer to 'prove ' the 'hypotheses·. An illustration of such a mapping situation is given in Box **4.vii.**

This simple example should serve to illustrate roughly the idea behind the mapping between a D-object and what may be called the <u>premises for its existence,</u> i.e., the hypotheses towards which it will be tested. The idea that design can be viewed as generating facts to match hypotheses indicate that a 'design language· must make possible the derivation of a specification and a later test of this specification against the formulated purpose. It should be reasonably clear that the mapping idea is valid also for other kinds of relationships. If the problem is to <u>relate the D-object to its environment,</u> sentences must be built to map between concepts describing the D-object and the environment. So also for other such relationships (for instance, between the D-object and the system of values used in determining its utility).

As a final example on the relationship between the knowledge of the D-object and other 'kinds of' knowledge, consider Simon's definition of an <u>artifact:</u> "An artifact can be thought of as a meeting point - an 'artifact' in today's terms - between an 'inner' environment, the substance and organization of the artifact itself, and an 'outer' environment, the surroundings in which it operates. If the inner environment is appropriate to the outer environment, or vice versa, the artifact will serve its intended purpose." (1981, page 9) This implies that the D-object has two 'lives' - one strictly internal and one depending on external circumstances, one of <u>internal composition and configuration</u> and one of <u>external role and function.</u> The characteristics of the object that are invariant with the state of the outer environment are here termed *intrinsic* concepts, while characteristics that varies with this state are termed *extrinsic* concepts. The latter characteristics would, in Simon's terms, seem to describe the artifact - the meeting point between the internal and the external environment.

As a matter of fact, the artifact may be viewed as all the extrinsic concepts of the D-object. To indeed map between the D-object and the circumstances, there must be relationships between the extrinsic and the intrinsic concepts of the D-object, as well as between the extrinsic concepts and concepts describing the circumstances. The artifact is thus a set of concepts that are defined through a <u>set of relationships</u> both to intrinsic concepts of the D-object and concepts describing the circumstances.

---

• "(In problem solving) the cognitive system attempts to find a sequence of actions that will transform the initial problematic state into a goal-satisfying state. An adequate mental model can accomplish the task by mimicking the environment up to an acceptable level of approximation, The model need only describe aspects of the environment and of the (cognitive) system's action that are relevant to the attainment of the goal-satisfying states." (Holland et. al., page 39)

**Box4.ix**

---

Assume now that the goal for the designer is to represent the inner environment separately from its outer environment. The inner environment consists of the concepts describing the D-object that remain unaffected by the 'operating conditions' whereas the artifact represents the behavior that depends on these operating conditions. There must therefore necessarily be a means to map between the outer environment and the inner environment to assess behavior, i.e., a means to construct relationships relating concepts 'belonging to' the inner environment to concepts 'belonging to' the outer environment. In other words, there must be a way to describe the artifact.

The above introduces the problem of how the designer knows where the inner environment (what will be termed the D-object) ends and the outer environment (what will be termed the environment) starts. One way could be for the designer to pose the question: "What relationships between concepts are constant regardless of the situation in which the D-object is placed?" Obviously, the composition of material is constant - steel is steel and plastic is plastic wherever it is placed, and a welded joint is welded whether it is placed on Earth or on the moon. Likewise, a beam is connected to the same strut regardless of the location of the structure - material composition preserves the "autonomy of the artifact."

The grand problem, however, arises when the designer is to determine what really are 'deduced ' concepts, that is, those which are not *intrinsic* to the D-object. The <u>speed</u> of a vessel is certainly dependent on the environment in which it operates and is therefore a non-intrinsic concept. Strictly speaking, the <u>weight</u> of the same vessel may also be viewed to be defined by a relationship

"The solution is that we need to combine at least two different kinds of descriptions. On one side, we need structural descriptions for recognizing chairs when we see them. On the other side we need a functional description in order to know what we can *do* with chairs. But it's not enough merely to propose a vague association, because in order for it to have some use, we need more intimate details about *how* those parts actually help a person to sit. To catch the proper meaning, we need connections between parts of the chair structure and the requirements of the human body that those parts are supposed to serve."(Minsky, 1986, page 123) between the inner and the outer environment (mass and gravity, respectively) and thus to be an extrinsic concept. What then about the mass? It would most certainly be treated as 'belonging to· the D-object, but if the argument above is stretched to its limits it may also be seen to be an extrinsic concept - it depends on the speed with which the D-object travels.[i] Of course, it makes little sense to pursue this to the extreme and there must therefore be a judgement made as to what reasonable variations in environment (what 'scenarios') are expected, and accordingly segregate the interior of the D-object from the circumstances - the concepts representing variations in the environment would then constitute the concepts to which it should be possible to relate the D-object (or rather the secondary concepts to which the D-object is related). The relationships between the object and the environment can then be called the artifact.

**Box 4.x**

## Applying relationships to build relationships

So far, the discussion has focused on sentences describing a D-object, and the relationships between the D-object and the premises for its existence. Little has been said of the manner in which these relationships are constructed, which is of fundamental importance in design. Hitherto the design process has been referred to as sequences of interventions and transactions that manipulate knowledge - interventions and transactions that manipulate sentences relating the concept *D-object to* other concepts. The purpose in the Design Process Language (DPL) is to allow these interventions and transactions to be discussed

---

m=mv=0/(l-v2/c2), according to Einstein's theory of relativity.

in the same manner, or by the same language, as the D- object and other knowledge. This section explores this possibility.

But first we need to find out what these interventions and transactions are. In the context of literature discussed in Chapter 2, they are what Goel and Pirolli and Newell and Simon call <u>information processes</u>. They are what Sternberg calls <u>components</u> and what Mistree et. al. call <u>activities,</u> primarily symbols processing and decision making. Yoshikawa terms them <u>mappings</u> while Coyne and Gero cast them as <u>operations</u> within a language (alternatively <u>actions).</u> One common factor is that they in sum represent the <u>active aspect in the </u>design    process – active in their manipulation of the information/knowledge base.

The interventions and transactions may also <u>themselves be</u> <u>viewed</u> <u>as knowledge,</u> as several have explicitly stated. (Guilford, 1967; Sternberg, 1985a and 1985b; Cristiaans and Venselaar, 1987) Another conclusion to draw from literature is that they may be conceived as <u>distinct entities</u>. Mistree et. al. distinguish them in the form of various classes of <u>activities,</u> Coyne et. al. view them as <u>specific rules</u>, while Sternberg and Newell and Simon consider them as particular <u>processes.</u> Yoshikawa implies that they can be identified as kinds of <u>"search and evaluate' activities,</u> Goel and Pirolli do not distinguish specific processes, but observe that whatever they are they appear to characterize design behavior. In short, there is support of the idea of discussing them as distinct concepts.

Finally, it may be concluded that <u>the activities are in some way related</u>. Some interventions and transactions must *come before* others (in sequential processing), some *depend on* how others are accomplished (component selection before component sequencing), some will arguably be *better than* others (according to some measures), while some are *parts of* other higher-level interventions and transactions (elementary information processes vs. composite, as Newell and Simon discuss them.)

.  This corresponds to the view described in the earlier sections; <u>the process may be viewed as sentences where concepts are combined through relations, and where</u> <u>new</u> <u>sentences, or relationships</u>, <u>are</u> <u>viewed</u> <u>as knowledge.</u> Two questions then arise: (l) Are concepts, relations, and grammatical rules for Let me try to illustrate what the design process may look like in real life and some characteristics that distinguishes it from natural evolution. Picture a river flowing, where the initial condition Is water present at a location above the sea level, and the implicit purpose for Nature is to transport this water down to the sea level. Nature intends to "change an existing situation into a preferred one· The solution favored is to choose the path of minimum resistance - in fact the chief (or only) priority is to get the water down to the sea in the cheapest possible way. The time it takes to reach there is unimportant, as is the choice of scenery en route. In the course of action, some implicit constraints are experienced, for instance disallowing the water to toke a shortcut over a hill and forcing it to travel around that hill instead, and disallowing tt to take a break to catch its breath at arbitrary locations. When the water reaches the sea Nature 'knows' its task is done without any 'stopping rule.' In fact, Nature designs without thinking, and it can do so because it has perfect 'knowledge' of goal,, constraints and pri0rities, and of the environment and the functional relationships outlining the most optimal solution (according to the priorities).

Picture the designer out to do the same job: First she would need to map the expected flow of water, including daily variations, and account for some 20-, 50- or 100-year flood condition. After this, she would start to mop the terrain along several possible routes to the sea and determine possible outlets. Then she would start to evaluate what consequences an outlet of water would have on fisheries, fog-conditions, the nearby aquaculture facility, and so on. She would proceed to screen out alternatives until she were left with a set to pursue. In the process, she has received a specification from her employer telling here that the water cannot toke more than two days to travel, and that it be used for generation of electricity at three locations along the way. In addition, every liter is to be used as drinking water by at least two persons before it arrives at the sea. The project cost has a roof of USD 1.5 billion, and the construction period cannot last for more than seven years. Our design engineer is in a bind. There are constraints all over the place, both imposed by nature, the employer, the communities, local politicians, and conservationist groups. There are multiple objectives, including the tangible ones like relocation of water, production of drinking water and of electricity, and the more intangible like preservation of the environment. She doesn't know very well how the running water in time will Influence climate and how it will erode the ground, nor does she know with certainty how much water can be expected. To investigate this, she needs to locate or make some method to assess these effects. In addition, the alternative solutions are numerous, and there Is a large number of decisions to make. In short, she must intervene constantly throughout the design process.

This Is to illustrate that the conscious design of an engineering artifact implies several interventions from the designer, which in natural evolution would be superfluous. It is these interventions that we seek to facilitate in the work on studying, explaining, or designing the design process. It is also the artificial relationships between phenomena that the designer Is to investigate and describe in the course of action.

**Box 4.viii**

describing the design process indeed so unique that a separate language is needed? (2) if not, is it useful to represent these "process" concepts and relations in the same language as the "object" concepts and relations?

The tricky issue of distinguishing object and process was briefly discussed towards the end of Chapter 3. In that section, there was a process model operating on a content theory. One way to interpret these two constructs is that the former may intervene upon the latter and change it while the opposite is not true. Let me, for the sake of discussion, refer to the process model as <u>active knowledge</u> (ability to perform) and the content theory as <u>passive knowledge</u> (no ability to perform) without stating whether they are separate or not or presenting formal definitions of the respective at this point.

---

Of course, the process model must be allowed to intervene upon Itself or else there must be something else that con intervene upon the process model. This must be so since a fixed and detailed strategy constructed once and for all at the outset of the design process is not congruent with most views of design as rather ill-structured problems and of the designer as an adaptable mechanism. The designer must change the process model continuously.

**Box 4.xi**

---

Let me then answer the last question stated above first; Is it useful to represent passive and active knowledge in the same language? Is it useful to describe a design object in the same language, using the same words, using the same grammatical rules, as the design process?

In the context of Coyne et. al. discussed in Chapter 2, the <u>language of form</u> (passive) is operated upon by the <u>language of actions</u> (active). In other words, there will necessarily have to exist relationships between the passive and the active knowledge. If we were to use <u>different languages</u> for the passive (or D-object and others) and the active (process) knowledge, it would seem that we needed to establish a third <u>meta-level language</u> in order to relate concepts in the two languages to each other - the mapping would have to be performed at a level 'above' both object language and process language.

There are several reasons for the relationships indicated above to be represented explicitly. One is that the choice of transactions and interventions in the design process depends upon what is known, and what is required to be known, about the D-object. This may be called <u>inputs of passive knowledge</u> to the design process. Another is that <u>the D-object results from the execution of the process</u> - the design process <u>generates output.</u> If we do not to represent the design process and the D-object in the same language, we necessarily need to represent the process separately from what the process operates upon. At the very least, this seems quite inconvenient in most cases. The increasingly popular work on developing generic product models will be hampered by this problem -  how is it possible to make models that can be set to all possible uses without having a generic process language that can access this model? The artificial intelligence community has struggled with it for years - how is it possible to construct intelligence mechanisms without "the intelligence knowing what to be intelligent about?" So, it may be expected that there indeed are great advantages of having a single language in which both the passive (object) and the active (process) aspects of design can be described.

Then the first question stated above remains: Is it possible to do this? Let me turn the question around, and ask: Why shouldn't it be possible?

The most difficult problem lies in the active/passive distinction. Intuitively there is a difference in nature between the two 'kinds of' knowledge. In linguistics, this difference is traditionally captured in distinguishing among transitive and intransitive verbs, or verbal relations, while the artificial intelligence community distinguishes between declarative and procedural knowledge.

In the linguistic approach, is is a non-transitive verbal relation, while writes is a transitive verbal relation. In natural language the rules regulating these relations are captured in the same system. So, the difference between the sentence "The ship has a hull" and "The designer writes a specification" lies essentially in the kind of verbal relation used. By knowing the meaning of the relations it is possible to distinguish knowledge of passive from knowledge of active.

Other kinds of relations can be classified as well. Consider "The ship is at the dock" and "The procedure X executes before the procedure Y." These relations belong to the same family -  the prepositional relations -  and have as such common characteristics. The same applies to conjunctive relations; "The designer has a computer and a calculator" and "The designer executes procedure X and procedure Y" -  the same general rules apply to interpret the sentences. Another example makes the connection in the language even stronger. Assume that the procedure X consists of two parts, 'read' and 'write'. This may be described as follows:

"X *has* a read-procedure"
"X *has* a write-procedure"

Compared to the example on page 75, the resemblance ought to be clear. In that case, the relation *has* was used in describing parts of an object, whereas in the sentences above it was used to describe 'parts' of a process. Process description has so many similarities to object description that it should be possible to establish a common language based upon natural languages.[k] The linguistic approach will be taken in developing the Design Process Language in the next chapter.

It is useful now to summarize by means of an example. Assume the role of an outside observer that is to describe everything (all transactions and interventions) that a designer performs in a design process to a listener not observing the process. As in the first section of this chapter, there is a communication situation. However, the context of the communication is now a design process as opposed to a design object. So, the moment the designer finds that "The ship *has* a hull," the observer could say something like "The designer *finds that* 'The ship has a hull'." If the designer makes note of this new relationship, the observer could say something to the extent that "The designer *writes down that* 'The ship has a hull'." It is characteristic of the comments of the observer that those sentences describing interventions and transactions can be described as a relationship between the designer and that at which the interventions and transactions are directed - in the examples above, the sentence "The ship has a hull." Indeed, every transaction that the designer performs can be described by the observer through using some verbal relation (and possibly other relations as well). Like the D-object was described as sentences relating it to known concepts, the design process can be described as sentences relating the designer to these sentences.

---

[k]       What is said thus far may be radical in terms of the traditional view of design -   i.e., process and object viewed as separate concepts. In natural language their relation is self-evident -   while there exist rules allowing for distinguishing active from passive these rules are different aspects of the same language.

# A note on knowledge and meaning

Earlier in the chapter it was distinguished between explicit and implicit iknowledge. For instance, "red" has a high implicit content to most whereas "girk" does not. This has to do with knowledge and meaning. Assume that the designer is a closed system, with no interface to the external world. Thus, the designer does not know the significance of the symbols being processed beyond their significance to other symbols - there is no implicit knowledge.' Assume further that the system is complete, in that all relationships needed to construct the required relationships are present - the designer can relate the symbols in a way such that the hypothesis is proven - such that the expectations are met.

In this system, it is possible to use any notation as long as the system remains complete - the symbols 'length ' and 'beam ' can be switched as long as all appearances of the two symbols are switched (for instance, "Length must be greater than beam" is replaced with "Beam must be greater than length"). The designer in this closed system can perform the design task and produce a consistent design that meets the expectations without knowing the 'meaning' of the symbols outside that system. The symbol 'length' is just any arbitrary symbol that is related to another symbol, like in "Length equals 120," as a consequence of the existing relationships. (For instance, "Length equals (displacement divided by (beam times draught times block coefficient)).") Why would it cause problems to switch the symbols 'length ' and 'beam ' as long as the complete design system is revised to make it consistent with this new terminology? Why isn't it sufficient to have internally consistent knowledge in the design system?

The answer obviously lies in the fact that the specification the designer produces is to be used in later processes. Some other system is to interpret the specification in terms of what the symbols mean in its own *frame of reference.* Just as any drawing that is to be used for construction purposes are affixed

---

Assume that the relations are known, such that the implication of a symbol having a relation to another symbol, like in "length equals 120," is known as far as symbolic processing is concerned. (For instance, the system then 'knows· that the symbol 'length· can be replaced with the symbol '120· in a formula where 'length ' appears.)

scales to relate the 'internal ' scale that the designer used to a scale known to the constructor, so must all symbols that bear significance to the constructor be understood by that constructor. In the end, the recipient of the specification must either possess or form a concept of what the symbol 'length' means - it must eventually be defined in terms of his or her implicit concepts. In this particular case, 'length ' must probably be related to a unit to make the symbol 'length' understood in terms of the concept designated through the symbol 'meter' (while 'length' to most is uniquely known to be a measure of dimension, it is not uniquely known what kind of dimension). Through such a process does 'length ' become a meaningful concept also outside the design system.

This illustrates the distinction between implicit and explicit knowledge - between the knowledge that is represented as relations among symbols and the one that exists between a symbol and its 'real' designation. The symbol 'meter· is a (semantic) concept because to most it carries meaning beyond itself, whereas the symbol 'length ' carries meaning when the design system is opened to allow for it to be related to concepts known outside that system, or semantic concepts.

This is not a trivial distinction. If the designer chose to specify the design object in a language not understood by the one that were to construct the system, there would be no way to secure that the artifact indeed would become what the designer intended it to be. If the designer then states that "A weld is to connect the web and the flange" and leaves the symbol ·weld· further undefined this could imply one of two things: Either (i) any 'weld· would do, implying that the general definition (for instance, the 'class-description') of weld is sufficient, or (ii) the constructor is expected to develop further the concept of 'weld' by relating it to 'height', 'length', 'fill angle', and so forth. While some aspects of this 'weld· are known to both the designer and the constructor, which is to say that both share some concept of 'weld ', there may be some aspects that are only known by the designer, and thus have to be explicated by her. This might, for instance, be the actual values of the respective attributes. Some is *implicit* knowledge (that a weld *has* a 'length') and some is *explicit* (what this 'length' actually *is,* both in value, unit, and measured from and to).

Consider the two cases above. In case (i) the situation is fairly simple to the constructor - as long as anything that can be called a 'weld' connects the 'web' and the 'flange', everything is fine. However, it is sufficient that there is only one weld that won't do to make the concept 'weld' under-defined - "My Kingdom for a horse" is a bad bargain if the 'horse· isn't alive! In case (ii) the designer expects the constructor to form the concept 'weld ' to suit the purposes - there is an expectation that the constructor will specify the concept further in an 'appropriate' manner - "Well, I guess when he says 'a horse' he presumably means a horse that is alive and well!?"

A designer can often assume (i) to apply to, for instance, concepts like 'meter' (can hardly be misunderstood), numeric symbols and unique names.[m] These are concepts that need no further definition. She would assume case (ii) in other design situations, where she would expect the target to understand or deduce the true meaning of the concept. To aid in this process, however, she could make one important modification; just as the designer works to fulfill expectations put to the design object, she might herself state expectations from, or functions of, 'weld · and assume the target to meet these expectations. For instance, the designer could describe 'weld ' through a relationship like (a) "The weld should resist forces of ..." instead of describing it through relationships like (b) "The dimensions of the weld are ...," where (b) is a combination of concepts that are expected to satisfy (a). Few designers would specify every tiny part of the design object at one go - they would probably tend to define some concepts through such expectations as in (a) and trust that the next stage, the target, ensures these expectations to be met by developing specifications as in (b). Indeed, the designer both proves hypotheses in the design process (combines concepts to meet expectations) and generates hypotheses (states expectations that the next stage is supposed to meet). This is all part of the concept formation process.

In plain terms, this amounts to a design process that could be described through the following steps:

---

[m] Most engineers who have functioned in both the metric (SI) system and the American system of notation would know the problem of confused concepts. Consider the use of the concept *tons* (SI metric, 1000 kg.; US short, 2000 lbs.; and long, 1016 kg.) and the American *lbs* (weight) and *lbs* (mass). Even in one system it becomes difficult - does *gross tonnage* refer to *register tonnage* (volume) or to *tons* (weight)?

(1)    The designer has a set of initial statements of functional expectations.[n]

(2)    The designer starts to develop a structural description that can either (i) completely meet some of the expectations or (ii) contribute towards meeting them. That is, she starts to develop a structural description that, when completed, can help prove all expectations to be met.

(3)    As aspects of the structure are described, the designer states new functional expectations that follow (are translated) from the initial expectations, but which are adapted to the particular kind of structure in development.

(4)    The new functional expectations are resolved through refining the structural description such that the new expectations can be provably met, and possibly through generating new functional expectations that are adapted to the new structure.

(5)    At some point, when the process through steps (1) through (4) has evolved such that the structural description and the functional expectations are sufficiently detailed, the designer concludes that (i) the structural description at her stage and (ii) the remaining functional expectations are sufficient, where 'sufficient· means (A) that the initial expectations are provable from the structural description and the functional expectations and (B) that the recipient of the design will understand the design sufficiently well such that the materialized result will prove the initial expectations to be met as in (6).

(6)    The designer delivers a description of the D-object which is a mixture of functional expectations (at a low level) and a structural description which is such that they together, if true in the materialized result, will cause the initial functional expectations in (1) to be met.

The ultimate result from the design process is then a mixture of 'facts· and 'expectations·, where the facts could be a description of the materialized structure and the expectations could be a description of what the elements of that materialized structure are to perform. An outline specification of a ship is a good example of this.

---

[n]    Throughout the process It may be that these expectations are revised as new information of opportunities or constraints is gathered. However, in the proceeding assume that the expectations do not change, that is, the overall purpose with the D-object remains constant.

## A note on function and structure

As argued above, the specification generated by the designer is composed of sentences belonging to, roughly speaking, one of two main categories: (1) Those used in the *structural description* of the concept, such as dimensions and 'possession' of specific parts, and (2) those used in stating *functional expectations,* or the roles that these parts are to fill. The distinction is frequently used in design and is a useful one except that it is somewhat elusive, as will be shown below.

Assume that the concept is a <u>part</u> of a <u>system</u> - a concrete and tangible concept. In this case, it could be expected that the concept had to <u>facets;</u> one functional and one structural. The description of the <u>functional facet</u> is then one that preconditions the description of the <u>structural facet.</u> The concept 'hull', for instance, has some description according to dimensions, as well as several other characteristics. Assume that 'hull' is <u>structurally </u>described as follows:

1. "Vessel has hull"
2.      "Hull of vessel has length"
3.           "Length of hull equals 140 meters"
4.      "Hull of vessel has beam"
5.           "Beam of hull equals 30 meters"
6.      "Hull of vessel has draught"
7.           "Draught of hull equals 14 meters"
8.      "Hull of vessel has block coefficient"
9.           "Block coefficient of hull equals 0.67"
10.     "Hull of vessel has volume displacement"
11.          "Volume displacement of hull equals 39,396 meters$^{3}$" {deduced}
12.     "Hull of vessel has weight displacement"
13.          "Weight displacement of hull equals 40,538 tons" {deduced}
14.     "Hull of vessel has weight"
15.          "Weight of hull is 4,000 tons"
16.     "Hull of vessel has net buoyancy of 36,538 tons"
17.     "Hull of vessel is leak-proof"

All the relationships above are supportive of the <u>functional</u> relationship:

18a.    "Hull of vessel should have net buoyancy larger than 36,000 tons" or
18b.    "Hull of vessel should ensure that the weights are supported" and
19.     "Weight of vessel excluding hull equals 36,000 tons"

The elusiveness in the division into structure/function is particularly clear in sentence 17. In order to satisfy that function (of 'leakproofness') one could later produce a functional description as in "All plate joins are impermeable" or a structural description as in "All plate joins are made of impermeable material." One could thus expect that a structural description at one level (for one concept) becomes a functional description at another level (for another concept). For instance, "All penetrations of the hull are fitted with leak-proof bearings" is part of a structural specification of the functional "Hull of vessel should be leak-proof."

This indicates that it may be flawed to call some description structural or functional once and for all. Even the relationship "Length of hull is 140 meters" may be difficult. It can be transformed in the next phase to become "Length of hull should be 140 meters" which may be seen to proximate a statement of functional expectation (as seen from the perspective of the constructor) - not of structural fact (as seen from the designer). When the constructor makes a hull that is 140 meters, he tries to meet the expectations posed by the designer!

In some sense, one may see the functional facet as capturing the rationale behind the structural facet. In other words, the latter is intended to satisfy the former. For instance, part of the functional expectations of the hull of a ship is that it should "Make the ship float," while parts of its structural description may be "Length equals 140 meters," "Beam equals30 meters," "Draught equals 14 meters," and "Block coefficient equals 0.67." Is then the statement "Displacement equals 39,396 meters[3,] an expression of structural nature or functional nature? It may be both, depending on where we stand. Seen from the 'perspective' of the ship, the statement is of structural nature, in that it satisfies the demands put to 'hull ' from 'ship'. Seen from the perspective of the 'hull ', it is of a somewhat more functional nature, in that 'length', 'beam', 'draught', and 'block coefficient' satisfies the demands put to them from the 'hull'. The issue is a difficult one to resolve. Suffices it to say here that there exists on ambiguity, and that it is difficult to separate the world into one dealing with structure (or 'facts) and one dealing with function (or 'expectations'). The division is context dependent and thus may be expected to change as the design process evolves.

Let me, in conclusion, propose the following:

> A design object is <u>S-F complete</u> if the combination of structural description and functional expectations is such that it, if true in the materialized result, would prove the initial expectations to be met.

In other words, if there is only one function needed to meet the initial specification that is either (i) not stated as a functional expectation to the constructor or (ii) does not result from the materialization of the structural description, then the D-object is not S-F complete.

It is further possible to say something about what is relevant knowledge related to the D-object:

> If the D-object is S-F complete and there are parts of the structural description or functional expectations that can be removed without causing the D-object to cease to be S-F complete, the D-object is <u>S-F overloaded.</u>

This says that if there may be statements about the D-object that can be removed without this causing any change in the degree to which expectations are met. This may mean either that (i) there is irrelevant knowledge about the D-object or (ii) there is redundant knowledge about the D-object. In any case, one would expect that the statements may be removed without any implication to the way the materialized result corresponds to the expectations.

# Axiomatization

As stated in the beginning, this chapter had two purposes. One was to discuss design from the particular perspective of concept formation, and to find characteristics of the process from that point of view. The other was to prepare the ground for building a Design Process Language through which concepts may actually be formed.

Some of the findings summarized here will be used in developing DPL in Chapter 5. There is no argument, however, that DPL will capture all aspects of the design process as formulated here. Neither is it necessarily so that all the statements are relevant for the sake of developing the DPL. The conclusions of this section, in the form of axioms, is therefore primarily to draw some lines from the discussions performed so far. That some of the axioms are used while others aren't used in the remainder of the thesis will hopefully not constitute difficulties.

### 4.1: Definition

> If a concept is known in terms of its relationship to other, known, concepts, it is considered as known in propositional memory.

This simply states a definition of propositional memory, such that if two concepts are explicitly related this is 'propositional· knowledge. A concept described in this way can be termed a propositional concept.

### 4.2: Definition

> If a concept is known in terms of its relationship to the real world, it is considered as known in semantic memory.

A concept that is known because it can be directly related to a 'real' phenomenon, this is a <u>semantic concept.</u> This is in contrast to the propositional concepts defined in 4.1.

### 4.3: Definition

> Knowledge generated during the design process is considered to reside in design memory, where design memory is a subset of propositional memory.

A quite self-explanatory statement saying that the relationships constructed throughout the design process are considered stored in design memory. Note that this does not necessarily mean that there is a physical location as such - it is merely a means to reference 'design knowledge'.

**4.4: Definition**

> A concept is formed by establishing relationships in propositional memory where the concept is related to other concepts.

The knowledge that is produced during the design process is represented as relationships in propositional memory. This is to say that it is assumed that one cannot define a concept semantically in the design process (which is again to say that a designer cannot describe a concept through actually point at a 'real' occurrence of the phenomenon to which the concept refers). Another way to put this is that all semantic concepts used in the process must initially exist.

**4.5: Definition**

> A <u>generally understood concept</u> is either a semantic concept or is related to such a concept through relationships in propositional, but not design, memory.

This definition states that a concept is generally understood if it is 'described' in semantic memory and in that part of propositional memory which is not design memory. The assumption is that all have access to a generally understood concept. ('All' in this context means all that are expected to understand the concept of the D-object.)

**4.6: Axiom**

> The design process may be viewed as a concept formation process.

This axiom follows from earlier discussion, where the D-object was a concept that was initially unknown. Since the D-object cannot be a generally understood concept since it is non-existing at the outset, the designer must define it in terms of (ultimately) generally understood concept. This process of definition is called <u>concept formation,</u> sometimes also <u>conceptualizing</u>.

**4.7: Corollary**

> The primary purpose of the design process is to form a concept
> that can meet the expectations put to that concept.

As described earlier, the design process is a directed search towards defining a concept- conceptualizing the D-object. The direction of the search is given by the expectations that the D-object is to meet - expectations that include requirements and constraints. Roughly speaking, the success of the concept that is formed depends on the degree to which it meets the expectations.

**4.8: Theorem**

> Design may be viewed as a process of generating relationships
> between the concept in formation and concepts that may
> already be known or may have to be formed in propositional
> memory.

In view of Axiom 4.6 and the definitions of how a concept becomes 'known', the concept being formed can only be formed by relationships being established between it and other concepts. If these other concepts aren't generally understood, they have to be described in terms of their relationships with other concepts. This repeats until all concepts initially not generally understood would become so if design memory could generally be accessed.

**4.9: Theorem**

> In the end, for a D-object to be generally understood, at least
> some concepts related to it must be generally understood.

In some sense, this theorem overlaps Theorem 4.8, but stresses the point that some generally understood concepts must be related to the D-object: It was argued earlier in the chapter that in a purely symbolic system this weren't necessary. However, design aims to define a concept ultimately to be materialized in the 'real ' world and by some 'constructor' that necessarily needs to know the implication of the D-object in 'real' terms.

### 4.10: Axiom

> The set of relationships describing the concept of the design
> object may be viewed as facts by the designer.

As discussed earlier, the designer is to conceptualize a D-object that can meet the expectations put to it. In short, the D-object is then described such that when placed in its environment, it will prove the expectations to be met. The designer models the D-object for performance, and may be viewed as treating the D-object as a set of facts - as already having been built according to the description.

### 4.11: Axiom

> The set of relationships describing the concept of the
> expectations may be viewed as hypotheses by the designer.

In relation to the above axiom, the designer may view the expectations as a set of hypotheses that she is about to prove with the D-object. As scientists test hypotheses against facts to verify the hypotheses, the designer tests the design result to see whether it meets the expectations.

### 4.12: Theorem

> The design process may be viewed as a process of proving
> hypotheses.

As a consequence from axioms 4.1O and 4.11, one may cast the design process as a process of proving hypotheses. The design process is shown to be a directed search for a solution - a D-object. In this view, the expectations may be cast as hypotheses that the D-object is to prove. The actual proofs are performed as a kind of 'performance modelling· where the specification and the outcome of the performance analysis of that model constitute the 'facts'.

**4.13: Axiom**

> The set of relationships describing the concept of the design
> object may be viewed as hypotheses to those that materialize
> the design object.

This axiom changes the perspective from the designer to the agent that is to materialize the D-object. For this agent, the D-object is a description of the expectations to be fulfilled by the process of materializing. The D-object specification may then be viewed as hypotheses which are to be proven by the result of the materialization. (In a ship design situation, a trial trip may be an appropriate 'test'.)

**4.14: Axiom**

> The set of relationships describing the materialized concept of
> the design object may be viewed as facts.

In consequence of the previous axiom, the implemented (materialized) result constitute the facts towards which the hypotheses (which now is the D-object specification) can be tested. In a ship design situation, the physical ship is the facts and the building specification is the hypothesis.

**4.15: Theorem**

> The design process may be viewed as the overall process of
> generating hypotheses.

As in the situation described in Theorem 4.12, with the perspective changed from that of satisfying expectations (posted by a 'customer'), the designer may now be viewed as one that aims to construct hypotheses to be met by the agent that is to materialize the D-object. Note that the D-object is more 'hypothetical' than 'real', and therefore is subject to verification by testing after its materialization as well as after its design.

**4.16: Definition**

> The creation of a product Is viewed as occurring through two phases or processes; design and materialization. The <u>phase of design</u> defines a <u>concept</u> that is called the D-object, while the <u>phase of materialization</u> Implements this concept into a <u>product</u> that can be described as the D-object.

This Introduces a view of design occurring in two phases, where one alms to define the concept of D-object while the other alms to transform this concept (which Is yet only abstract) to a concrete and existing concept. It Is clear that this segregation Is only one of many possible, but it Is Introduced here to ease the discussions.

**4.17: Axiom**

> A relationship may be viewed as a sentence In a language.

It has become apparent in the course of discussions of literature on design and knowledge modelling, both in Chapter 2 and Chapter 3, that the relationships discussed so far are similar to sentences In some language. Earlier in this chapter the Idea has been further developed. The language in question will be referred to as a <u>design language.</u>

**4.18: Definition**

> A sentence in the design language has two main categories of symbols; concepts and relations.

It should be clear from the discussions so far that the two main 'carriers' of meaning are concepts and relations.

**4.19: Definition**

> A relation in a sentence applies to relate concepts or sentences within the sentence.

This Is quite self-evident. It has frequently been shown that concepts are related by means of relations, and also that sentences themselves may be related. Pronouns that are useful in a design language are normally used as place-keepers for concepts and may thus be treated as ordinary concepts. Some exceptions are discussed In Chapter 5.

### 4.20: Corollary

>   A concept is formed according to what kinds of relations are
>   used to relate it to other concepts.

This is a quite self-evident statement, since what relates concepts are relations, and what brings meaning to concepts is the way concepts are related. Thus, the relation used in a sentence is critical in determining what meaning that sentence 'carries to ' the concepts.

### 4.21: Axiom

>   The design object may be described through a set of
>   sentences.

In the course of discussions in this chapter it has been shown that the design object may, as any other concept, be described through its relationship with other (ultimately known) concepts. It may thus be described through a set of sentences.

### 4.22: Axiom

>   The process of design may be described through a set of
>   sentences.

This has been developed throughout the course of this chapter. Since the process Is described, both by myself and others in Chapter 2, as a set of activities related In some way (ordered), it follows that it is possible to construct sentences in which the related concepts refer to activities or other processes.

### 4.23: Axiom

>   It is possible to define relations to express any kind of
>   relationships needed in design.

It Is hard to prove the correctness of this axiom. However, since design (as viewed In this thesis) deals with 'symbolic· knowledge, or that which may be denoted by words, It Is self-evident that there exists means with which the concepts receive meaning through other concepts. There must necessarily exist means of representing such knowledge, and the axiom claims that the relations used are definable.

**4.24: Theorem**

> Such relations as in Axiom 4.23 may be categorized. This categorization may be seen to correspond to what kinds of relations between concepts are required to be expressed in a design situation.

The intention behind this corollary is to indicate that various relations play various roles in a language. The relations may be grouped according to what kinds of relationship they express in the design context. That is, whether there is a possessive relationships (mainly transitive, verbal relations), a relationship of intention and future (modal auxiliary verbal relations), or spatial and temporal (prepositions and adjectives).

**4.25: Axiom**

> A sentence may be interpreted through grammatical rules operating on the relations in it.

In linguistics, the main grammatical rules operate on the verbal and other relations in the sentence. What the sentence says about how a concept becomes defined in a design language depends on how the relations connecting these concepts are interpreted. It seems proper to claim that grammatical rules apply to the relations, rather than the concepts, in the language.

**4.26: Axiom**

> It is possible to separate between those sentences describing how a particular role of a concept is or will be fulfilled (structural description) and those sentences describing the roles that the same concept indeed does or shall fulfill (functional description).

This says that there is a distinction between a concept and the implication of the concept. For instance, it is possible to distinguish between a ship and its role or function. Some (Goel and Pirolli, e.g.) have called this a distinction between a causal structure specification and function. Yoshikawa distinguished, on his side, between an entity and a concrete function. The distinction may be illustrated in two ways; one in the hypothesis-proving view of design, and another in a more 'concept-formation' view:

Hypotheses -> Facts -> Hypotheses -> Facts

> ? floats? -> planned ship floats! -> materialized ship floats? -> materialized ship floats!

Expectations -> Structure -> Environment -> Function -> Expectations.

> "D-object should float in water"-> "D-object is a ship"-> "Ship in water" -> "Ship floats in water"-> "D-object does float in water"

### 4.27: Hypothesis

> There exist relations that may be used to construct the needed sentences in order to <u>form a concept</u> in the design process.

I hypothesize the existence of such relations, and aims to show these categories in Chapter 5. This hypothesis applies to the knowledge <u>being generated</u> in the process.

### 4.28: Hypothesis

> There exist relations that may be used to express the needed relationships between concepts <u>applied</u> in the design process.

As above, but the relations are <u>used</u> to express knowledge that is used in the design process - not that which is <u>produced in</u> it. The two hypotheses separate between that which is known and that which is becoming known in the process.

### 4.29: Hypothesis

> The categories of relations whose existence were hypothesized in 4.27 and 4.28 may be seen as categories in the same language.

This statement hypothesizes that the 'process' and 'content· knowledge that exists, and that one which is produced, may be described in the same language and interpreted through the same kinds of grammatical rules. This will be treated in the course of Chapter 5.

# Chapter summary

In this chapter I have performed parts of the main development in the thesis, through introducing the *view* of the design process as a process aimed at forming concepts. To perform the concept formation, a designer needs to form the concepts of plans and sequences of activities to use in forming other concepts -  produce this new knowledge. The design process is described as a set of relationships between abilities that may be represented as concepts in the same language as that which describes the design object. Thus, like "Activity A follows activity B" is a <u>process</u> description and "Deck A is above deck B" is a <u>content (object)</u> description, they are both relationships among concepts and both representable in the language. These relationships are termed <u>sentences</u> in the <u>Design Process Language</u>. A designer tries to form new concepts by relating them to known things in a way such that all information relevant for assessing the general behavior of the design object, may be defined from knowledge about known things.

I have developed an alternative view of design as proving and/or constructing hypotheses. The designer may be viewed to attempt to prove hypotheses about how the design object will behave in the future, by designing her design object such that it has the best chances of displaying such behavior as expected. On the other hand, nothing that the designer does is necessarily 'objectively' true in the future - there is another stage after the design in which the description is to be <u>materialized.</u> Seen from that angle, the designer constructs hypotheses that the constructor performing in the next stage (the 'materialization stage) will take *over* and try to satisfy through materializing a product that has the best chances of proving the hypotheses that the designer constructed. The designer may actually be seen as a  <u>mediator</u> between a 'customer ', who <u>expresses expectations,</u> and a 'constructor ' that <u>materializes</u> the design object as shown in Box 4.iv.

Towards the end of the chapter, the commonly referenced division between structure and function is argued to be somewhat elusive. It is difficult to see where the one ends and the other starts. It is claimed that the specification of a design object is a combination of expressions of structural and functional nature. This specification should be <u>S-F (structurally-functionally) complete</u> at the end of the process, i.e., the combination of new functional expectations and structural description presented by the designer should be such that the specification can <u>prove</u> the expectations that the designer set out to meet - or such that it can <u>prove the initial hypothesis.</u> In other words, all parts of the design object are either described structurally, in terms of dimensions and other 'tangible' information, or functionally, in terms of behavior expected from a specific part. As a designer in a <u>conceptual design phase</u> may lay out the main dimensions of the hull in terms of 'length', 'beam', and 'draught· of a vessel (structural), the same designer in the same phase may describe the 'engine' in terms of expected behavior (functional).

# Chapter 5: Developing relations

In previous discussions the *relations* were shown to be crucial in representing knowledge. This chapter will briefly summarize the most important findings presented so far and draw some implications of them in terms of the Design Process Language (DPL). The purpose is chiefly to discuss different phenomena related to design and to develop the kind of relations used to express them. I do not intend to define these particular relations nor any grammatical rules. I intend to show that it is possible to classify the relations so that the classes are meaningful in the context of design. I will further argue that it is possible to build a language from these relations that is an expressive one in the context of design, both for the general purpose of developing theories about design and for the specific purpose of developing a computer-based design 'assistant'.

Thus, while it is probably easiest to see the application of DPL in the context of automated design, it may be less apparent that the discussion of language is closely related to the discussion of the phenomena that the language is to express. Thus, the DPL may be seen as a means towards an end which is to shed some light over what phenomena that are involved in design process and object description. I thus hope that the dual purpose of this work will shine through; to discuss phenomena regarding design as the language develops, and to develop the language as the phenomena are discussed.

## Requirements of the Design Process Language

In the course of the previous four chapters, several characteristics of the design process, and the knowledge used and produced by it, have been distinguished. It is thus possible to develop some requirements by which the DPL should be judged for 'epistemological adequacy' in design contexts.[a]

---

[a]   "A representation is called epistemologically adequate for a person or a machine if It can be used practically to express the facts that one actually has about the aspect of the world." (McCarthy and Hayes, 1973, page 399)

Let me recapture the hypotheses 4.27, 4.28, and 4.29, from the previous chapter:

> There exist relations that may be used to construct the needed sentences in order to <u>form a concept</u> in the design process.

> There exist relations that may be used to express the needed relationship between concepts <u>applied</u> in the design process.

> The categories of relations whose existence was hypothesized above can be seen as categories in the same language.

Recall that a concept is formed through sentences ultimately relating it to concepts that are known either in propositional memory or semantic memory. For a D-object to be <u>generally known</u>, at least some concepts used to define it, as well as all the concepts that defines the D-object, must be generally known. A sentence is built from two main categories of symbols - concepts and relations. The relations could (according to Axiom 4.25) be operated upon by grammatical rules and categorized according to the phenomena that they express (Axiom 4.24).

It then remains to develop two main ideas in order to establish a framework for the Design Process Language; (i) the particular phenomena that are necessary to express as relationships in design, and (ii) the groups of relations used to express these. In this section I will show some examples of sentences describing phenomena that one would expect to be relevant in the design context. I will also, with reference to these examples, introduce some requirements that the relations in the Design Process Language must satisfy. Thus, (i) above will be started here but continued through the discussion of each category of relations. It is necessary to develop these ideas in a stepwise fashion, since in my view it makes little sense, as frequently stated earlier, to discuss the phenomena to be described separately from the language used to express them.

I have two main reservations to state. The first is that I do not claim these requirements to constitute a <u>complete</u> set of requirements that the relations are to satisfy. One reason for this is that to do this would bring me far beyond the limits of my abilities, specifically in terms of temporal resources at disposal. Another reason is that it may simply not be possible to prove sufficiency in a language used for the purposes of expressing phenomena in design because too little is known about what design is. The other reservation is that I do not claim to satisfy all the requirements stated here, grossly of the same reasons as above. What I <u>can</u> do is to claim that all requirements are <u>relevant</u> and that the Design Process Language is a <u>useful</u> continuation on the road towards achieving 'epistemological adequacy' in the context of design.

One kind of phenomenon is related to that of expressing activities, events, processes, and the like. The relation of interest here is the verb that describes the kind of activity being performed:

> "The program *calculates* the weight of the ship"
> "The designer *starts* the weight calculation

• It must be possible to define the activity of doing something.

Another kind of phenomenon is that concerned with what a concept is, how it is identified, and what it possesses of parts and characteristics. In addition, it is necessary to describe in what circumstances a concept exists - that is, how it relates to other concepts not possessed by it:

> "The vessel *is* a container ship" and  "The vessel *is* 50 meters long"
> "The vessel *is* the  container ship"
> "The vessel *is at* the  dock" and "The vessel *satisfies* the requirements" "The vessel *is* the  one at the dock"
> "The vessel *has* a length of 50 meters"
> "The company *owns* the vessel" and  "The vessel *is* the property of the company"
> "The vessel *is at* the  dock"

\* It must be possible to define the 'content' of a concept,
It must be possible to define how a concept relates to other concepts not included in its 'definition'.

Another kind.of relation is that which simply has to do with stating the existence of a concept:

> "There is a D-object"
> "There are expectations"

\*      It must be possible to define the existence of a concept.

An important category of relations are those that concern ability, futurity, possibility, determination, expectation, and in general those phenomena that have to do with planning and purpose. In addition, the relation *do* is important in stating 'facts' in different tenses:

> "The program *can* calculate the number"
> "The program *could* calculate the number"
> "The number *may* equal 7"
> "The number *might* equal 7"
> "The program *might* calculate the number"
> "The D-object *should* satisfy the requirements"
> "The program *will* calculate the number"
> "The vessel *ought* to look nice"
> "The hull *must* resist water-pressure"
> "The designer *must* calculate stability"
> "The hull *does* resist water-pressure"
> "The designer *did* calculate stability"

\*      It should be possible to describe an ability to perform an activity.
\*      It should be possible to describe the necessity of performing an activity.
\*      It should be possible to describe the determination of performing an activity.
\*      It should be possible to describe the purpose or expectations of a concept.
\*      It should be possible to describe what is required of a concept.

Another kind of phenomenon of importance is how concepts, or objects, relate in space and time. Further, it is important to refer to a concept through its affiliated concepts or Its source - that is, to related concepts:

"The machinery is *on* deck 3"
"Plan A will be executed *before* plan B"
"Machinery is subsystem *of* vessel"
"The formula is calculated *by* the program"
"The propeller is connected *to* the propeller shaft"
"The vessel is analyzed *as* a structure"

* It should be possible to describe how concepts spatially and temporally relate.
* It should be possible to refer a concept through its related concepts ("owner' or 'origin).

Another important kind of relations are those joining sentences or concepts. Such joining may, for instance, be logical, causal, inclusive, or exclusive:

"The beam increases *with* increasing length"
"The program runs *while* increments are large"
"The vessel satisfies requirements *if* length is less than 140 meters"
*"If* one concept is cheaper than another concept *then* choose the cheapest concept"
"The Color is Red *or* Green."
"The machinery is a steam turbine *or* a gas turbine"
"The Crane is on the Deck *because* it should reach *(both)* Cargo Hold *and* Shore"
"The length is larger than maximum length *therefore* the designer must decrease length"
"The designer must decrease length *since* the length is larger than maximum length"
"The designer executes Program A *when* Program B has executed"
"The designer executes Program A *or* Program B"

"The designer executes Program A *if* Program B fails"
*"If* Program A is faster than Program B *then*
the designer executes Program A"

* It should be possible to logically relate sentences or concepts *(if, while).*
* It should be possible to causally relate sentences or concepts *(since, because).*
* It should be possible to inclusively relate sentences or concepts *(both, and)*
* It should be possible to exclusively relate sentences or concepts *(neither, nor).*
* It should be possible to describe the purpose of performing an activity.
* It should be possible to define rules of selecting among concepts. *(if, then)*
* It should be possible to make alternative definitions of concepts. *(or)*

In the design process, an important activity is to compare alternatives, or describe one concept in comparison to another concept:

"Vessel A is *larger* than vessel B" "Program
A calculates *faster* than Program B"
"Choose the *cheapest* vessel"
"Vessel A is *cheaper* than vessel B"
"Vessel A is the *cheapest'*

* It should be possible to describe how concepts compare.
* It should be possible to describe a concept as compared to others.

Some other modifiers are discussed in the case where fuzziness or uncertainty is introduced into the design process:

"Vessel A is *possibly* cheaper than vessel B"
"Vessel A is *not* cheaper than vessel B"

\*    It may be useful to introduce fuzziness in otherwise deterministic relations.

\*    It should be possible to represent negation of a statement.

The sentences shown so far are propositions stating facts about the world. However, it is necessary to be able to enact, or mobilize, abilities and to make queries regarding what is stated about the world:

> "The weight is 12,000 tons"
> "Execute program A"
> "What is the weight?" - "Which program is executed?"

\*    It should be possible to distinguish between commands, propositions, and questions.

Another issue is that dealing with a historical record of a design process. In addition,it is often necessary to distinguish between future and present, specifically as goes for planning activities:

> "The program *will calculate* weight" - "The program *calculates* weight"
> - "The program *calculated* weight"

\*    It should be possible to distinguish between past, present and future.

Some terminology is traditionally introduced to denote knowledge about knowledge. These are specifically those that have to do with classification, attribution, and systemizing. All these terms imply at least two concepts and a relation between these concepts:

> "The D-object is a ship" - "Ships is a *class"* - "'Anne Lise' is an *instance* of ships" - "Container ships is a *subclass* of ships"

> "The ship has a length" - "Length is an *attribute* of the D-object"

"The length is 140 meters" -   "140 meters is the *value* of length"
"The ship has length 140 meters" -   "140 meters length is a *property* of the
D-object"

"The ship is a *system"* -   "The machinery is a *part"* - "The machinery is an
*assembly'* -   "The machinery is a *subsystem"*

\*        It should be possible to define classes and class-related concepts.
         It should be possible to define attributes, values, and properties.
\*        It should be possible to define systems and system-related concepts.

━━━━━━━━━━━━━━━━━━━━━━━━━━━━━━━━━━━━━━━━━━━━━━━━━━━

# The framework of natural language

The remainder of the thesis develops the Design Process Language (DPL), As
has been frequently stressed, one of the main purposes with the DPL is to
enable design process modelling and D-object modelling in the same general
language. It has also been stressed that process and content are tightly
interwoven, and that they are grossly inseparable. The inter-dependency
between process and object is particularly apparent when the designer as an
intelligent system learns from experience or constructs new tools to solve new
problems. In both cases, adaptation requires the system to access all
knowledge in the content theory introduced in Chapter 3.

Concretely, the aim with DPL is to both give tools for describing the content of
memory at any given state and for describing interventions that affect the content.
These interventions may be perceived as performed by 'agents' that may be seen
as parts of the content theory, in McDermott's words. Thus follows: In DPL the
process model is a particular aspect of the content theory, in which agents
(belonging to the 'content) are mobilized to perform a process. The *mobilization*
is performed by using certain relations in the language, in particular (latent)
active (or modal, transitive) verb relations. For instance, (1) "Joe designs a
vessel" is a mobilization of the relationship (2) "Joe can design a vessel." The
relationship (2) belongs to a content model, while (1) belongs to a process model.

It is useful also to view the design process as a process of constructing working models. A designer will necessarily plan ahead the steps to make in course of design - the interventions are premeditated. However, just as no chess-player can plan all steps to take in detail before the game starts and therefore must revise and extend the process model as the game develops, so must the designer. This highlights the role of Sternberg's 'components' that were discussed in Chapter 2, in particular the components "component selection" and "component ordering". When the designer chooses a plan of work, it is possible to describe this as a plan that ensures a mobilization of latent abilities (can-relationships) to enact in the future. Such a situation is illustrated in Box 5.ii.

Assume that the following is represented in content theory:

1.     "3*2" *is* a Formula
2.     Calculation_f *can calculate* a Formula
3.     Calculation_f *can return* a Result
4.     Calculation_f *is* a C-program
5.     Calculation_f *has* Code
6.          Code *of* Calculation_f *is*
                                        double main (Formula)
                                        {

                                        return (Result)
                                        }

The following might belong to the process model:

7.     Calculation_f *calculates* "3*2" *and returns* Result

The process model may be viewed as a *mobilization* based on sentences in the content model. The specific process enacted in sentence 7 is performed by removing the relation *can* (a modal verb) from sentence 2. After the process has been mobilized and de-mobilized, the following sentences might be added to the content theory:

8.     The Result *is* 6
9.     Calculation_f *calculated* "3*2" *and returned* 6 (a historical record)

As is seen from the very superficial illustration, the language used in both content and process description is the same - the difference lies in the kinds of (modal *con*) or also aspects of (infinite to present) relations used. In this case the process was described through the verb relation *calculate.*

It should easily be recognized that there must exist a connection in the language used in both content theory and process model. The construction of the sentence (7) (that is part of the process model) depends on access to the sentences (1) through (6).

**Box5.i:**     An example of a representation of 'content' and 'process' in the same language.

| Problem: | Find and modify a prototype vessel to introduce into 'content theory': |
|---|---|

Plan:

| Step | Action (relationship in 'process model') | Ability (in 'content theory') |
|---|---|---|
| (l) | Find a comparable, existing vessel | *"can* find a …" |
| (2) | Adjust parameters to fit the current conditions by ... | *"can* adjust parameters ..." |
| (3) | Test whether vessel is OK by ... | *"can* test whether ..." |
| (4) | Proceed If vessel is OK | *"can* proceed if vessel is ok" |

| Box5.iI: | Planning a mobilization of abilities in design. |
|---|---|

As has been argued in Chapters 3 and 4, to represent knowledge is essentially to establish relationships among concepts, whether these relationships describe 'process· or 'content ', I focus on the kinds of relations that may be relevant in developing a language that will make possible the representation of knowledge as it pertains to design object and design process. I will first argue for the use of <u>natural language</u> as an appropriate platform, and will then pursue to indicate some invariants in the classes of relations in the DPL.

---

"(T)here are psychological grounds for supposing that mental models are constructed by a semantics that operates on superficial propositional representations of the premises. In general terms, this process consists in mapping strings of symbols into models." (Johnson-Laird, page 167)

**Box5.iii**

---

It is noteworthy that most models and model languages indeed do apply selected elements from natural language to describe relations. <u>Predicate calculus and formal logic</u> tend to apply terms categorized as <u>prepositions</u> in natural language, such as *Above, Below, On,* and so forth, as well as <u>verbs</u> like *implies* (=>) and *equals(=)*. The <u>conjunctives</u> *and, or,* and *if* are used frequently both in <u>general computer programming</u> and specifically in 'expert systems' applying production rules and formal logic based on fundamental axioms and theorems. In addition, a variety of representational systems make use of composed relations like <u>part-of(noun-preposition</u> relations), *has-subclass(verb-*<u>.noun</u> relations), *composed-of* (<u>verb-preposition</u> relations), *less-than* (<u>adjective-conjunctive</u> relations), *if-exists* (<u>conjunctive-verb</u> relations), and *can-calculate* (<u>verb-verb</u> relation). Again, it is apparent that <u>such relationships are closely</u>

related to sentences in natural language, where the relations in specific are
selected terms from natural language.

---

"Once we assume that language and thought are different things, we're lost in trying to
piece together what was never separate in the first place." (Minsky, 1986, page 198)

.**Box 5.iv**

---

Some of the reason for key terms in natural language appearing in 'artificial'
languages is that it simply is convenient. Using terms that are 'naturally'
understood may help the users of the 'artificial' languages grasp the semantics
of these terms. It is, of course, not necessary to use 'natural' terms in these
languages. Both automated and human systems may operate on entirely
absurd terminology (as seen from the outside) provided that these terms are
understood by the system that applies them and that there is a consistent
grammar. In predicate calculus, for instance, the particular terms could easily
be redefined without any loss in efficiency - all semantics is explicitly
introduced in the universe of discourse without the terms having to mean
anything outside this universe -  the system is closed.

In most computing languages, the names of particular relations, like IF ...  THEN,
DO ... WHILE, FOR …, GOTO ...,   are probably transferred from natural
language for sakes of 'convenience' in order for the programmer to more
easily understand the semantics attached to the terms - not because the
words themselves are unreplaceable.

---

"How much more convenient it would be if theories of thinking, instead of employing
conventional mathematics, could be expressed formally using the same kind of symbolic
representation as are used in thought itself!" (Langley et. al., 1987, page 33)

---

When it comes to meaning the picture is different; the semantics, or the
operations, affiliated with the terms is indeed central in performing operations
upon and representing knowledge. This should come as no surprise. Natural
languages have evolved to reflect the needs of humans to represent
phenomena -  in particular to support cognitive processes. As Johnson-Laird
indicates above, the possibility of performing intellectual processes rests upon
the fact that the language itself has a grammar that supports such processes.

Thus, Johnson-Laird implies that the development of early abilities to make inferences does not follow from a child's ability in formal logic, but rather from a child's ability in language, specifically to understand the power of the grammar. For instance, a child knowing that "a cat cannot fly" will know that if something flies it Is not a cat. Johnson-Laird argues that this deduction is tied to knowledge of the semantics of *if* - not to knowledge of the logical rules (e.g., *modus tolens)* that formalize this deduction. This is illustrated in (1) and (2) below.

1  $\forall x \; Cat(x) \Rightarrow \neg CanFly(x)$
2  $\forall p \forall q \; (p \Rightarrow \neg q) \Rightarrow (q \Rightarrow \neg p)$  {*Modus Tolens*}
3  $CanFly(Hekate)$  {*Observation*}          **(1) Representation in predicate calculus.**
4  $\forall x \; CanFly(x) \Rightarrow \neg Cat(x)$  {1 *and* 2}
5  $\neg Cat(Hekate)$  {3 *and* 4}

1.  *If "x is a cat" then "x can not fly"*
2.  *If "if p then not q" then "if q then not p"*
3.  *Hekate can fly*
4.  *If "x can fly" then "x is not a cat"*          **(2) Representation in 'natural' language**
5.  *Hekate is not a cat*

This 'semantics of language' is one of the prime reasons for the Design Process Language to be developed on the platform of natural language. One fundamental premise is that if main classes of relations in natural language are identified, this will shed light on how design process and object can be represented. In addition, as key terms are found in this language, it may be expected that key characteristics of the design process and object - or design knowledge - can be affiliated with these terms. As each relation, roughly speaking, has a unique role in language so will the phenomenon it describes have a unique role in a relationship describing some 'real ' phenomenon, as indicated at the beginning if this chapter.

It does not require a language that is as rich as the spoken or written one to apply elements, or at least the general structure, of it in design object and process modeling. The design process is for the most part the result of a quite premeditated effort, where activities are intended to be tightly controlled. This reduces the need for a large set of qualities and words found in natural languages, as for behavioral and mental 'processes' (see Table 5.1).

It is not argued here that the English terms used are necessarily the best ones to use in DPL. Neither is the purpose to necessarily distinguish some relations as more important than others. It is instead to argue that there are some groups of relations that are important in describing the design process and the knowledge that this process manipulates. It is further to show that the linguistic categories used here are good categories for discussing the role of the relations in any language such as the DPL.

There may be some reasons for natural language not to be considered appropriate in the context of DPL. I will counter some possible arguments in brief here:

(1)   Natural languages are too ambiguous for use in highly structured processes, that is, where logic or computer application is central.

If DPL were to be a complete language for describing any conceivable phenomenon, this would be a legitimate concern. However, the scope of the DPL is to provide a language where important (and not necessarily all) phenomena can be described and manipulated. It is further not to be a means for a <u>machine</u> to understand the way <u>humans</u> communicate. To the extent that machine/human communication is the issue, the crucial idea is to let their respective languages 'meet' at the point where the DPL is structured according to rigorous rules, but where those rules and the terminology is a subset of the natural language. Thus, DPL singles out the more structured parts of the English language and defines relations such that the terms denoting these relations are uniquely understood in a design environment. I claim that this may be an epistemologically sound (or strong) and intuitively appealing form of communication and processing.

(2)   Natural language representation is inefficient for speedy search and retrieval, as is required from large database systems.

There are particularly two reasons why this is not a relevant question in the context of DPL: (i) DPL is not developed with a view towards achieving efficiency (in terms of information processed per CPU) - it is geared towards achieving epistemological adequacy (see footnote (a)) in the context of design; and (ii) if DPL indeed were primarily focused on developing a language for computer purposes, it is more than likely that developments in computing technology would increase the processing speeds of information to the extent that the problem would be more <u>what kinds of</u> information to process rather than <u>how fast</u> it should be processed.

(3)     It is too rich in concepts for most purposes, as in most traditional computer programming, and requires a too extensive syntax. Thus, the grammar would be overly complicated.

As already said under (1), DPL is a segment of a natural language, where the relevant grammatical rules are chosen to meet the requirements of a language constructed for describing design process and object. As in any traditional (computing) languages, the syntax and grammar may be reduced to only be epistemologically adequate, where any user and context determines the interpretation of 'adequate·.

(4)     A complete syntactical description (including grammatical rules) is too extensive to practically implement in computers.

This is discussed under (1) and (3); the extent of the DPL can be adjusted to any purpose, and computer technology will probably reduce this problem to a detail.

(5)     It is not sophisticated (scientific) enough.

Since there already exist languages that have shown to be powerful in terms of representational abilities, I can see no need to construct an entirely new language from scratch. It seems more appropriate to extract from those existing languages the rules needed to represent a phenomenon adequately. This approach may be pragmatic, and in some sense cynical, but there are chances that the result is a more complete and intuitively 'right' kind of language.

# Relations and modifiers in the Design Process Language

The main classes of relations in DPL are the *verb relations, preposition relations,* and *conjunctive relations.* In addition, there are the classes of *adverb* and *adjective modifiers.* These latter terms are briefly described - the focus is to develop the verb relations since these may be argued to have a most central role, particularly as goes process modelling. Towards the end are introduced some abstract terms that are frequently used in 'talking about knowledge ' - terms that refer to what 'kind of' concept being discussed. For the sake of discussion, let me focus the scope of the discussion to simple *propositions,* or relationships that are either true or not. Thus, questions and commands, for instance, are discussed separately - the general understanding of the relations are assumed to be the same regardless of whether they are used in a question, command, or proposition.

## Verb relations

Verbs are central *carriers of meaning* in the language - they are present in ll complete sentences in daily conversation. This section discusses the main kinds of verb relations in a Design Process Language.

The earlier discussions have referred to roughly four kinds of verb relations, those that concern existence, those that concern possession, those that concern action, and those that concern future, determination, requirement, and the like. This division is by no means exclusive - one verb may play different roles according to the context in which it is used. Furthermore, it is difficult to establish criteria for determining whether one verb belongs to one class or the other, and several verbs do not necessarily belong to either of these classes, as may be the case with those verbs that concern mental states, such as *love, like, hate,* and so forth. As mentioned on page 120 the classes of verbs discussed here are those that are viewed to be most relevant in describing design processes as structured and goal-directed problem solving, and in describing objects that essentially are quite tangible.

Researchers into linguistics have been involved in discussing the roles that verbs play in communication between humans and in representing information, and results from work within this field can be applied in the context of DPL. Halliday summarizes the verb relations (process types) in the manner as shown in Table 5.1.

| Process type | Category meaning | Participants |
|---|---|---|
| material:<br>    action<br>    event | 'doing·<br>    'doing·<br>    'happening· | Actor. Goal |
| behavioral | 'behaving' | Behaver |
| mental:<br>    perception<br>    affection<br>    cognition | 'sensing'<br>    'seeing·<br>    'feeling'<br>    'thinking· | Senser, Phenomenon |
| verbal | 'saying· | Sayer, Target |
| relational:<br>    attribution<br>    identification | 'being'<br>    'attributing·<br>    'identifying· | Token, Value<br>Carrier, Attribute<br>Identified, Identifier |
| existential | 'existing' | Existent |

**Table 5.1:** "Process types. their meanings, and key participants:" (Halliday, 1989, page 131)

The scope here is limited to three of Halliday's 'process types· - <u>material</u>, <u>relational,</u> and <u>existential</u> verb relations.[b] This is not to say that the other relations are irrelevant or unimportant, and not to say that these relations cannot be included in the same frame of reference. It is merely to say that the kinds of relations presented here are the more relevant in a design situation. The last category of relations are the <u>modal</u> verb relations.

### Material verb relations

<u>Transactions,</u> as referred to earlier, have the characteristic feature of causing <u>changes in state.</u> In other words, they in some way alter the content theory or process model when enacted - there is generally an <u>object of their actions.</u>

---

[b]     Wherever Halliday uses the terms processes, in the sense of process carried between segments in a sentence, the term relation is used here, in the sense of relations between concepts in a sentence, i.e. a relationship. Also, Halliday refers to processes as process *clauses,* including adverbs, for instance. The discussion here singles out the verb, which after all is the 'carrier·.

It is clear that a language such as DPL must be capable of expressing such transactions. Grossly speaking, it is possible to cast them as relations between themselves and some aspect of the world. These relations are the *material verb relations.*

The *material verb relations* imply that "some entity 'does' something - which may be done 'to· some other entity. So we can ask about such processes, or 'probe' them, in this way: *What did ... do? What did ... do to ...?"* (Halliday, 1989, pp. l 03-104) In other words, the transactions and interventions experienced in the design process are material processes. Whenever there is something that in common terms would be called activities, these activities are described through material verb relations.

It can generally be assumed that this class of verb relations belongs to the class of *transitive verbs.* This is to say that the verb carries a process from a subject to an object, i.e., the verb relates a subject to an object. The object may be only implicit, like in "The program calculates," where it is understood that something (an object) is being calculated.

In the context of the material verb relations, the transitivity is intuitively strong - the relationship between the subject and the object is one where the subject actually may change the object. This is not so with all transitive verbs, such as *have, contain, include,* and so forth. While these latter verbs represent relations between that which has, contains, or includes and that which 'is' had, included, or contained, they do not represent any change in the object the recipient - of the relation. This is as opposed to the material relations *execute, calculate, remove,* and so forth. To *calculate* something affects the state of that something, whereas to *have* something does not.

This interpretation introduces at least one new problem: What does it mean to affect the state of something? To answer this, the ·universe of discourse· must be kept in mind - the task here is primarily to represent the knowledge that pertains to the design process and the object of this process. In addition, it is focused on using and developing symbolic representation through sentences in the DPL. In this view, it is clear that whenever a new sentence is introduced that differs from a previous sentence, this brings new knowledge into the domain and thus the state (of design memory) is affected.

Say then that whenever a sentence is constructed this reflects some behavior or phenomenon that can be observed in the design process. In other words, the sentence is a  underline{true} description of a phenomenon. Assume further that the sentence describing the phenomenon and the phenomenon itself appear simultaneously -  there is no time-lag. Thus, it is possible to consider a sentence as underline{causing} something, although the something actually is caused by the phenomenon the sentence describes. In other words, a process description may be treated as the process itself,

Under this assumption it appears that if a change in design memory directly results from the construction of a sentence, the relation in that sentence is a material and a strong transitive verb relation. For instance, "The program calculates the weight of the ship" causes a new sentence to appear, saying that "The weight of the ship equals ...," The relation *calculate* can be said to be a underline{direct material verb relation}, henceforth referred to as DMR. Said in another way, if a new sentence B appears underline{immediately following} the sentence A, and if B would not have been constructed were it not for A, then A is a DMR sentence.

A new potential problem may seem to arise when a sentence such as "The designer starts the weight calculation program" is constructed. In this case, no new knowledge apparently is introduced before "the program calculates the weight of the ship" appears. However, recall that design memory also consists knowledge of the process itself. Thus, the latter sentence represents the knowledge added to memory. These relations, like *start, trigger, execute,* are henceforth referred to as indirect material relations, or IMRs. Other relations in the same category are *stop* and *terminate.* These relations describe the *destruction* of an DMR and are also referred to as IMRs. In view of this, the following definitions can be proposed:

**Definition 5.1:**

> If one of the direct consequences of a sentence is that other sentences are introduced into design memory, then the verb in that sentence is a direct material verb (DMR) relation.

**Definition 5.2:**

> If the only direct consequence of a sentence is that other SMR
> sentences are introduced into design memory, then the verb in
> that sentence is an indirect material verb (IMR) relation.

In this framework, the "process model ' referred to earlier is an ordered collection, especially of sequences, of MR sentences. The segregation into direct and indirect MRs roughly corresponds to Halliday's division in Table 5.1, where 'action' may be viewed as a DMR and 'event' as an IMR.

**Intention, circumstance, and possession**

Halliday terms as *relational processes* those relations that concern, in a wide sense, the existence of a concept or rather the way a concept exists. He separates into three types;

| | | |
|---|---|---|
| "(1) | intensive | 'x is a' |
| (2) | circumstantial | 'x is at a' |
| (3) | possessive | 'x has a' |

Each of these comes in two modes:

| | | |
|---|---|---|
| (i) | attributive | 'a is an attribute of x' |
| (ii) | identifying | 'a is the identity of x '" |

(Halliday, 1989, page 112)

In the context of the D-object, examples of these types and modes are:

| | |
|---|---|
| 1.i. | "The vessel *is* a container ship" |
| | "The vessel *is* 50 meters long" |
| 1.ii. | "The vessel *is* the container ship" |
| 2.i. | "The vessel *is at* the dock" |
| 2.ii. | "The vessel *is* the one at the dock" |
| 3.i. | "The vessel *has* a length of 50 meters" |
| ·3.ii. | "The company *owns* the vessel" |
| | "The vessel *is* the property of the company" |

In a design situation, it would seem that the most relevant mode is the *attributive* mode. Most likely, the matter of identifying a concept is less of a problem than attributing 'content· to that concept. There is an important exception in representing choices among alternatives. In that case, the process would involve the *identification* of the alternative to pursue among a set of alternatives. However, the attributive mode is focused here, since this is the mode that is operational when a concept is described, or attributed meaning.

Halliday divides each <u>circumstance-attributive</u> relation into two additional categories of clauses - (a) "circumstance as attribute" and (b) "circumstance as process." In (b) the relation is carried by the verb alone, while in (a) the <u>clause</u> is composed of a verb and a preposition or adjective. Thus, we get:

2.i.a.   "The vessel *is at* the dock"
2.i.b.   "The vessel *satisfies* the requirements"

The relations predominantly discussed here, the <u>attributive modes</u> of the intensive, circumstantial, and possessive verb relations, may all be used for capturing different aspects of concepts. I will briefly discuss their relevance as applies to the D-object, but the discussion can be extended to other concepts as well.

"In the intensive type, the relationship between the two terms is one of sameness; the one 'is' the other." (ibid., page 114) These <u>intensive relations</u> are used for relating the D-object to secondary concepts that may serve to identify the D-object. There are particularly two 'kinds of· secondary concepts that enter into such a sentence; those that concern concepts generally viewed as attributes and those that concern concepts generally viewed as classes. An instance of the first kind is the sentence (A) "50 meters is the length of the vessel" while an instance of the second kind is the sentence (B) "The vessel is a (kind of) container vessel." These two sentences are fundamentally different in nature. Whereas in (A) the vessel is identified by a particular characteristic, unaffected by any classification, in (B) the vessel is identified through classification, unaffected by any particular characteristic aside from class membership.

There is a sense to which this is a logical segregation in the design context. Seen from the perspective of the D-object, (A) relates the D-object 'downwards' to a more elementary and concrete concept that serves to define the particular characteristic 'length.' (B), on the other hand, relates the D-object 'upwards' to a more composed and abstract concept that serves to define a set of characteristics of the D-object. In the context of object- oriented programming, (A) is a relationship between the D-object and local properties, while (B) is a relationship between a D-object and its super-class. Another way to view this is that in (A) the vessel is described such that it is possible to classify it as something, while in (B) the vessel is described such that it is possible to characterize it (through class properties).

"In the circumstantial type, the relationship between the two terms is one of time, place, manner, cause, accompaniment, matter or role," (ibid., page 119) The circumstantial clauses (CC), as applied to the D-object, are highly relevant in describing the (not surprisingly) circumstances of the D-object. In other words, connections between the D-object and the expectations or the environment are predominantly circumstantial. Examples of this are the sentences "The D-object *satisfies* the requirements" (process) and "The D-object *is situated* at the dockside" (attribute). As said earlier, these relations are either composite (attributive - from verbs and prepositions or adjectives) or simple (process) verbs.[c]

The easiest way to visualize the circumstantial relations is to distinguish between what a concept A means 'in isolation ' and what it means with respect to another concept B not included in the definition of A It is possible to place A and B in different circumstances without changing their 'isolated' definition. A circumstantial relation is, however, established when "A is (placed) next to B." Both A and B are internally the same, but their mutual relationships is different - they have a new circumstantial relationship.

A circumstantial clause built from a verb and a preposition differs in one respect significantly from a clause built from a verb and an adjective. Whereas a CC including a preposition will not be applicable to relate the D-object to its <u>intrinsic</u> concepts (see definition in Chapter 4), the CCs including an adjective may be used for this purpose.

---

[c]    When discussing the circumstantial relations as attributive it is clear that they are clauses rather than specific relations. Thus, they are either composed from verbs and prepositions or verbs and adjectives. They are discussed both here, as well as under prepositions and adjective modifiers, because of this mutual influence.

The D-object is by definition a concept that receives meaning only through its relations to secondary (intrinsic) concepts. To claim that the D-object has a circumstantial relation including a preposition to intrinsic concepts is dangerous, since the intrinsic concepts are all related (as 'content) to the D-object. For instance, to say that "The vessel (D-object) is *above* the keel" becomes contradictory since 'the keel ' is involved in defining 'the vessel'.

As it turns out, it is difficult, if not impossible, to construct a meaningful sentence in which the D-object has a circumstantial relationship including a <u>preposition</u> to its intrinsic concepts. At best, such a sentence would be unresolvable, like in "The vessel is connected *to* its rudder." (<u>What</u> is connected to the rudder?) At worst it becomes contradictory, like above. (Is the keel above itself?) Of course, the intrinsic concepts may have such circumstantial relations with other <u>intrinsic</u> concepts, like in "The rudder is connected *to* the autopilot" or "The engine is *above* the keel." There are thus three ways in which circumstantial relationships including a preposition apply to a D-object; either as relationships between intrinsic concepts, between intrinsic and non-intrinsic concepts ("The hull is *in* the water"), or between the D-object and non-intrinsic concepts ("The D-object is *at* the dockside").

When it comes to the CCs including adjectives the picture is different. Generally, the adjective modifiers are applied to <u>compare</u> some <u>aspects</u> of the D-object to its parts. Thus, statements such as "The vessel is *longer* than the hull" are allowable. Roughly, it is possible to say that a D-object may be <u>compared</u> to its parts, but not <u>positioned</u> according to them.

"In the possessive type, the relationship between the two terms is one of ownership; one entity possesses another." (ibid., page 121) The <u>possessive relations</u> describe relations between a concept and concepts 'had ', or possessed, by that concept. This applies to <u>parts</u> serving to define a D-object (semantically, *has-part),* <u>members</u> of a class *(has member, contains, includes),* and <u>characteristics</u> 'had ' by concepts *(has attribute, has value, has property).* In a sense, the possessive relations <u>mirror</u> the intensive relations described earlier - "The vessel *is* a container ship" is analogous to "The class of container ships *has* the vessel (as a member)," and "50 meters *is* the length *of* the vessel" is analogous to "The vessel *has* a length which *has* value (or equals) 50 meters."

The target of possessive relations can generally be referenced by using the preposition *of.* The 'rudder' in "The vessel *has* a rudder" becomes "The rudder *of* the vessel." The 'vessel· in "The class of container ships *has* the vessel (as a member)" becomes "The vessel (which is a member) *of* the class of container ships," The 'length· in "The vessel *has* a length" is referred through "The length *of* the vessel." This will be discussed further in the section on (referential) preposition relations.

### Existential relations

The existential relations "represent something that exists or happens." (ibid., page 130) "There is a D-object" refers to the mere existence of the concept 'D-object' and does not tell anything about what this concept means. It may be viewed as a means of signaling the existence of a concept that either (1) can be used to describe other concepts, (2) must be described by means of other concepts or (3) that is irrelevant.

In the context of the Design Process Language, the first steps to take in the design process include to introduce the sentence "There is a D-object" into design memory, and to interpret the expectations ("There are expectations") in terms of this new concept, like "The D-object should float." If these are the only sentences in design memory, the task would be to relate the D-object in some way to the concept 'float'.

### Modal relations

The design process can be safely assumed to be characterized by some key terms like 'future', 'uncertain', 'planning', 'ability', 'hypothetical', and other terms signaling what is to be, become, or be done. None of the relations introduced so far can adequately deal with such situations. In a sense, the verbs previously discussed are somewhat deterministic in the sense that they are either true or not. There is yet no way to describe an expected, desired or required state as outlined in Chapter 4, for instance "Vessel has length 50 meters". In terms of the prior relations, either the vessel 'has' this length or it 'has not'. Fortunately there are tools that modify these verbs - *modal auxiliary verbs,* here referred to as <u>moda</u>l relations.

In the Design Process Language the modal relations are used for indicating that a relationship is possible, plausible, able, to become true, and the like. Grammatically, a modal auxiliary verb is used for "expressing the mood of a verb," where 'mood' is defined as "a verb form or a set of verb forms inflected to indicate the manner in which the action or state expressed by a verb is viewed with respect to such functions as factuality, possibility, or command." (American Heritage Dictionary) The modal verbs include *can, could, may, might, must, ought, shall, should, will,* and *would.* A modal verb always occurs in conjunction with another verb in the sentence. (Butler, 1988) I will briefly discuss the semantics of each of these terms as applied to a design language.

*Can* is used here as a relation applying exclusively to the *material relations.* In other words, it applies to relations that <u>express transactions,</u> or state-changing processes in the design process. In this interpretation, *can* is a means of expressing ability of performance. For instance, the sentence "The program *can* calculate the number" is legal whereas the sentence "The number *can* equal 7" is illegal.[d] This modal relation is necessary in a language where content and process is to be expressed in the same language and memory. The reason for this is that there must exist means to distinguish between an *ability* to do something and the *mobilization* (or enactment) of this ability. If *can* did not exist in the language it would be difficult to resolve whether something expressed an action or a possible action.[e]

---

[d]     If something *can* something, ii should be possible to ask "What *can* it do?" Obviously 'the number· *can't* do anything while 'the program' *can.*

[8]     To completely resolve the notion of *can* is a quite involved philosophical question. McCarthy (1973) has an extensive discussion on what the term actually implies in terms of artificial intelligence. The discussion by McCarthy goes along the lines: If a program can't do something until told to do it, *can* it do it? Such discussion will not be pursued here.

*Could* is the past tense of *can.* Although it has no central position in DPL one useful Interpretation is that It expresses the same as the composed modal relation *"can* and *might"* below. The sentence "The program *could* calculate the number" then means that the program has both the ability to calculate the number, and that there Is a possibility that this ability will be mobilized in the future. (That is, the conditions for the ability to be enacted are present.) Thus, possibility and ability to perform a certain transaction in the design process may be expressed through this modal relation.

*May* is used for expressing possibility and applies to all other verb relations. In a design situation It would probably be used for introducing a relationship whose truth is not established. There are, however, multiple interpretations, like in "The number *may* equal 7." This could mean either (1) that the number is 'allowed' to equal 7 or (2) that there are indications that the number is 7. For sake of simplicity, the interpretation (I) is chosen here for *may* while (2) is captured through *might,* such as in "The number *might* equal 7." In a sense, this is congruent with traditional interpretations In linguistics where *might* is seen as a 'weak' form of *may* - the latter is permissive in nature while the former is rather informative.t

An additional ambiguity arises when *might* is used in conjunction with a material relation. To say that "The program *might* calculate the number" could, with respect to the interpretation (2) above, mean that either (a) there are indications that the program will calculate the number in the future or (b) there are indications that the program can calculate the number. It would be convenient If it were possible to use double modal relations, such as in "The program *may can* calculate the number" for interpretation (b). It is questionable whether this should be permitted, particularly due to the danger of making absurd sentences, such as "The program *should might can* calculate the number." One way to make such sentences more 'readable' could be to use conjunctive *and* to combine modal relations, like in "The program *may* and *will* calculate the number" for (a) and "The program *might* calculate the number" for (b). This interpretation is adopted here.

---

Contrast "You may do it" with "You might do it.· When *may* is used the statement generally would grant permission - when *might* is used the statement generally would be Informative (of option).

The modal relation *shall* is a quite complex modal relation. Lexically it is defined as follows: "1. Used to indicate simple futurity: ... *(The vessel shall sail the seas).* 2. Used to express: a. Determination or promise: ... *(The job shall be done soon).* b. Inevitability: ... *(The process shall take time).* c. Command: ... *(The designer shall do the job).* d. A directive or requirement.... *(The vessel shall meet the requirements)"* (American Heritage Dictionary, comments in brackets added here) Due to these multiple interpretations and due to the fact that *should, must,* and *will,* seem sufficient for capturing these interpretations, the modal relation *shall* is not used in DPL.[g]

There are two interpretations of the relation *should* that are relevant in the Design Process Language: (a) As expression of probability or expectation, as in "The D-object *should* satisfy the requirements," or (b) as expression of conditionality or contingency, like in *"Should* the result be wrong then the work must be redone." The interpretation chosen here is that *should* expresses an expectation of a future state, such as In (a). The relation particularly relates to the <u>expectations</u> put on the D-object, as discussed in Chapter 4. Interpretation (b) seems to be manageable through the conjunctive relation *if,* and there is no apparent reason for making the definition of *should* ambiguous in order to 'duplicate' the role of *if* in the language.

The modal *will* is of the more multi-faceted such relations. American Heritage Dictionary gives the following definitions: "Used to indicate: 1. Simple futurity: ... *(The vessel will eventually be built.* 2. Likelihood or certainty: ... *(The vessel will be large).* 3. Willingness: ... *(I will design it).* 4. Requirement or command. ... *(You will design it.* 5. Intention: ... *(I will design it if I get paid)* 7. Capacity or ability: ... *(The vessel will float."* (comments in brackets added here) The interpretation used in DPL is one In the intersection of 1, 2, and 4. Will is used to modify the material relations as a <u>transaction that Is to be mobilized</u> (with certainty) at a later stage in the design process, as in "The program *will* calculate the number." It thus has a particular application in <u>executing plans of transactions</u> in the process, as shown below.

---

[g]  It may, however, be argued that interpretation 2.d. is a substitute for the weaker *should,* as discussed below. That would make the expectation stronger, bordering to a competition with *must.*

*Would,* as a past tense of *will,* does not have a central role in DPL. Its only signification might be to conditional actions, in the sense that if certain conditions were present then an action would follow - "If the length were longer than maximum length then the program *would* alter the length." However, such hypothetical events are either captured through *will,* as through conjunctive relations like "If the length is longer than maximum length then the program *will* alter the length."

The modal relation *ought* is a weak modifier, and is scarcely used in DPL. Semantically it is close to *should,* in that it generally indicates obligation or duty, desirability, or probability and likelihood. The definition used in DPL is that it expresses a <u>desire</u> of a future state in the design process, in contrast to *should* that expresses <u>expectations</u> of a future state. It may thus be used for planning purposes, such as in "The vessel *ought* to look nice." (Note that this relation normally is followed by the preposition relation *to.)* For most practical purposes in the design process, *ought* is not a necessary relation.

The last modal relation is *must,* one of the stronger such relations. In DPL it expresses a <u>required</u> <u>transaction</u> <u>or state,</u> without any promise of this transaction or state to actually 'appear' in the future. One general usage in design would be to express rigid constraints, such as in "The hull *must* resist water-pressure" or as "The designer *must* calculate stability." For <u>non-materia</u>l verb relations all *must* relationships have ultimately to be matched with relationships either (a) replacing *must* with *does,* as in "The hull *does* resist water-pressure," or (b) change the infinite tense of the verb to its <u>present</u> tense, as in "The hull *resists* water pressure.[h] As <u>materia</u>l relations are concerned, the easiest way to handle the *must* relationships is probably to delete it as soon as the required transaction is performed, and replace the sentence with either (i) *did,* as in "The designer *did* calculate stability," or (ii) replacing the material verb relation with its <u>past</u> tense, as in "The designer *calculated* stability."

An additional verb relation is useful, if not necessary; the auxiliary (not modal) verb *do.* This relation may be used for expressing either relationships that are true or that have been true at some time. This is illustrated above. However, the *do* relations are not absolutely required -    only helpful in creating meaningful sentences that are also easily understandable. The modal relations in DPL are summarized in Table 5.2.

---

[h]     It is possible that the definition of *will* should be extended to include non-material relations as well, such that it became possible to say that "the hull *will* resist water-pressure" as a promise.

Note that no modal relation expresses absolute certainty of things to be. Even *will* and *must* point to the future, and are thus uncertain. This conspires well with the lack of certainty of the future with which designers are concerned, as has been frequently mentioned. Let me illustrate with a small example:

Initially, there is knowledge of abilities and a conditional (potential) plan:

| Abilities | Sequence | Conditional actions |
|---|---|---|
| *Can (do)* C ... | ? | If (condition) then *should (do)* C ... |
| *Can (do)* A ... | ? | If (condition) then *should (do)* A ... |
| *Can (do)* B ... | ? | If (condition) then *should (do)* B ... |

| Modal relation | Description | Scope |
|---|---|---|
| Can | Ability to perform | Material |
| Could | Ability, possibility, and plausibility to perform | Material |
| May | Possibility | Non-material |
| Might | Possibility and plausibility of relationship being true. Possibility and plausibility of ability' | Non-material Material |
| Shall | Not used | None |
| Should | Expressing expectation of future | Both |
| Will | Certainty of 'execution' (or mobilization) | Material |
| Would | Not used presently | None |
| Ought | Expressing desire for the future | Both |
| Must | Expressing required state Expressing required transaction | Non-material Material |
| (Do) | Modifying tense of verb relation | Both |

**Table 5.2:**      Different modal relations in DPL.

Upon determining that the "condition is true," and making a sequence of enactment, there is a concrete plan to follow in <u>mobilizing those abilities</u> in the future:

| Abilities | Sequence | Action to perform |
|---|---|---|
| *Can (do)* C ... | 3 | *Will (do)* C ... |
| *Can (do)* A ... |  | *Will (do)* A ... |
| *Can (do)* B ... | 2 | *Will (do)* B ... |

The <u>actual execution</u> of the ordered plan (mobilization) could be represented as "Do A then B then C":

| Abilities | Sequence | Plan in-execution - mobilization |
|---|---|---|
| *Can (do)* A ... |  | *(Did)* A ... {Transaction already undertaken} |
|  | (then) |  |
| *Can (do)* B ... | 2 | *(Doing)* B ... {Transaction being undertaken} |
|  | (then) |  |
| *Can (do)* C ... | 3 | *Will (do)* C {Transaction still to be undertaken} |

The auxiliary verb *do* signals what has been done, what is currently being done, and what remains to be done.

## Preposition relations

In describing design process and object, it is necessary to be able to represent how concepts that may <u>exist</u> independently are placed together, or configured, either to perform a certain task (as in a plan) or meet a certain functional requirement (as in a geometrical layout of a structure). In other words, concepts may be treated as 'blocks' that are related to each other in space, time, or 'cause·. The blocks themselves may. in turn, again consist .of more 'elementary' blocks that are related. For instance, subsystems of a vessel such as machinery, cargo holds, and fuel tanks, may be moved without any change to these subsystems' internal description. Moving an engine does not alter the relationship between the cylinder head and the cylinder rod, and using a sub-routine at a new location in a process does not change the relationships between that subroutine's 'own' subroutines.

The relations expressing *temporal* or *spatial* relationships would be expected to be absolutely required in the Design Process Language. It is frequently necessary to express such relationships through using <u>preposition relations,</u> as in "The machinery is *on* deck 3" and "Plan A will be executed *before* plan B." In addition, relations expressing more 'causal ' connections may be encountered frequently - for instance in modelling event-consequence scenarios in technical systems, where what is important is interactions of causal nature rather than interactions of temporal or spatial nature.

One instance of the use of such relations are in structural design. A structure is determined as much from relationships among structural members as the 'content' of the members themselves. Indeed, it commonly is the case in structural design that such relationships are determined <u>before</u> the content of the members.

The use of preposition relations is closely related to the use of circumstance verb relations as described earlier. A preposition relation will never act without some kind of verb relation, notably the circumstance verb relations. In some sense, one can say that the preposition relations are <u>directing</u> while the verb relations are <u>enacting.</u>

Grammatically, a preposition is defined as "... a word (or word construction) that indicates the relation of a substantive to a verb, an adjective, or another substantive." (American Heritage Dictionary) Said in another way, they are relations that apply between concepts, indicating how concepts relate to each other in terms of particularly space and time. Examples of preposition relations, henceforth referred to as PRs, are; *In, Out, Inside, Around, Outside, From, To, After, Before, Above, Below, With, Without, Of, For, Between, At, By, Within, In-between, Next, Beside, Previous, Under,* and *Over.* Examples of typical relationships including circumstance verb/preposition relations are: "The crane is *on* the deck." "The crane moves the cargo *from* the deck *to* the shore," and "Subroutine 'Read' executes *before* subroutine 'Display'."

It is particular of the preposition relations that they have various characteristics some of which 'belong ' to one particular PR and some of which relate one PR to other PRs. Thus, the semantics of the relations can be defined in terms of how they compare to other similar relations. One would expect that knowing the semantics of prepositions would help humans in spatial and temporal (and to some extent causal) reasoning.

Some aspects of the PRs are helpful to classify the by semantics. The classes are introduced here as *transitive, reversitive, inversitive, synonymous, implicating,* and *referential.* Formal definitions of each term in these classes are not given here - the semantics is generally understood. It should however be realized that whether some PR has a particular characteristic depends on how that PR is defined. For instance, if *on* is defined as a <u>temporal</u> PR then it wouldn't be classified as <u>implicating</u> with *above.* The purpose here is thus to suggest the semantics of PRs in general and to classes of PRs -  not the particular meanings of individual PRs.

<div align="center">**Transitive**</div>

Transitive PRs have the characteristic as illustrated in: *If "A  is PR B" and  "B is PR* C" *then "A is PR C."* Examples are the PRs *Above, Below, Before, After, Inside, Around, Within, With, Of, Under,* and  *Over.* All PRs that have inverses, as described below, are transitive.

*Transitivity* is of particular use to design in two respects; understanding how two concepts relate in space and time although they are not explicitly related initially. For instance, in a spatial design object one may know that if "Deck two is *above* deck one" and "Deck three is *above* deck two." From the semantics of *above* it should be possible to find that "Deck three is *above* deck one." If developing a plan of actions to pursue in the design process it might be necessary to know that if "Action A is executed *before* action B" and  "Action B is executed *before* action C" then "Action A is executed *before* action C."

### Reversive

These are individual PRs that have the characteristic as illustrated in: *If* "A *is PR B" then 'B is PR* A." Examples are the PRs *Next, Beside,* and (possibly) *Outside.*

*Reversitivity* enables reasoning about both concepts related through the PR. A typical example would be the relationship "The engine room is *beside* the fuel tank" where it could then be found that "The fuel tank is *beside* the engine iOOm."

### Inversive

These are pairs of PRs that have the characteristic as illustrated in: *If* "A *is PR, B" then "Bis PR$_2$* A" *and  If* "A *is PR$_2$ B" then "B is PR,*

| PR,: | *Below* | PR$_2$: | *Above, Over* |
| PR,: | *Before* | PR$_2$: | *After* |
| PR,: | *Inside* | · PR$_2$: | *Around* |
| PR$_1$: | *Under* | PR$_2$: | *Above, Over* |
| PR,: | *Within* | PR$_2$: | *Around* |

All of this kind are also transitive. Typical use of such inversitivity is to find that if "Deck one is *below* deck two" then "Deck two is *above* deck one." This semantics would enable the system to answer to command like "Calculate weights *above* deck one" although all relationships might be stated with the *below* relation. Equivalently, it would not be necessary to question plans in terms of relationships involving *after,* if two concepts are already related by *before.*

### Synonymous

These are pairs of PRs that have the characteristic as illustrated in: *If* "A *is PR, B" then* "A *is PR$_2$ B" and  If "Bis PR$_2$* A"  *then "Bis PR,* A."

| PR,: | *Below* | PR$_2$: | *Under* |
| PR,: | *Above* | PR$_2$: | *Over* |

The use should be apparent. Synonymocity makes unnecessary a dual representation even though two different transactions are based on different terminology. If "Deck one is *below* deck two" then "Deck one is *under* deck two" and vice versa.

### Implicit

These are  pairs of PRs that have the characteristic as illustrated in: *If "A is PR, B" then "A is PR$_2$ B"* but not <u>necessarily</u> *If "A is PR$_2$ B" then "A is PR, B"*

R$_1$:      *On*              PR$_2$:    *Above, Over*

As other aspects, this leaves unnecessary  dual representation with the relation *on* to  also support interpretations that.focus on other relations.  If something is *on* something it is also *above* it. (Of course subject to the definition of *on.)* Thus, if "Engine  is  *on* the floor" then "The engine is  *above* the floor," but not necessarily vice versa.

### Referential preposition relations

Some of the PRs do  not  necessarily describe spatial, temporal, or causal relations. One particular category may be termed *referential* PRs, including *of, by, to,* and *as.* These PRs are  useful relations in the language, and *of* is particularly frequent in existing knowledge representation languages as in *part-of, instance-of,  subclass-of,* and  *member-of.*

The PR *of* is used in DPL to  refer back to a <u>possessive</u> relationship. Generally it can  be  said  that if "A has B" then B can be  referred to as (i) "B *of* A." This reference can  be  performed through referring to what aspect of A that B describes  -  whether  it  is  'part·,  'member',  "subclass',  'instance', 'attribute', "value', or "property' as discussed in the section "Abstract concepts" below. A relationship might look like "A has? B" then (ii) "Bis? *of* A," where '? ' can be any of the above terms_! Consider the difference of (i) and (ii). In the relationship "Vessel has subsystem machinery" the latter term can be referred

---

As will be argued later, these 'abstract concepts· are not necessary in DPL. However, since they are very frequently used in daily speaking and in design terminology, they are discussed here.

to as (A) "Machinery *of* vessel" or as (B) "Machinery is subsystem *of* vessel." The statement (A) is more of a <u>reference</u> to 'machinery· whereas (B) is more of a <u>definition</u> of 'machinery'. The *of* relation may be considered as a transitive PR as defined above. Consider "A is subclass *of* B" and "B is subclass *of* C" which imply that "A is subclass *of C"*.

The PR *by* has another function in terms of reference. This PR is frequently used to relate a concept to something which has performed transactions on that concept like in "The program calculates the formuia" which implies that "The formula is calculated *by* the program." In general terms, it can be said that if "A MR B" then "B Is MR'ed *by* A," where MR stands for a <u>material relation</u> as described earlier. The PR can also be used in conjunction with other transitive verbs, like In "The weld connects the two beams" which becomes "The two beams is connected by the weld." However, this is not a rule - consider "The vessel has machinery." It becomes queer to claim that "The machinery is had by the vessel," even though *have* is a transitive verb.

The preposition relation *to* is often used in conjunction with an adjective, as in *adjacent to, next to* and *previous to,·* or with verbs other than· *is,* as in *is connected/connects to* - "The propeller is connected *to* the propeller shaft." It is difficult to resolve the *to* preposition relation due to this characteristic - the semantics is more tied to the verb or adjective relation than to the PR. However, there may exist a general rule for the case of <u>reversitive</u> relations, such that If "A is VR/ADJ *to* B" then "Bis VR/ADJ *to* A," where ADJ stands for adjective modifier (described in a later section) and VR stands for a verb relation - most relations including verbs or adjectives and *to* can be reversed. Examples of reversitive VRs are *Connected* and *Linked,* and of reversitive ADJs are *Adjacent, Close, Near,* and *Opposite.*

However, other verb relations, apparently including all MRs, have an opposite interpretation. Consider *Moved* and *Sent,* where the rule is false, as is seen in "The cargo is moved *to* the shore" which then would become "The shore is moved *to* the cargo." This is because the MRs are transitive verbs, where the agent (subject) affects an object, and where the *to* phrase constitutes a complement - in this case the direction of the transaction.

The last referential PR discussed here is *as*. It has an interesting property when used to reference <u>a particular context.</u> Consider, "The vessel *as* a transport device" which signals what aspect of the vessel that is interesting in the actual context. Depending on the definition of a 'transport device', the vessel would 'inherit' its characteristics, such as its capacity, the kind of cargo, its speed, and so forth. "The vessel as a structure" would probably cause quite different characteristics to be focused, such as in the form of structural members, their dimensions, and their geometric configuration. It is then possible to state "The vessel is analyzed *as* a structure."

As was discussed in Chapter 3 under Object Oriented Programming, an object may inherit properties from the class to which it belongs. However, since there is an infinity of classes to which one object may be assigned, it may be appropriate to reference its particular class membership through *as*. This PR becomes more important when a class is more seen as a particular concept that is used to define other concepts. In that case, there is not necessarily a class-subclass hierarchy, and the way to point to particular characteristics commonly defined through a *member-of* relation becomes to point to particular concept through an *as* relation. This will be discussed further under the section 'Abstract concepts'.

## Conjunctive relations

The conjunctive relation is used to connect other complete propositions or concepts within a proposition. A conjunctive is formally defined as a word or word construction that is " ... serving to connect elements of meaning and construction in a sentence." (American Heritage Dictionary) Examples of conjunctive relations (CRs) are; *With, While, Both, And, Either, Or, Neither, Nor, Also, As, Than, Ergo, Therefore, When, Although, Since, Because, Where,* and *Whether.*

There are several reasons why the conjunctive relations are absolutely needed in a Design Process Language. One of the reasons is that design involves frequent use of decision rules, which are efficiently represented through using the conjunctive relation (CR) *if,* particularly in 'companionship· with the adverb relation *then.* Another reason is the need to represent 'logical' relationships among concepts or other relationships, traditionally as carried by the CRs *and* and *or,* but possibly including *either, neither, both,* and *nor.* Another reason is that it might be necessary to relate a cause with its effect or a condition with an effect *(if-then,* or only *if),* or to express logical properties of these conditions or effects, or to represent alternatives *(or).*

---

"The effective procedure for propositional reasoning embodies one fundamental principle that I shall argue governs the psychology of reasoning: the logical properties of connectives *(conjunctive relations)* derive from their meaning...........The solution to the problem of *if* is to search, not for a single uniform logic of the terms, but for a single uniform semantics from which its logical vagaries emerge. Such an analysis will strengthen ... the case for the fundamental semantic principle that the meaning of a term gives rise to its logical properties." (Johnson-Laird, pages 54 and 56, comment in bracket added here)

---

I will discuss the conjunctive relationships in three categories; <u>inclusive</u> (conjunction), <u>exclusive</u> (disjunction), and <u>causal</u>. There may be other categories of interest, but the three proposed here seem most relevant in a design context. Some conjunctions apply to connect other complete sentences, such as *because, if, since,* and *while.* Others may also be applied to connect <u>individual concepts,</u> such as *and* ("The ship is red and blue"), *both* ("The ship is *both* red and blue"), and *or* ("The ship is red *or* blue").

### Inclusive

In this category are those conjunctive relations where the two relationships joined by the CR are both true. These are statements of the kind *"A CR B"* such that A and B are both true sentences to concepts. Among these are the CRs *and, both,* and *also.* Examples of their use is as follows:

"If the length *and also* the beam are given then the L/B ratio may be found"
"The L/B ratio may be calculated if *both* length *and* beam are given"

### Disjunctive/exclusive

In this category are those conjunctive relations where at least one of concepts or sentences joined by the CR may be false. These are statements of the kind *"A CR B"* where A or B or both may be false. Among these are the CRs *either, or, neither,* and *nor.*

Intuitively, there are three such expressions that may be of particular use; either (i) all sentences or concepts are false or (ii) one, and only one, is true (exclusive *or)* or (iii) all may be true: (A <u>true</u> concept is one that is a correct concept to link to the 'target' concept. For instance, in "Color is red or blue,. Here, either 'red ' or 'blue ' is the concept to relate to 'color' - the <u>true</u> concept.)

Sentences like in case (i) uses two particular conjunctive relations; *neither* and *nor.* They should be quite self-explanatory; "The color is *neither* red *nor* blue."

For sentences as in case (ii) it is more difficult to find unique relations. However, it is probable that the conjunctive *either,* operating with *or,* is a natural relation to use for this purpose. "The color is *either* red *or* blue."

For sentences as in case (iii) the ordinary *or* relation will do. This does not say anything about exclusiveness; the statement is true even if both A and B are true. Consider; "If the displacement is lower than required displacement then I should alter length *or* beam *or* draught (or all three)."

### Causal

Intuitively, there are two kinds of causal conjunctives; retrospective (predominantly historical) and hypothetical (future). In this category are those conjunctive relationships where one relationship in some sense describes or causes the other relationships joined by the CR to be true. Thus, *"A CR B"* means that either (i) B is true if A is true, (ii) A is true if B is true," or (iii) A is true despite the fact that B is false.

Among case (i) CRs are *therefore,* and *ergo.* "The maximum length is 140 meters *therefore* length is too long."

Among CRs of type (ii) are *with, while, as, because,* and *if,* "Length must be reduced *if* it is longer than maximum length," "Length changes *with* displacement," "Length changes *as* displacement changes," and "The crane is on the deck *because* it should reach *(both)* cargo hold and shore.

Among case (iii) CRs are *although* and *however,* as in "Length must be reduced *although* displacement is to be constant" and "Displacement is to be constant *however* the length must be reduced."

## Adjective modifiers

The adjectives, as well as the adverbs, are more *modifiers* than relations. They are discussed here, however, since they closely proximate the relations. The adjective modifiers are similar due to the way they are used in DPL - in comparisons (see below), while the adverbs modify verbs that themselves are relations - 'following · the verbs into these verbs ' roles as relations.

The adjective modifiers are generally terms that apply in comparison to other terms. Lexically an adjective is defined as "(grammatically) any of a class of words used to modify a noun or other substantive by limiting, qualifying, or specifying." (American Heritage Dictionary) In general, the adjectives thus do not assign any objective value to the concept to which they apply - they assign a property valuable only in terms of some lower level values. For instance, the truth of the sentences "The vessel is *large*" or "The program calculates *fast'* depends on what the adjectives *large* and *fast* imply.

The most common usage of these modifiers are when concepts are compared to other 'equivalent' concepts, such as in "Vessel A is *larger* than vessel B" and "Program A calculates *faster* than Program B." However, it is typical of several adjectives that the interpretation is relative as well -  does *large* mean heavy, voluminous, lengthy, or what? One can always 'probe' a description of the kind "The vessel is *large*" with "How *large?",* whereas it normally does not make sense to probe quantified attributes in this way.

One important exception to this relativity are numbers, which actually function as adjectives in some respects, like in "The vessel has length 140 meters." Here, 'meters' is a (semantic) concept while '140' is an adjective that modifies 'meters' but which Is not relative. The same information could be stated as ."The vessel has length," "Length (of vessel) equals 140." and "Unit (of length) is meter." In this case, '140' is grammatically a noun, or a proper concept in the DPL.

It is important to distinguish when the number plays the role of a noun and when it plays the role of an adjective, particularly since the grammatical rules discussed later on separate between the roles of adjectives and the nouns (which in DPL are the concepts). Some other adjectives display similar behavior as the numbers, notably those adjectives that have their 'origin' from nouns, such as terms designating names of colors - 'red', 'blue', and so forth are both nouns and adjectives. The discussion below concerns those adjectives not originating from nouns.

In design, most adjective modifiers will be used in comparisons, or <u>circumstance verb relations</u>, rather than in describing a concept 'in isolation'. One reason is seen in the following example; if (i) "Vessel A has length 140 meters" and (ii) "Vessel B has length 130 meters" it could be concluded that "Vessel A is *longer* than vessel B," whereas stating that "Vessel A is *long*" and "Vessel B is *long*" wouldn't tell much in their comparison as regards length. Adjective modifiers could be used to state that "Vessel A is 140 meters *long*" and "Vessel B is 130 meters *long.*" Here, values are 'given to' the adjectives and 'long' assumes the same role as the attribute 'length'. Another way of 'quantifying' the adjective/attribute could be through using adverbs like *somewhat, slightly, significantly, very, extremely,* and so forth. As will be argued in the next section, this might be useful in a system supporting 'fuzzy' knowledge, but will not be used in DPL.

To the extent that adjective modifiers are used in DPL they will refer to attributes - not to values of these attributes (or actually complete properties - attribute/Value pairs). Thus, just as the attributes tell something about the concept to which they belong, leaving some relationships to be constructed between the attribute and its value, so may adjectives be valued. However, when describing a concept as in Chapter 4, through relationships, the adjectives will be avoided. Instead will be used a noun functioning as an attribute to capture the adjective, as in "has length" for 'long' and "has weight" for 'heavy· or 'weighty'. This removes some of the 'relative' nature of such statements.

The adjective modifiers <u>are</u> used in DPL when the relationship indicates an <u>existing or possible comparison</u> among concepts that share the same kind of attribute, as in (i) and (ii) above. This is used in (1) stating inequality criteria for selection, as in "Choose the *cheapest* vessel," (2) comparing alternatives in terms of inequality, as in "Vessel A is *cheaper* than vessel B," and (3) identifying one concept as the answer to an inequality criterion, as in "Vessel A is the *cheapest."*

Two exceptions to the relativity discussed above, aside from those adjectives originating from nouns, are the adjectives *false* and *true.* These adjectives are absolute in the sense that for most practical purposes in design there is no interpretation that makes *true* become 'less true'. If anything, if something is not *true* it is *false* - if a sentence becomes "untrue' it becomes false.

Typically, the adjective relations deal with relationships in terms of value, space, time, and certainty. Incidentally, this corresponds to several researchers' view of what characterizes objects. (See for instance Guilford, 1968 and Baron, 1987)

The adjective modifiers share certain properties with the preposition relations, possibly because they are "comparative ', which is to say that they bear meaning only with relation to some other concepts and a verb relation. Thus, as with the prepositions, they are possible to categorize according to some main characteristics, although the categorization according to these groups is not as efficient as for the preposition relations.

### Transitive

These individual ADJs have the characteristic as illustrated in: *If "A is ADJ PR B" and "Bis ADJ PRC" then "A is ADJ PRC."* Examples are the ADJ-PR pairs involving the preposition *than,* as in *taller than, larger than, smaller than,* and *less than.*

### Reversive

These are individual ADJs that have the characteristic as illustrated in: *If "A is ADJ PR B" then, "B is ADJ PR* A." Examples are the ADJ-PR pairs involving the preposition *to,* as in *close to, next to,* and *adjacent to.* The same applies with the preposition *as,* such as in *as low as, as good as,* and *as large as.*

**Inversive**

These are pairs of PRs that have the characteristic as illustrated in: *If "A is ADJ, PR B" then "B is $ADJ_2$ PR A" and If "A is $ADJ_2$ PR B" then "B is ADJ, PR A."*

| $ADJ_1$: | *Better* | $ADJ_2$: *Worse* |
| $ADJ_1$: | *Higher* | $ADJ_2$: *Lower* |
| $ADJ_1$: | *Longer* | $ADJ_2$: *Shorter* |
| $ADJ_1$: | *Heavier* | $ADJ_2$: *Lighter* |
| $ADJ_1$: | *Closer* | $ADJ_2$: *Farther* |

## Adverb modifiers

The adverbs are words that modify verbs, adjectives, or other adverbs. The most common usage is as a modifier to verbs. In design, one would not expect the adverbs to be too frequent for the same reasons as with the adjectives - they are relative. They may, however, be of use in modifying verb relations in the DPL according to certainty, as in "Vessel A is *possibly* cheaper than vessel B." indicating that the relationships must be resolved in the process to either remove *possibly* or change the relationship. It is conceivable also that an extensive analysis and subsequent use of adverbs, introducing levels of certainty as *certainly, probably, possibly, plausibly, viably, likely, uncertainly, unlikely, hardly, improbably, impossibly,* and so forth, could provide a useful terminology for representing 'soft' information regarding certainty. Some have advocated that an 'intelligent' system must be able to cope with such 'fuzzy' terminology. (See, for instance, Minsky, 1991)

An exception to the avoidance of adverbs is the adverb *not.* This term is omnipresent in both computing languages, as a 'negation operator', in logical problem solving systems, and in standard design terminology. It is a 'substitute' for the use of the adjectives *false* and *true,* and is necessary due to the fact that it is almost as important to say what something isn't as what it is. One might, of course, simply leave sentences unstated and treat them as false until they are explicitly established, as in non-monotonic reasoning. (Genesereth and Nilsson, 1988) However, it is often necessary to actively use the *not* adverb, as in "The displacement is *not* larger than weight," causing a particular (active) focus on the displacement. The alternative would be to leave the sentence out and let the designer imply any not-relationships that must be handled.

## WH- terms

WH- terms are particularly used in questions. These include *who, what, which, where, why, when, how,* and so forth. (See Halliday, 1985, pp. 83-85)

The WH- terms are used here as interrogative - as applied to questions. They typically occur in the beginning of a sentence and demand an answer. *"Who* calculated the number" will be answered as "Program A" if the proposition "Program A calculated the number" or "The number was calculated by Program A" resides in memory. *"Why* is length 140 meters" will trigger a search for a conjunctive relation such as *because, since, therefore,* and so on, that related one sentence A with another sentence (B) "Length is 140 meters" like in "A *therefore* B" or "B *since* A" or "B *because* A." The sentence "What is length" is ambiguous (it could ask for whether length is an attribute or a part, or what length *is* in value). Assuming the query is for a value, the answer would be "140 meters" upon locating (B) above. Similar behavior can be 'coded' as rules into the Design Process Language to produce correct mapping into memory for different WH- clauses.

One can of course use these terms for planning design, as in "What (who) can calculate the number" which returns "Program A (and ...)" assuming "Program A can calculate the number" is found.

## Some issues on general grammar

The issue of grammar is an immensely difficult question to resolve in general, since there is a large number of rules that interact in so many different ways. It is far beyond the scope of this thesis to treat the issue of grammar - I have pointed to categories of relations relevant in the design context upon which could be expected that the grammatical rules should operate. I have thus no intention of delving deeply into this question, but give some indications as to the kind of grammar needed to understand sentences, or relationships among concepts.

The set of grammatical rules needed to resolve sentences in the Design Process Language is significantly less complex than the set of rules needed to resolve sentences in a natural language. The reason for this is that the relations, which is the prime object of resolution by the grammatical rules, are fewer and more rigidly defined in DPL than in ordinary language. The issue of grammar is, of course, most important when the DPL is conceived as <u>applied in a computer environment.</u>

There are two issues that are of particular interest when discussing grammar in the context of DPL the issue of whether (i) a statement is a true/false proposition, a question, or a command, and (ii) a statement is in past, present, or future, and how to change the tense of statements as the design process evolves.

A predominant view within linguistics is that sentences can be analyzed through parsing them into particular phrases, each of which can then be further broken down into new phrases. The grammatical rules, in this conception, operates then upon such a *phrase-structure* as a *context free* construct. Major phrases are noun phrases (NPs), verb phrases (VPs) which in turn are decomposed into more 'fundamental' phrases, such as nominal phrase (NOM) and preposition phrase (PP) and categories as nouns, adjectives, pronouns, verbs, and the like. In phrase-structure, the sentence "The length of the vessel equals 140 meters" could be parsed as follows:

NP: "The length of the vessel"
      Determiner: 'The'
      NOM: "length of the vessel"
           Noun: 'length'
           Relative clause - PP: "of the vessel"
                Preposition: 'of'
                NP: "the vessel"
                    Determiner: 'the'
                    Noun: 'vessel'
VP: "equals 140 meters"

      Verb: 'equals'
      NP: "140 meters"

Adjective: '140'
Noun: 'meters'

Chomsky developed *transformational* rules to interpret sentences. These rules had various roles in the grammar, some for transforming present to past, while others used for transforming active to passive. For instance, with respect to the discussion on the referential preposition relation *by,* one could get the following result from the sentence "The Program calculated the number":

$NP_1$ - VP-> $NP_1$ - V - $NP_2$ is transformed to passive through:
$NP_2$ - was - V – *by* + $NP_1$ {Would predominantly apply to <u>material relations.</u>}

$NP_1$: "The Program"
VP: "calculated the number"
    Verb: 'calculated '
    $NP_2$: "the number"

To passive:

$NP_2$: "The number"
'was'
VP: 'calculated '
'by'
$NP_1$: "the Program"

This results in the phrase "The number was calculated by the Program."

There is a large number of rules in transformational grammar, each having a unique role in rephrasing sentences. Most of these rules operate upon the relations, as cast in this chapter. (See Johnson-Laird for an extensive discussion.)

Among the rules proposed in the transformational grammar are those that have to do with <u>questions,</u> specifically WH-clauses. These clauses, determined by a WH- pronoun as discussed earlier, are important in asking questions of what is in stored in memory. Thus, "What is the length of vessel" would trigger a transformational rule that resolved the question and matched this with an existing proposition. "Length of vessel is 140 meters" or "140 meters is the length of the vessel." "What *can* calculate the number" could be matched with "The Program *can* calculate the number" and "What calculated the number" with "The Program calculated the number."

When the system must resolve <u>commands,</u> such as "Calculate the number," the picture is different. Which ability (which procedure) should be enacted to perform this activity? In this situation, one would expect that the command would primarily mobilize an ability that was planned ahead in the design process by using a <u>modal relation</u>, like in "Program A *will* calculate the number." A rule resolving a sentence with a <u>material relation</u> *(calculate)* and a <u>missing subject</u> could then be to search memory for the modal relation *will* in conjunction with the desired VP "calculate the number" and distinguish the missing subject as 'program A·. Such a rule could then cause the construction of "Program A calculate the number."

If a *will* relation is not found with the VP, one would expect the rule to query the designer (or another authority) something like *"What* will calculate the number?" Such rules are, like most other issues of formalized grammar, only relevant in case of automated implementation of DPL - not when DPL is to be used as a basis for describing a design process as viewed from an outside observer. In the latter case, one focus more on what the particular relations imply than how they are identified and resolved in an automated system.

Johnson-Laird distinguishes procedures that a system may have to ensure a correct interpretation of propositions, commands, and questions. Several of these propositions have to do with verifying that a new proposition does not conflict with existing propositions. Such verification must be closely related to the resolution of relationships. If the proposition "The length must be larger than 140 meters" exists, and a proposition is suggested like "The length is 120 meters," the procedures (semantics) of the system would resolve a conflict through "The length is *not* larger than 140 meters" and disallow that proposition to be accepted. This goes back to the discussions in Chapter 4; one should establish relationships to prove or disprove hypotheses. In this case, the <u>hypothesis</u> that "Length is larger than 140 meters" would be disproved by the <u>'fact'</u> that "Length is 120 meters" and the latter statement would either be rejected or else the conflict would have to be mediated by an authority.

## Abstract concepts

Among the terms that have become usual in design terminology as well as in dally speaking are those that point to 'knowledge about knowledge', This is to say that they refer to some other concept through describing the role that this concept plays in the knowledge representation system. They always presuppose the existence of another concept. The most common terms are *system, subsystem, part, class, subclass, member, instance, property, attribute,* and *value.* Sometimes the abstract concepts *assembly (as* a sort of subsystem) and *element* (as a sort of part) are included among such concepts Also the mere concept of relation is often included among such abstract concepts. (See, for instance, Guilford, 1967)

It is important to note that the terms above do not refer to empirical truth, as in "The vessel is a system" in all situations, but may be more appropriately viewed as context specific. The terms are more pragmatic means for human (and automated) Information processing systems to organize knowledge so as to use this knowledge efficiently in particular situations. For instance, in one Interpretation a vessel may Indeed be a system, as in a ship design situation, while in other interpretations the same vessel may be a part of a system, as in transport systems design. 'Vessels' may also be classes of objects. In some situations 'Volvo 640' may be an instantiation upon a class (for Instance, if I were to choose a new car - "Which mark should I choose?"), and in others it may be a class Itself (for instance, If the Volvo factory were to categorize this line of cars according to 'color' - "How many red 640s did we send out last month?").

Thus, the terms are helpful aids In categorizing concepts and referring to them in particular situation, and whether one·concept Is called a system or a part or a class or an Instance is relative to the situation. I will discuss the terms under three headings; composition and aggregation, classification and abstraction. and attribution.

## Composition and aggregation

To call a concept S a *system* generally implies that it is possible to perform transactions on S that affects the circumstances (extrinsic description) of S while not affecting the (intrinsic) description of S. For instance, it is possible to *move* S without altering the relationships among the parts that make up S. Further, to put the label, or relate the concept, "system' on S implies that there are more fundamental concepts through which S is described.

It is easy to be led into thinking that a representation of a system remains internally consistent as transactions are performed on that system. This is not necessarily true - consider a representation of an engine as (l) "Engine has engine block," (2) "Engine has carburetor," and (3) "Carburetor is *above* engine block." If the engine for some reason is flipped over ("Rotate engine vertically 180°"), the relationship (3) ceases to be true, whereas a 'move' transaction may preserve the internal consistency of the system.[j] Thus, care must be taken in treating a system as a block that remains internally unaffected by the context in which it is placed.

The most important role of relating the 'system· concept to S is to refer to a whole that 'is more· than the sum of its components. In other words, the system exhibits characteristics - has functions - that results from a particular combination of components. The system S is a concept that becomes defined through components serving to both define S directly and that attribute to it some characteristics that neither of its components have in separation.

The components of a system are commonly referred to as *parts* or *elements.* Traditionally, these are considered as some kinds of 'atoms' and are treated as building blocks. In relation to the discussion in Chapter 4, the parts are intrinsic concepts used to describe concepts that in turn are used in describing the system to which the parts belong. In the end, the parts will serve to describe extrinsic concepts of the system S, or serve to enable S to meet the expectation put to it. In this concept formation process both the individual characteristics of the parts themselves and their mutual relationships are important.

---

[j] This depends of course on how the relation *above* is defined, whether it is local or global. In a local definition, where direction is given with reference to a part of the system, S might remain internally consistent.

The parts themselves are, of course, often systems in another interpretation or context. In the conceptual design of a vessel, like in lay-out of general arrangement (GA), the 'engine' is normally treated as a part. As the design process evolves, the 'same' engine may change status to become a set of parts combined into the system, engine. It is questionable whether it is possible to find 'parts' that cannot be considered systems in other contexts - the search for truly elementary particles will probably never end. To denote 'intermediate' systems - systems that assume the role of 'parts' in some contexts - the terms sub-system and assembly are introduced. There is little to say about these except that the terminology implies that there is a collection of parts that constitute a coherent whole - the sub-system or assembly has some distinguishable characteristics that differ from the sum of the characteristics of the individual parts.

### Classification and abstraction

A *class* is a term that designates that the concept in question refers to a collection of other concepts, their only necessary (mutual) relationship being that they share some characteristics outlined through the class definition. A class is traditionally not considered a concept existing in the 'real' world - it is rather an abstraction of a set of such concepts. However, it is an often-overlooked fact that a class may be defined so as to include only one particular concept. In this conception, an instance may then be seen as a class containing only one member, where the class definition is such that it identifies a concept by either (I) a unique characteristic, such as an identifier, or (2) having a class description that is identical to the description of the concept it is to include. Situation (I) would be commonly referred to as *instantiation,* whereas situation (2) would be referred to as *class reduction.*

For the most parts the concepts used in a D-object description are similar to class-descriptions. It is rare to identify an instance by a unique identity - it is more frequent to identify a concept by some particular characteristics that may significantly reduce the members of the class. Identifying a particular kind (by name and type) of gas turbine to use in a ship is often seen as instantiating that gas turbine, but strictly speaking it is rather a (dramatic as it may be) restricting class description that reduces its membership to include only one gas turbine. For example, while "Volvo 640" may seem like an instantiation on the class "Cars to buy" it is instead a class membership criterion by name-kind-type identity. A truly instantiated class would be a class whose description identified a member identical to one, and only one, entity. Strictly speaking, if it were possible to identify more than one 'real' entity that fitted the description, the class would not be instantiated and there would exist multiple possible solutions.

In this view it is not entirely correct to say that the designer has chosen one unique instance to use further in the design process. She has either chosen a class description with which to work or chosen a concept description that meets the requirements of the class description.

It is possible to establish an infinity of classes that contains a particular concept. The requirement is that the description of the class must subsume that of the concept - the concept must satisfy the classification criteria. The ·taxonomy' may have different characteristics. It may be hierarchical, in that the classes are divided such that one concept belongs to only one 'branch·. This is a dichotomous classification - either a *car* is a vehicle or not. The taxonomy may be heterarchical, such that one concept belongs to two classes that themselves are not contained in each other. For instance, a *car* may be a concept contained in the class of *killing_devices* as well as in *vehicles.* This is the problematic issue of 'multiple inheritance', as was discussed in Chapter 3.

As was intimated in the discussion of the preposition relation *as,* it may be more proper to inherit properties from a 'class' concept through viewing the (A) "Vessel *as* a transport device" or (B) "Vessel *as* a structure." The important point is not necessarily to know the class-membership of the vessel - it is more to know the characteristics that are focused when the vessel is viewed as belonging to one class or the other. In practice, this would make the designer able to "Analyze vessel *as* a structure" and "Analyze vessel *as* a transport device." In designing that vessel, one might refer through *as* to find what attributes were to be assigned values. In (A) one might find that an unvalued attribute were 'capacity' while in (B) one might filter out parts of the underlying description, such as 'loading conditions' and a decomposition into structural components and values assigned to the attributes of these components. The description residing in memory is the same and no classes need to be defined. The relation *as* signals what kind of class concept (structure or transport device) should be used as a 'query' condition.

<div align="center">**Attribution**</div>

Hitherto the term *attribute* has been frequently used. Lexically, an attribute is defined as "the quality or characteristic of a ... thing." (American Heritage Dictionary) A concept that falls into this category describes a particular characteristic of another concept. A 'rule of thumb· could be to say that if something can be assigned a particular value, it is an attribute of another concept, like in 'length', 'weight', 'color', and so forth.

It is always possible to assign a *value* to an attribute. Sets of attributes and values are commonly referred to as *properties.* For instance, "Vessel has length" (where 'length' is an attribute), "Length (of vessel) is 140 meters" (where '140 meters' is a value) and "Vessel has length of 140 meters" (where "length of 140 meters" is a property).

In most situations it is easy to see whether something Is an attribute without putting that specific label on it. There are, however, situations where the distinction between an attribute and Its value is diffuse. Consider "The plate has material" and "Material (of the plate) is steel." Here, 'steel' seems to be a value, but the problem is that 'steel' itself has properties, like "Steel has density" and "Density (of steel) is 7.2 g/cm$^3$.· This is as opposed to "Thickness (of plate) equals 2.5 cm" which is clearly a particular characteristic of the plate where '2.5 cm' is a 'true' value. Coyne et. al. handle this issue by saying that "we can treat properties (attribute and value pairs) as objects (and) depict the 'properties' of properties." (Coyne et. al., 1991, page 93)

<div align="center">**The ambiguity of abstract concepts**</div>

. As is seen, the terms frequently used to denote the knowledge that designers generate are relative to context - the 'status' of a particular concept may change as the design process evolves. To view the design process as a concept formation process alleviates this problem of relativity. Whether I propose that "This car is a Volvo" (instantiation or concretization) or "This car is a vehicle" (abstraction) the semantics is the same - 'Volvo' and 'vehicle', respectively, are concepts serving to define the particular concept of 'this car', Whether I refer to a super-class ('vehicle') or a subclass ('Volvo') is immaterial as long as both concepts are described such that they provide to the formation of the primary concept 'this car'. If I, in addition, said that "This car is a killing_device" it wouldn't matter whether 'killing_device' were defined as a class or not - my purpose would be to relate 'this car' to another concept that could clarify what I meant to say about 'this car'.

It is possible to make a similar argument with respect to the system/part distinction. Whether some concept is termed a 'system' or not is rather immaterial - it is the description of the characteristics and composition of that concept which is the primary focus. If sufficient relationships are established to relate the 'system' to its components and characteristics, it is possible to treat the concept as any other concept, whether 'part' or not. If the concept description is not such that it satisfies the expectations, the description must be revised. Most probably, one would then get into the issue of revising the description of 'parts' in an equivalent manner as one would revise a description of a property, such as 'length'. Indeed, as was seen in the discussion of the preposition relation *of*, the reference to a part and to an attribute follow along the same lines; "The hull (a part) of the vessel" and "The length (an attribute) of the vessel" are grammatically identical, as are "The vessel has a hull" and "The vessel has a length.".

It follows from the above that the abstract concepts do not offer much 'content' with respect to describing a particular concept. Their primary roles are to 'assist' in describing still other concepts. In the end, a D-object is thus a collection of relationships where the distinctions into attribute, value, system, part, class, and member become less important than the actual descriptions of the particular concepts going into the description of the D-object.

## Chapter summary

In this chapter the framework of the Design Process Language has been laid out. Requirements to be put to the DPL were presented in the beginning and the categories of relations in the language were developed from this angle.

| Category of relation | Description | {may} appear with |
|---|---|---|
| Material (MR) | Describes activities and processes | {MOD, PR, CR, ADJ} |
| Intensive (INT) | Describes what a concept is | Alone |
| Possessive (POS) | Describes what a concept has | Alone |
| Circumstance (CC) | Describes a concept in relation to other concepts not related Intensively or possessively | Either PR or ADJ and a verb relation |
| Existential (ER) | Describes that a concept exists | Alone |
| Modal (MOD) | Describes futurity, possibility, determination, and intention | MR |
| Preposition (PR) | Used with circumstance relations - used to express spatial, temporal, referential, and other circumstance relationships | CC {ADJ} |
| Conjunctive (CR) | Relates sentences or concepts logically, causally, inclusively, or exclusively. | {MR} |
| Adjective modifiers (ADJ) | Used predominantly with circumstance relations - used to express comparisons among concepts. | CC |
| Adverb modifiers (ADV) | Used to modify verb relations - only *not* is presently considered. | Verb relations |

**Table 5.3:**      Relations and modifiers in the Design Process Language.

These relations, that combine concepts in sentences, were developed from linguistic categories, notably as various verb, preposition, and conjunctive relations, as well as adjective and adverb modifiers. Each of the categories play a particular role in describing different phenomena experienced in, and characteristics of, the design process. The categories, their main characteristics, and the relations with which they interact, are summarized in Table 5.3.

Towards the end of the chapter I argued that several of the terms that are commonly applied in describing objects, such as classes, systems, attributes, and others, are usefully viewed in the concept-formation framework. Such concepts are like any other concepts - they are defined by and serve to define other concepts. A class Is more a concept being related to other concepts constituting class membership criteria, i.e., a specification, than a set of concepts 'belonging to' that class.

Some examples of the usefulness of using a language as the DPL were given towards the end. For instance, to treat a 'class' as an individual concept there must be ways to reference that concept properly -  I argued that this is closely related to the use of language. For instance, in DPL the sentence "The ship *as* a structure" is not a message classifying 'ship' as a structure, as is commonly seen in several representational systems *(has-member* and *has-subclass* relationships, for instance). It is rather a sentence stating that attributes from 'ship' should be  extracted that are relevant for viewing it as a 'structure' concept. Such a sentence could imply that relationships between the 'ship' and its 'paint' and 'rudder' were irrelevant, whereas the relationships between the 'ship' and the 'plates'', 'beams', and  'columns' of the 'hull' would be focussed. In other words, a 'class membership' can be viewed to filter knowledge from a larger, underlying description of the world. In plain terms, the vessel is conceptualized as a structure -  not classified as a structure.

# Chapter 6: Drawing lines

The previous two chapters had two different purposes; Chapter 4 developed a particular view of design, chiefly as a concept formation process, while Chapter 5 developed the framework of the Design Process Language with respect to the requirements resulting from viewing design as in Chapter 4. In the present chapter I will try to show, by means of some basic examples, how the language developed in Chapter 5 can be applied in a computer-assisted design situation, where the design process is one similar to that described in Chapter 4.

It may be appropriate first to briefly repeat the main conclusions. Chapter 4 introduced design as a concept formation process, or as a process of building relationships in a language. The DPL arose from the fact that a language to describe all the relationships potentially found in design, had to be more broad than existing programming or other formal languages. In fact, it had to be almost as powerful as the natural languages.

The proposed framework will support describing most of the relationships that can be identified, or at least understood, in the design process. Whether intelligent behavior is something that can be described propositionally by using definite relations is not to be debated here - it suffices to say that language is an essential ingredient in explaining intelligent behavior, as we know it. Therefore, the expressiveness of the language will Increase the ableness to support intelligent behavior as displayed by the designer.

In the following is assumed that there exists a machine with a processing unit, an interface for communication, a short-term memory and a long-term memory. The imaginary system has encoded grammatical rules that makes possible operations on a set of relations and that checks for consistency in sentences. Several of the sentences (Design memory) represent or are intermediate results from internal processes, so the examples look more involved than they will be experienced by someone using this imaginary machine. In addition, there will (like in any ordinary computer program) be operations that are not described or that can not be "seen".

For the sake of this illustration, the following conventions are introduced:

[DPL System]    **"Prompt from the system for information input, or response from the system as information output"**

[Operator]      **"Information given to the system from the operator"**

*[LT Storage]*    *Sentences that are stored permanently in long term memory - gathered in the system through earlier experiences. At this point, assume that sentences stored in short term memory are transferred into long term memory only if the (human) operator commands the system to do so. That is, the illustrations presented here do not describe anything close to autonomous machine learning.*

*[Design memory]*   *Sentences that are stored temporarily in short term (alternatively design) memory. These sentences are established during the course of this particular design process, as a result from any internal operation or any input to the system. Sentences stored in short-term memory may or may not be retained in long-term memory after completion.*

## The design object

As stated in Chapter 4, the design object (D-object) is viewed as a concept that is defined, or specified, in the course of the design process. If the term (or token) denoting the concept of the D-object is viewed as a 'core', there are in the end numerous sentences, formed in the DPL that connect this term to other, known, concepts. The D-object has parts or sub-systems, which again has parts, and both the D-object, its related sub-systems, and their parts have properties which may be of shape, functionality, color, weight, capacity, and so forth. The properties themselves may even have other properties, such that the steel of a hull skin will have its own properties 'isolated from' the hull skin itself. These properties may be yield strength, density, E-modulus, brittleness, and so forth. It is possible to trace all properties and parts back to the D-object, however. For instance, the below concept definitions are sufficient to construct the sentence "Yield strength of steel of hull skin of hull of D-object equals 27 MPA". Here, hull is sub-system of D-object, hull-skin is part of hull, steel is attribute/property of hull skin, yield strength is attribute of steel, and  27 MPA is value of yield strength.[a]

### The example starts with the designer (operator) constructing a sentence, starting the specification of the design object: ###

**[Operator]**          **"D-object has hull"**

### The sentence is then stored into short term memory by the system: ###

*[Design memory]*     *D-object has hull*

### The concept of hull (which is not yet "known" to the system since it is not yet located in memory) may already have been formed at an earlier point. I.e., it may reside in LT storage. Thus, in LT storage it extracts a general (parts and properties that any hull will have) and partial definition: ###

*[LT Storage]*        *HULL:*
                      *{*
                      *Hull  has  hull  skin*

                      *}*

### The new concepts introduced by 'hull', here 'hull skin', are then found also to be partly defined in LT storage: ###

*[LT Storage]*        *HULL SKIN:*
                      *{*
                      *Hull  skin  has  thickness*
                      *Hull  skin  has  material*

                      *}*

### Traversing further down into the concepts, it may be that 'thickness' and 'material' are partly defined concepts, and subsequently found in LT storage. Assume here that the new concepts introduced below, 'dimension', 'numeric value' and

---

[a]     This is just one interpretation. Hull skin may be a sub-system in others. In addition, it is difficult to say whether it is most correct to view steel as a part, an attribute, or a value. However, this is of no relevance as goes the possibility to map and construct sentences in DPL.

'type' (not their values, though), are defined in the system (as 'semantic' concepts): ###

*[LT Storage]*     *THICKNESS:*
                   *{*
                   *Thickness has dimension*
                   *Thickness has numeric value*
                   *}*

*[LT Storage]*     *MATERIAL:*
                   *{*
                   *Material has type*
                   *}*

### The 'missing' definition of the different concepts are prompted by the system. In this case, the system asks for the value of dimension (it is assumed that the system knows what kinds of dimensions are allowable): ###

**[DPL System]**   **"What is dimension of thickness (of hull skin of hull of D-object)?"**

### The designer responds to the question from the system: ###

**[Operator]**     **"mm"**

### The new information then is registered as a sentence in short term storage: ###

*[Design memory]*  *Thickness has dimension mm*

### Similarly to the above situation, the human designer is prompted by the system for missing information: ###

**[DPL System]**   **"What is value of thickness (of hull skin of hull of D-object)?"**
**[Operator]**     **"16"**
*[Design memory}*  *Thickness has value 16*

**[DPL System]**   **"What is type of material (of hull skin of hull of D-object)?"**
**[Operator]**     **"A36 steel"**
*[Design memory]*  *Material has type A36 steel*

### Here, the designer responded with A36 steel. This is now found to be defined earlier as a concept in long term memory: ###

*[LT Storage]*     *A36 STEEL:*
*{*
*A36 steel has density*
          *Dimension (of density) is tons/m$^3$*
          *Value (of density) is 8.0*
*A36 steel has yield strength*
          *Dimension (of yield strength) is MN/m$^2$*
          *Value (of yield strength) is 27*

*}*

### The 'new' concepts introduced by the concept 'A36 Steel' are then also found to be defined in LT storage:###

*[LT Storage]*     *DENSITY:*
*{*
*Density has metric dimension*
*Density has numeric value*

*}*

*[LT Storage]*     *YIEW  STRENGTH:*
*{*
*Yield strength has metric dimension*
*Yield  strength has numeric value*

*}*

As is seen here, the design process becomes a sort of "fill in the blanks" process. Information that is added to the description of the D-object is used by the system to identify what information is missing. Some concepts, such as A36 steel, are completely defined (in this context). Others are incompletely defined, such as D-object. Between these two extremes, the concepts introduced are defined to a varying degree.

The only statements entered in the system about the object are the four sentences listed below. The remaining definitions were present in LT storage. Thus, only the information that can not be found as part of the contained concept definitions must be entered as part of the D-object description. The role of the designer thus was to introduce the following information into design memory:

*[Design memory]*    *(l)  D-object has hull*
*[Design memory]*    *(2) Dimension (of thickness of ... ) is mm*
*[Design memory]*    *(3) Value (of thickness of ... ) is 16*
*[Design memory]*    *(4) Type (of material of ... ) is A36 steel*

If the sentence "D-object has hull" were replaced with "D-object is a ship" the system would conceivably identify the general concept of 'ship' as 'having' a 'hull ', assuming the system had encountered this concept earlier and retained its definition in long term memory. Sentence (1) would then be unnecessary. If it were replaced with the sentence "D-object is 'Berge Viking'", a perfectly defined concept assuming that "Berge Viking" has already been built and defined in memory, all sentences would be unnecessary since the information could be extracted from the concept 'Berge Viking ·. Sentence (2) is an example of knowledge that can be learnt by, or taught to, the system ("Dimension of thickness of hull plates is mm") as a general rule.

This illustrates the use of the DPL in describing objects in a design process. Note that the only DPL relations used here are *has* (possessive verb relation), *is* (intensive verb relation) and *of* (referential preposition relation).

It is important to recall from earlier discussions (Chapter 5) that the DPL does not distinguish whether some concept 'has· a property, a part, a sub-system, or anything else. The concept being 'had· is defined partly isolated from its possessor, and the relationship between the possessor and the possessed is identical whether it is a dimension or a hull that is 'possessed ·. The identification of a hull as a part comes into the picture only if there is something in the definition of 'hull ' that distinguishes it as a part (for instance, it having dimension or weight). Tangible versus intangible concepts is of no importance in a system operating with the DPL.

# The D-object and the "World"

Above was illustrated how an object could be described by means of DPL. In this section, an illustration is given to indicate how the D-object may be 'mapped' to its requirements and to concepts describing the external environment (e.g., "sea water') of the object.

### The process starts with a requirement that is added to memory: ###

**[Operator]**      **"D-object should float"**
*[Design memory]*    *D-object should float*

### The system finds a useful relationship in long term memory. In this case, the term 'something' is a place-keeper, equivalent to a *universally quantified sentence* in predicate calculus. The following sentences will probably not be generated until the system is asked a question as the one below, but are shown here for the sake of clarity: ###

*[LT Storage]*    *Something floats if maximum of buoyancy is larger than maximum of weight*

### Then the system instantiates D-object for 'something': ###

*[Design memory]*    *D-object floats if maximum of buoyancy is larger than maximum of weight*

### The system instantiates 'float': ###

*[Design memory]*    *D-object should have maximum of buoyancy larger than maximum of weight*

### Now, the designer asks a question (which is then added to short term, or design, memory): ###

**[Operator]**      **"What is maximum of weight of D-object?"**
*[Design memory]*    *What is maximum of weight of D-object?*

### The system then finds some general and usable relationships in long term memory. Significant shortcuts are made with respect to the interpretation and processing of the modifier 'maximum': ###

*[T,T Storage]*    *Maximum of buoyancy equals maximum of weight of displaced liquid*

| | |
|---|---|
| *[LT Storage]* | *Maximum of weight of displaced liquid equals maximum of suppressed volume times density of liquid* |
| *[LT Storage]* | *Maximum of suppressed volume equals Cb times L times B times T* |

### The system then will prompt for missing information, the designer respond, and the new relationships are added to short term memory. If the system could find relationships that made it able to resolve the unknown concepts, the designer would not need to be involved in this process: ###

| | |
|---|---|
| **[DPL System]** | **"What** is Cb? " |
| **[Operator]** | **"0.62"** |
| *[Design memory]* | *Cb is 0.62* |
| **[DPL System]** | **"What is L? "** |
| **[Operator]** | **"120 m"** |
| *[Design memory]* | *L is  120 m* |
| **[DPL System]** | **"What is B? "** |
| **[Operator]** | **"20 m"** |
| *[Design memory]* | *B is  20 m* |
| **[DPL System]** | **"What is T? "** |
| **[Operator]** | **"7 m"** |
| *[Design memory]* | *T is  7 m* |

### After accessing all required concept definitions, the system will calculate (or resolve) the relationship that was detected in long term memory. (This is after instantiating values for the symbol). The first sentence has been found earlier (above), the second is an instantiation, and the third results from a transaction (resolution): ###

| | |
|---|---|
| *[LT Storage]* | *Maximum of suppressed  volume equals Cb times L times B times T* |
| *[Design memory]* | *Maximum of suppressed  volume equals 0.62 times 120 m times 20 m times 7 m* |
| *[Design memory]* | *Maximum of suppressed  volume is 10416 $m^3$* |

### The system now must resolve the concept "Density of liquid", after having detected 'liquid' as being a concept 'having' a density, and must get specific information from the designer: ###

| | |
|---|---|
| **[DPL System]** | **"What is type of liquid? "** |
| **[Operator]** | **"Sea water"** |

*[Design memory]*     *Type of liquid is sea water*

### 'Sea water' is then detected in long term memory. Had the response above been "Type of liquid is x" the system might have prompted something like "What is density of x?" since x is  undefined in memory, and since **it** necessarily must have a density.###

*[LT Storage]*       *Density of sea water is 1.025 tons/m³*

### The system calculates (or resolves a relationship): ###

*[Design memory]*     *Maximum of weight of displaced liquid equals 10676 tons*

### The system replaces 'weight of displaced liquid' with 'buoyancy' in the relationship found earlier: ###

*[Design memory]*     *Maximum of buoyancy equals 10676 tons*

### Instantiating: ###

*[Design memory]*     *D-object floats if 10676 is larger than maximum of weight*

### The system performs a grammatical operation, 'reversing' the relationship above:###

*[Design memory]*     *D-object has maximum of weight less than 10676 tons*

### The system is now capable of "answering" the question stated in the beginning: ###

**[DPL System]**     **"D-object has maximum of weight less than 10676 tons"**

### The designer now asks a new question: ###

**[Operator]**       **"What** is **maximum of deadweight?"**
*[Design memory]*     *What is maximum of deadweight?*

### The system finds a relationship stored in long term storage: ###

*[LT Storage]*       *Weight equals deadweight (of D-object) plus weight of lightship (of D-object)*

### The system performs a grammatical operation (reverses the relationship) and instantiates 'deadweight': ###

[Design memory]   *Maximum of deadweight equals maximum of weight (of D-object) minus maximum of weight of lightship (of D-object)*

### Now, the system has found no usable relationship to calculate weight of lightship, and the designer is asked. If the designer wanted to, she could at this point enter a relationship, such as a formula including, e.g., L, B, D, Cm, and Cb. The system would then search for definition of parameters (concepts) and prompt for the missing information if no method existed to find them, as in "What is Cm?" and "What is Cm of D-object?" Upon learning the new relationship (new formula), this might be added to long term memory. However, the designer responds with a fixed value: ###

[DPL System]   **"What is weight of lightship (of D-object)?** "
[Operator]       **"3500 tons"**

[Design memory]   *Weight of lightship is 6500 tons*

### The system instantiates (maximum weight is known from earlier): ###

[Design memory]   *10676 equals maximum of deadweight (of D-object) plus 6500 tons*

### The system performs a grammatical operation (reversal): ###

[Design memory]   *Maximum of deadweight (of D-object) equals 10676 tons minus 6500 tons*

### The system performs a calculation (through a subtraction relation, which is a preposition relation): ###

[Design memory]   *Maximum of deadweight (of D-object) equals 4167 tons*

### The system performs a translation. *Equal* to *is,* in this case: ###

[Design memory]   *Maximum of deadweight is 4167 tons*

### It is now possible for the system to answer the question asked earlier by the designer: ###

[DPL System]   **"Maximum of deadweight is 4167 tons"**

# Design process

The process description that follows introduces some material relations (see Chapter 5) that are used in solving the design problem. It is impossible here to list all possible transactions that take place -  assume that the relations have an underlying definition that enables the system to chose appropriate lower-level transactions. The concept definitions are not given here, and some other shortcuts are also made in the course of the illustration.

### The question asks a question that is registered into short term memory: ###

| | |
|---|---|
| **[Operator]** | **''What is weight of hull skin?''** |
| *[Design memory]* | *What is weight of hull skin?* |

### The system commands itself to attempt a resolution of the concepts introduced in the sentence: ###

*[Design memory]*     *Resolve concepts!*

### The system then cannot find any value of weight in short‑term memory. Here is assumed that the system interprets this question as one asking for value -  not for other possible parts of the definition of the concept "weight of hull skin". Furthermore, it is assumed that 'hull skin' refers to the last referred 'hull skin', namely that of the hull of the D-object: ###

*[Design memory]*     *Weight of hull skin is unknown*

### Since the concept is unknown, the system constructs a command to itself aiming to find the missing information (define the unknown concept):

*[Design memory]*     *Find weight of hull skin!*

 ###
### New internal command to detect a transaction that can make the system able to solve the problem (find, or define, the concept): ###

*[Design memoryⅉ*     *Locate method!*

### One relationship that may be used is found. Assume here that the system has used some decision rules, for instance If ... then ... :  ###

*[LT Storage}*          *Weight equals density times volume*

### The relationship above is instantiated: ###

*[Design memory]*      *Weight of hull skin equals density of hull skin times volume of hull skin*

### The system commands itself to attempt a resolution of the concepts introduced in the sentence: ###

*[Design memory}*      *Resolve concepts!*

### The system finds that the .concept has been defined earlier. ###

*[Design memory}*      *Hull skin has material A36 steel*

### The system finds necessary relationship through searching memory and.interpreting relationships found there: .###

*[LT Storage]*          *A36 steel has density 8.0 tons/m$^3$*
*[Design memory}*      *Density of hull skin is 8.0 tons/m$^3$*

### The system could not find volume (concept introduced above) to be defined in relation to this problem. It commands itself to find a means to do this.: ###

*[Design memory]*      *Volume is unknown*
*[Design memory]*      *Find volume of hull skin!*
*[Design memory]*      *Locate method!*

### **A** <u>possible</u> relationship is found and instantiated. ###

*[LT Storage]*          *Volume equals surface area times thickness*
*[Design memory}*      *Volume of hull skin equals surface area of hull skin times thickness of hull skin*

### New concepts (in this case, parameters in a formula) are introduced with the chosen relationship, and a new internal command is constructed: ###

*[Design memory]*      *Resolve concepts!*

### The system detects a definition of one of the concepts from earlier: ###

| | |
|---|---|
| *[Design memory]* | *Hull skin has thickness 16 mm* |
| *[Design memory]* | *Thickness of hull skin is 16 mm* |

### Here is assumed that the system finds a rule that determines that dimensions need to be consistent.

*[Design memory]*       *Convert dimension to m*

### The system finds a conversion (mapping) in long term memory: ###

*[LT Storage]*          *1 mm equals 0.001 m*

### And performs a transaction to convert: ###

*[Design memory]*       *16 mm equals 0.016 m*
*[Design memory]*       *Thickness of hull skin is 0.016 m*

### The system finds no trace of the remaining parameter: ###

*[Design memory]*       *Surface area of hull skin is unknown*

### Searches and fails to find a useable relationship: ###

*[Design memory]*       *Find surface area of hull skin!*
*[Design memory]*       *Locate method!*
*[Design memory]*       *No method found!*

### The system is left blank - no useful relationship has been found in memory: ###

**[DPL System]**        **"No method found. Define method, give value or identify range?"**
**[Operator]**          **"Identify range"**

The example is further developed in the appendix.

# Chapter summary

The chapter has given some illustrations of how the DPL may be conceived applied in a design situation. It is beyond the scope of this thesis to state any proofs that the DPL will work, but the illustrations presented in this chapter should indicate that the language itself may be used in a machine assisted design process.

The examples have shown that it is possible to use the DPL for both object and process descriptions as a tool in the design process. The main requirement is an operating system that can handle it. Nothing has been said as to how to represent the various abilities that are used in such a design process, and how. the system detects the actual processes used to calculate, find, plan, and so forth. Neither has anything been shown as to how the grammatical rules used to interpret sentences are constructed, refined, or operated. Most of these problems have been discussed at some point throughout. A brief discussion of limitations and possibilities is presented in Chapter 7.

# Chapter 7: Concluding remarks

In the thesis I have developed some different ideas of how design may be viewed and, towards the end, tried to develop the framework of a language - the Design Process Language - with which it may be expressed, The fundamental hypothesis underlying the entire work is that design may essentially be viewed as a process of using and producing relationships. This idea was used and developed further until the point where the framework of the DPL was suggested. I will here propose a critique and draw some conclusions from the work.

# Critique

The (few) 'invariants' in design that I have discussed and argued for, particularly through comparing theories regarding intelligent behavior and design research in Chapter 2, treated <u>high level processes</u> and were only <u>indicative.</u> Little was said on how designers actually perform in design activities. The reason for this is that the works within design research show little convergence as to what indeed are invariants at the lower levels of operation.

I suggest in the introduction that this might be so because the ways individual humans design in individual tasks differ so much that it is impossible to establish all but very general theories. Presumably, design researchers must focus on coherent theories that have chances of being consented to. But the high-level processes involved in intelligent activity are not what the DPL are to describe - DPL is to represent knowledge used in actually planning and performing tasks specified by an external (intelligent) agent or some other authority.

This does, however, cause problems in verifying that the DPL is of virtue - if little is known about what a language is to express there is little chance of proving that the language is expressive! The DPL has then, so far, become a <u>framework</u> where quite general categories are identified but where few concrete relations or concepts are defined.

An obvious problem in using the Design Process Language framework is that the structure it presupposes is propositional, where relationships are embedded in sentences. It is not oriented towards databases and file structures as all of us have increasingly fallen victim to - indeed, the knowledge represented must probably be directly accessible in the DPL format. As with all new approaches, this causes compatibility problems in case of implementation of the DPL - it makes difficult any communication with today's program-specific data storage. It may, of course, be possible to create artificial linkers and to let concept like 'data model' define another format, such as "Data drawing has format" and "Format of data model is DXF." The content of a file may then treated as the definition of a concept - somewhat like a semantic concept.

Nothing has been mentioned regarding how to encode semantic concepts (those that are understood through their signification in the real world) in a DPL implementation in a computer. This may be a large problem if the purpose is to create intelligent programs. However, the part of the design system residing in the computer may here be dumb, or closed. In other words, it may be purely symbolic. The designer is viewed as both the 'intelligent authority' that uses a computer as an assistant, and is viewed as the interface with the world.

There are obvious problems of spatial and temporal efficiency in using DPL in an automated environment. However, this is probably an increasingly inferior problem to that of enabling representational, or epistemological, adequacy in the languages used to describe what is known about the world.

Little has been said regarding how to represent 'knowledge about knowledge'. Such information may be at what point a particular relationship is established, which may be of importance for purposes of mapping design history. Other such information may be issues regarding internal storage of information, for instance location of sentences in memory. These are serious problems that remain to be addressed.

The frequently cited concept of control knowledge is not discussed. Obviously, there must be knowledge embedded in a human mind or in a computer such that the application of (grammatical) rules is controlled to find what actions to take upon learning a new relationship. It may be expected that some of this control may be captured through a kind of hegemony that can be described through transitivity in the grammar, as in "A controls B" and "B controls C" implies that "A controls C."[a] Related to this is that, in the bottom line, some knowledge must be <u>behavioral,</u> like the interpretation of <u>fundamental</u> (material) relations and grammatical rules operating upon relations. This knowledge can obviously not be described in terms of concepts and relations in propositional or semantic memory although composite relations (those propositionally defined by means of other, more fundamental, relations) can be. The encoding of the <u>grammatical rules</u> that interpret the meaning of sentences is a different issue, but does in principle not differ that much from the technology of encoding operating systems in today's computers.

The DPL is based on representing and manipulating symbolic knowledge, whereas a lot of information relevant in a design process may be figural, in particular graphical. This limitation is similar to that experienced in any language, and is thus not particular to DPL - there probably does not exist enough words in the world to describe all relationships that may be found in an image of a human face.

## Conclusion

I will propose that the contributions from this work can be categorized in two main categories; those related to the development of the view of design as a concept-formation process and those related to the DPL itself. I will conclude the thesis by summing up the main points in these two categories.

### On design

It has become apparent that there is no single way to look at design. It is a multi-faceted phenomenon - sometimes a multi-headed monster. With each new angle that we take as observers, with each new light-setting, with each new pair of glasses we put on, different aspects of the design process surface. I have come to land on three views of design; as a <u>concept-formation process</u>, as a <u>hypothesis-proving process</u>; and as a <u>hypothesis-generating process.</u> Each of these views discloses different characteristics of design that I will summarize here.

---

[a]   It is not shown here under what conditions this issue of control is relevant. During the course of my research, however, it became clear that this is a <u>possible</u> approach to making the grammar controlled.

### Design as concept formation

It is widely recognized that the design process is a process of which the aim is to discover some knowledge about a non-existing object that, when materialized, is to have a purpose. In other words, the process has as an ultimate goal to describe a product that is not yet known. The way to describe this unknown object is to conceptualize it in terms of known things, or known concepts. The designer relates the unknown concept to other concepts that ultimately carry meaning to the <u>initially abstract and unknown</u> and <u>ultimately concrete and known</u> concept of the design object.

This sheds light on some more fundamental characteristics. In order for the design object to have the slightest chance of filling the purpose that it is set to fill, it must be possible to design it with a view to (understand it in terms of) that purpose. Or, said in another way, it must be possible to understand concepts describing the purpose in terms of concepts describing the design object. If one concept describing purpose is 'transport', then the design object must necessarily have a structure that in some way enables it to 'transport'. In other words, there must exist a way to relate the 'transport' concept to the design object.

The argument above can be extended. In order to describe the <u>role</u> of a design object, or how it will behave when materialized and put to work, it is also generally recognized that there must exist means of relating this object to the environment in which it is to exist in the future. <u>The design object must be understood in terms of how it will perform in that environment.</u> In other words, it must be possible to create relationships between concepts describing the environment and those describing the design object. The 'artifact', as proposed by Simon, is then a set of relationships serving to connect concepts from the two 'worlds', respectively. Phrased in parametric terms; to model how a product will function in a given environment, the variables used to model the product and those used to model the environment must both be accessible to the program that is used to evaluate their interaction.

As a matter of fact, it must be possible to merge <u>all</u> knowledge that is relevant for assessing to what degree an object performs satisfactorily through <u>relationships</u> that connects all this knowledge into a whole - in some sense, it must be possible to construct a 'network' of concepts. This network may also include knowledge of constraints, of evaluation criteria, of procedures, and so forth.

The last enlightenment to achieve from the concept-formation view of design is that in order for anyone else to understand what the designer describes, there must in the description of the design object, be concepts that are understood by others. While nobody would know very much about the design object through simple knowing the phrase "this is an x," a lot more understanding ould be gained <u>by others</u> if this sentence were followed by "x is a ship." It should not be very controversial to claim that this arises from the fact that the more share the same understanding of the concept "ship ' than they do of the concept 'x '. However, "this is an x" might make perfect sense to the designer, assuming that the designer were accustomed to the substitute 'x ' for "ship·. This leads to the conclusion that at least <u>some</u> of the concepts used in describing the design object must be <u>known</u> by the one that is to materialize this object. Further, <u>all</u> (relevant) concepts (that may initially be unknown) used in describing the design object must somehow be <u>understandable</u> in terms of other, known, concepts.

Thus, the key to understanding a design object description is to understand the concepts used to describe it. In this conception, 'understand' implies that the designer and the constructor 'understand' in a similar manner -  if there is a divergence in understanding there is an  opening for a materialized design object to be different from that which was intended from the designer's head and hand. The designer must find those generally known concepts that can be used as 'vehicles of meaning'. This also goes further - it may be necessary to specify standards so as to provide a <u>sufficient set of carriers of</u> <u>meaning</u> for a successful communication between the designer and the constructor to take place.

**Design as proving hypotheses**

It is shown that the design process is in some sense similar to the process of proving hypotheses. The knowledge that the designer initially has access to is a set of statements about what the design object is to achieve in the future. These statements are in nature similar to hypotheses. The design object thus becomes a specification of an object that appears to prove correct hypotheses about the future. When designers test whether their design will meet expectations, they might easily cast this test as something like "the design object satisfies requirements." If this proves wrong, the hypotheses are not proven and the design process is not finished.

This brings up some additional points. The 'facts' that the designer 'learns' about the design object all have an overall purpose - to support the claim that the hypotheses are correct. If a fact does not contribute, directly or indirectly, towards proving a hypothesis, it may be an irrelevant or redundant fact. In some sense, this is like saying that all aspects of the design object must have a reason for existing, and that this reason in some way should be possible to trace back to the initial expectations - the hypotheses.

Another conclusion to draw in the hypothesis-proving view of design is that the test - the performance modelling - must be made such that the results from these tests are understood in the same terms as those used in stating the hypotheses. This actually is mapping between the hypotheses and the concepts that describe the design object.

**Design as generating hypotheses**

It is also shown that design may be viewed as a process of generating hypotheses. The specification that the designer produces is only a hypothetical construct - the result is still an abstract concept that becomes concrete only if its description is such that its materialization is ensured. The designer must consider how to ensure a correct materialization - possibly through also including in the description procedures and instructions of how to implement the result.[b] She must design hypotheses that are easily understood and easily tested. Arguably, it is easier to test an hypothesis that states <u>structural configuration</u> than one that states <u>functional behavior</u>. At least it is easier to *a priori* ensure that the finished result corresponds to the hypotheses if they state requirements to <u>physically observable phenomena</u> and not simply <u>functional behavior.</u>

It seems probable that this is part of the reason why there is a need for a designer to mediate between the one who states the initial expectations (the customer) and the one who materializes the result (the constructor). <u>The designer is an expert in translating hypotheses of a functional (intangible) nature to hypotheses of a structural (tangible) nature, while the constructor is an expert in translating hypotheses of a structural (tangible) nature to a physical construct (extremely tangible).</u> As a product evolves from idea to material fact, the hypotheses describing it evolve along the scale of 'tangibility' - the ratio of <u>hard to soft knowledge</u> increases. (See Mistree et. al., 1990 and Smith, 1992)

Furthermore, this view stresses the point that the designer cannot assume her specification to be an exact picture of the materialized result. This depends on to what degree the constructor is able to prove her hypotheses (specification) right. The role of the designer then becomes to ensure that the hypotheses are stated such that the chance of the constructor being able to prove them is high. In other words, if the constructor produces a product that 'proves' the designer right in her description, the constructor has 'proved' the hypotheses suggested by the designer. But this chance then depends on how these hypotheses are formulated. It is probably easier for a constructor to construct a block of wood described as dimensions of length, breadth, and height than one that is simply described as a volume. The design object should then be hypothesized as a construct in terms of those three dimensions. As seen from the perspective of the designer, assuming that all that is needed is a certain volume regardless ot how it is achieved, both approaches are sound, since both satisfy requirements. However, the designer will probably specify dimensions to increase the probability of the constructor making a product that satisfies the requirements to volume.

---

[b]    This is also said by others, for instance Goel and Pirolli as discussed in Chapter 2. However, they <u>define</u> this as part of an <u>artifact (design object) description</u>. I claim it to be a <u>consequence</u> of the design object as being hypothetical in nature!

## On the Design Process Language

As a natural consequence of viewing design as a concept-formation process, as relating concepts, it is quite near in thought to presume that design could fruitfully be viewed as a process of using and producing sentences in some language. I have termed this language the Design Process Language and chosen to use a natural language - English - as a platform.

### Process

One of the obvious requirements to place on the Design Process Language is that it shall be possible to represent process with it. Thus appeared the material verbal relations, usefully viewed as a subset of the transitive verbs in natural languages. In this category are those relations that normally describe what an agent (a subject) 'does' to an object when they are used in a description.

Another category was less obvious in describing process. This was the category of modal verbal relations. The design process implies a planning of activities, and the existence of activities imply that there exist abilities to perform activities. The modal relations include exactly such verbs that have to do with, for instance, representing abilities *(can (do)),* future required and necessary activities *(must (do), will (do)),* and future desired activities *(should (do)).*

Some other relations also turned out to be necessary for the purpose of describing processes; the 'composite' category (including a verbal relation) of circumstance relations, a subset of which "operate· with prepositional relations or adjective modifiers. For instance, the (verbs) prepositions *(executes) before* and *(executes) after,* and the (verbs) adjectives (prepositions) *(is) faster (than)* and *(is) slower (than).*

Other requirements in the process view were even less obvious. Commonly, processes are treated in a similar manner as the objects in that they are decomposable into sub-processes just as a structure is decomposable into parts. The possessive relations then turned out to be relevant. One need not go far to see this decomposability and the use of possessive relationships - most "programs *have* subroutines."

Obviously, processes need to exist, something represented through the category <u>existential relations</u>. ("There *is* a program") Obviously, processes <u>are</u> something as well, creating the need for the <u>intensive relations</u> in the DPL. ("The program *is* a C-program").

In going even further into the process aspects of design, it became apparent also that <u>historical information</u> could be relevant, whether for re-design purposes, for purposes of learning, or for purposes of documentation. This was made possible by the category of <u>conjunctive relations</u> ("The attempt failed *because* ...") and the <u>(grammatical) rules in the language</u> enabling a distinction between past, present, and future ("The program will *calculate* ...," "The program *calculates* ...," and "The program *calculated* ..."). The conjunctive relations also are important in representing, for instance, rules for choosing activities; *"If(...) and* (...) then (...) *or* (...)."

In short, the relations that could be found in the language of English proved to be efficient, at least in supplying terminology, for a DPL to handle a general design process description.

**Object**

I will here use the term <u>object</u> to refer to the general class of non-process knowledge. It should be quite clear what is meant with this term; non-process knowledge gathered, used or produced in the course of the process, like "The length of the vessel is 140 meters."

The categories of relations differ not as much as might be expected from those needed to represent process. However, the importance of the various categories is different. The <u>existential relations</u> were needed to state the presence of an object ("There *is* a design object") and the <u>intensive relations</u> to state what the object actually <u>is</u>. ("The design object *is* a ship" and "The length *is* 140 meters"). The <u>possessive relations</u> were probably more important here than in describing processes. It is a frequent activity in design to decompose into less general or less encompassing concepts; "The ship *has* machinery" and "The ship *has* length."

The prepositional relations most relevant for describing object were somewhat different from those relevant for describing process. Whereas the most relevant (circumstantial - prepositional) relationships among otherwise 'independent' processes are temporal in nature, the most relevant such relationships among otherwise independent objects are of a more spatial nature; "The machinery is *on* deck 1" and "Deck 1 is *below* deck 2." Others are relevant to both process and object; "The program writes *to* a file" and "The propeller is connected *to* the shaft," and "The program executes *at* (time)" and "Crane is located *at* (position)." The way the prepositions are interpreted depends on relations they cooperate with, in this case the verb relation. If the verb relation is material, the prepositional relation is temporal.

The same difference applies partly to the adjective modifiers as well. Those of temporal nature apply less to objects than those of spatial nature, such as in "Beam A is *longer* than Beam B." However, some are not process or object specific, such as in "Program A is *better* than Program B" and "Alternative A is *better* than Alternative B."

The conjunctive relations may, of course, be applied to object as well as to process; "The machinery is *either* a steam turbine *or a* gas turbine" and "If the machinery is a gas turbine then the RPM is high."

**Requirements and expectations and hypotheses**

The modal relations have an important role in describing the initial statements of expectations and requirements, what was earlier referred to as hypotheses. These relations are all in some way or another used for expressing conditionality, futurity, expectations, hypotheses, possibility, and so forth. They thus seem well suited to describe that which is to become, like in "The length *must* be less than 140 meters," "The ship *should* cost less than 20 million dollars, and "The ship *ought to* be safe." To state it simply, a designer using DPL would need to resolve all such modal relations - prove them 'right' before ending the work. These relations are some of the more powerful ones, and probably the more 'design specific' ones. in the DPL.

# Further work

## Theoretical

From the theoretical point of view, several angles could be taken to increase understanding of design, as viewed in the context of this thesis. In this section are discussed some of the more important works to be performed in the future.

> Investigate whether design activity is usefully explained in terms of Sternberg's seven components.

In Chapter 2 several indications were presented to the extent that design is closely related to the view of intelligence by Sternberg. This particular aspect of Information Processing Theory may thus be a useful basis for undertaking research into abilities that make humans able, and well able, to design. By using the seven components presented by Sternberg as categories in which to 'place' different observations it might be expected that new light is shed on what is design,

> Investigate what kinds of relations among concepts are most important in a language used for describing the design process and the knowledge it uses and produces.

The categories of relations presented in Chapter 5 were all shown to be of some relevance to describing the design process and the knowledge used and produced in the process. It may be useful to both perform theoretical and empirical research to find what role the various kinds of relations play in building a model in design.

> Investigate what kinds of material relations are most important in design.

The material relations are the most crucial in describing the process aspects of design. They are grossly transitive verbs. It may be of use to study the design process to find what kinds of material relations could most frequently be used to describe it.

> Either (a) describe the relations in terms of other, more
> fundamental, relations, or (b) trying to classify them further
> · through identifying common properties among sets of relations.

Some material relations will probably be more important than others. In addition, it may be fruitful to view individual relations, such as *analyze,* to constitute classes of other relations, like *calculate, decompose,* and so on. In addition, there may be the case that some relations are <u>decomposable</u> into gradually more elementary material relations (processes), like Newell and Simon argue.

> Investigate whether the set of <u>modal relations</u> is an
> 'epistemologically adequate ' set of such relations.

The modal relations were shown to be of ultimate importance in planning for the future and treating potential or possible events or facts. I would expect that the set of modal relations could be extended to be a sufficient set in practice - I haven't made the effort to prove sufficiency, if such a proof is indeed possible.

> Classify phenomena detected in design behavior in terms of
> the relations used to describe these phenomena as defined in
> the Design Process Language framework.

It may be an interesting experience to find whether indeed design behavior can be expressed through the categories of relations suggested here. It was beyond the scope of this thesis to test the DPL against design theories and observed behavior to investigate the expressiveness, although I have tried to show, by means of examples, some of the expected expressiveness of the DPL.

## Practical

Some further tasks are of a more practical nature, seeking primarily to give evidence for some of the theoretical findings in this Thesis and paving way for a possible implementation of them.

> Investigate in what way the relations interact, and what
> grammatical (interpretational) rules are specific to each
> category of relations.

Several of the relations suggested in Chapter 5 display different behaviors when joined to form 'composite· relations. Consider the relation *is* alone, *is above, is close to,* and *is closer than.* It is a task not only to find rules that resolve a relation when it is functioning alone to relate concepts, as in "The car *is* red," but a!so when it functions with other kinds of relations, like "The engine *is above* the keel." For instance, some of the circumstance relations (attributive) always presuppose that a verbal relations is cooperating with another (prepositional relation or adjective modifier). So, pairs of relations from different categories may display characteristics that differ from that of both categories viewed in isolation.

> Find invariant features among relations in the various
> categories.

There are probably also some characteristics of the relations that are constant within each category - that is, rules that are category specific. It will be useful if these characteristics could be identified.

> Investigate how to treat data in a computer environment using
> the DPL particularly with respect to storage and search.

An obvious area of research is to find what demands to put on hardware and software, particularly memory and operating system, to facilitate storage of sentences and search to find concept and relations occurring in these sentences. For instance, while conventional data management systems are based on pointers to locations in the database, it is more likely that when DPL is used the words are treated as they are - not as symbolizing some 'deeper' knowledge. An exception are the semantic concepts - those that Indeed do have some deeper meaning for the interpreting system. Presumably, these need to point to other entities that 'carry meaning' to the system. (Like, for instance, the mapping in digital computers between the number 9 and the binary representation 00001001.)

> Investigate rule structures and procedures inherent to the system (behavioral knowledge).

This is a field where extensive research has already been undertaken. However, the aim with such research in terms of the DPL would be to find how the general rules and procedures that are needed to understand the meaning embedded in sentences should be stored, structured, and operated. In other words, how the language is controlled.

# Appendix: Example from Chapter 6

The appendix contains a complete listing of the example started towards the end of Chapter 6.

### The designer asks a question that is registered into short term memory: ###

**[Operator]**          **''What is weight of hull skin?''**
*[Design memory]*      *What is weight of hull skin?*

### The system commands itself to attempt a resolution of the concepts introduced in the sentence: ###

*[Design memory]*      *Resolve concepts!*

### The system then cannot find any value of weight in short term memory. Here is assumed that the system interprets this as a question for value - not for other possible parts of the definition of the concept "weight of hull skin". Furthermore, it is assumed that 'hull skin' refers to the last referred 'hull skin'. namely that of the hull of the D-object: ###

*[Design memory]*      *Weight of hull skin is unknown*

### Since the concept is unknown, the system constructs a command to itself aiming to find the missing information (define the unknown concept.:###

*{Design memory]*      *Find weight of hull skin!*

### New internal command to detect a transaction that can make the system able to solve the problem (find, or define, the concept): ###

*[Design memory]*      *Locate method!*

### One relationship that may be used is found. Assume here that the system has used some decision rules, for instance If ... then ...: ###

*{LT Storage]*         *Weight equals density times volume*

### The relationship above is instantiated: ###

*{Design memory}*        *Weight of hull skin equals density of hull skin times volume of hull skin*

### The system commands itself to attempt a resolution of the concepts introduced in the sentence: ###

*{Design memory}*        *Resolve concepts!*

### The system finds that the concept has been defined earlier. ###

*[Design memory]*        *Hull skin has material A36 steel*

### The system finds necessary relationship through searching memory and interpreting relationships found there: ###-

*{LT Storage}*           *A36 steel has density 8.0 tons/m³*
*[Design memory]*        *Density of hull skin is 8.0 tonslm³*

### The system could not find volume (concept introduced above) to be defined in relation to this problem. It commands itself to find a means to do this: ###

*[Design memory]*        *Volume is unknown*
*{Design memory}*        *Find volume of hull skin!*
*{Design memory}*        *Locate method!*

### A <u>possible</u> relationship is found and instantiated. ###

*[LT Storage]*           *Volume equals surface area times thickness*
*[Design memory]*        *Volume of hull skin equals surface area of hull skin times thickness of hull skin*

### New concepts (in this case, parameters in a formula) are introduced with the chosen relationship, and a new internal command is constructed: ###

*[Design memory]*        *Resolve concepts!*

### The system detects a definition of one of the concepts from earlier: ###

*[Design memory]*        *Hull skin has thickness 16 mm*
*[Design memory]*        *Thickness of hull skin is 16 mm*

### Here is assumed that the system finds a rule that determines that dimensions need to be consistent.

*[Design memory]*      *Convert dimension to m*

  ###
### The system finds a conversion (mapping) in long term memory: ###

*[LT Storage]*         *1 mm equals 0.001 m*

### And performs a transact10n to convert: ###

*[Design memory]*      *16 mm equals 0.016 m*
*[Design memory]*      *Thickness of hull skin is 0.016 m*

### The system finds no trace. of the remaining parameter: ###

*[Design memory]*      *Surface area of hull skin is unknown*

### Searches and fails to find a useable relationship: ###

*[Design memory]*      *Find surface area of hull skin!*
*[Design memory]*      *Locate method!*
*[Design memory]*      *No method found!*

### The system is left blank -  no useful relationship has been found in memory: ###

**[DPL System]       "No method found. Define method, give value or identify range?"**
**[Operator]         "Identify range"**

### The system finds and instantiates the relationship describing range. 'Instantiation' is performed by !!Sing the referential preposition relation *of* ###

*[LT Storage]*         *Range is less than maximum and larger than minimum*
*[Design memory]*      *Range of weight is less than maximum of weight and larger than minimum of weight*
*[Design memory]*      *Range of weight of hull skin is less than maximum of weight of hull skin and larger than minimum of weight of hull skin.*

### The system can not find the necessary concept in memory: ###

*[Design memory]*    *Resolve concepts!*
*[Design memory]*    *Maximum of weight of hull skin is unknown*
*[Design memory]*    *Find maximum of weight of hull skin!*
*[Design memory]*    *Locate method!*

### One useful relationship was found earlier in the process: ###

*[Design memory]*    *Weight of hull skin equals volume of hull skin times density of hull skin*

*[Design memory]*    *Maximum of weight of hull skin equals maximum of volume of hull skin times density of hull skin*

*[Design memory]*    *Resolve concepts!*

*[Design memory]*    *Volume of hull skin equals surface area of hull skin times thickness of hull skin*
*[Design memory]*    *Maximum of volume of hull skin equals maximum of surface area of hull skin times thickness of hull skin*

*[Design memory]*    *Resolve concepts!*

### The value of 'thickness' is given earlier, but the system can not find the maximum of surface area: ###

*[Design memory]*    *Maximum of surface area is unknown*
*[Design memory]*    *Find maximum of surface area of hull skin!*
*[Design memory]*    *Locate method!*

### Here the system 'knows' that the maximum surface area is that which is enclosed in a box bounded in the three dimensions. Presumably, several other transactions are undertaken before this is deduced: ###

*[LT Storage]*    *Maximum of surface area equals L times B times D*

*[Design memory]*    *Resolve concepts!*

*[Design memory]*    *L is 120 m*
*[Design memory]*    *B is 20 m*

*[Design memory}*    *D is unknown*

| | |
|---|---|
| *[Design memory]* | *Find D!* |
| *[Design memory]* | *Locate method!* |
| | |
| *[LT Storage]* | *D equals T plus F* |
| | |
| *[Design memory]* | *Resolve concepts!* |
| | |
| *[Design memory]* | *T is* 7 |
| | |
| *[Design memory]* | *F is unknown* |
| *[Design memory]* | *Find F!* |
| *[Design memory]* | *Locate method!* |

### At this point the system has traversed through a series of relationships until it stopped upon finding no useful relationship: ###

| | |
|---|---|
| *[Design memory]* | *No method found* |

### The system resigns and resorts to the designer. Note that 'F' does not have to be defined further - it is of no importance to the system what 'F' actually signifies, since it is only used as a 'temporary' concept in this case: ###

| | |
|---|---|
| **[DPL System]** | **"What is F?"** |
| **[Operator]** | **"4 m"** |
| | |
| *[Design memory]* | *F is  4* |

### The system instantiates the relationship it trued to resolve: ###

| | |
|---|---|
| *[Design memory]* | *D is 7 m plus 4 m* |
| | |
| *[Design memory]* | *D is  11* |

### The system calculates some of the required information based on relationships found earlier: ###

| | |
|---|---|
| *[Design memory]* | *Maximum of surface area of hull skin equals 26400 $m^2$* |
| *[Design memory]* | *Maximum of volume of hull skin equals 422 $m^3$* |
| *[Design memory]* | *Maximum of weight of hull skin equals 3379 tons* |

*[Design memory]*    *Minimum of weight of hull skin is unknown*

*[Design memory]*    *Find minimum of weight of hull skin!*

### The system tries to resolve the second concept in the relationship describing the range of the weight: ###

*[Design memory]*    *Locate method!*

*[LT Storage]*    *Minimum weight of hull skin equals minimum surface area of hull skin times density of hull skin*

*[Design memory]*    *Resolve concepts!*

*[Design memory]*    *Minimum of surface area of hull skin is unknown*

*[Design memory]*    *Find minimum of surface area!*

*[Design memory]*    *Locate method!*

### The system 'knows' that the shape with least surface area with respect to volume is a sphere and uses this relation (in lack of better methods): ###

*[LT Storage]*    *Minimum of surface area equals surface area of sphere*

### Retracting from long term memory necessary relationships (formulas). ###

*[LT Storage]*    *Surface area of sphere equals 4 times $\pi$ times radius$^2$*

*[LT Storage]*    *Volume of sphere equals 1.33 times $\pi$ times radius$^3$*

### Deduced by some transactions and relationships not described here: ###

*[Design memory]*    *Volume of sphere equals volume of displaced water*

### This is found earlier in the process: ###

*[Design memory]*    *Volume of displaced water is 10416 m$^3$*

### Instantiating: ###

*[Design memory]*    *Volume of sphere equals 10416 m$^3$*

### Assume now that the system chooses a new strategy - it <u>plans</u> how to attain the necessary information: ###

*[Design memory]*        *Find plan!*

### The plan might be constructed based one experiences in the process of finding the maximum weight: ###

*[Design memory]*        *The plan is should find radius of sphere then should find surface area of sphere then should find volume of skin of sphere then should find weight of skin of sphere*

### The first step of the plan is then initialized. The system will know the sequence from interpreting the then-relationship describing the plan: ###

*[Design memory]*        *Should find radius of sphere*

### Satisfying the requirement established in the sentence above. A mobilization of an ability: ###

*[Design memory]*        *Find radius of sphere!*

### The system simply calculates relationships found earlier: ###

*[Design memory]*        *10416 equals 1.33 times 3.14 times radius$^3$*
*[Design memory]*        *10416 equals 4.18 times radius3 of sphere*
*[Design memory]*        *Radius3 of sphere equals 10416 divided by 4.18*
*[Design memory]*        *Radius of sphere equals 2494$^{\circ 333}$*
*[Design memory]*        *Radius of sphere equals 13.56*
*[Design memory]*        *Radius of sphere is 13.56*

### The second step in the plan is initialized: ###

*[Design memory]*        *Should find surface area of sphere*

### And mobilized: ###

*[Design memory]*        *Find surface area of sphere!*

*[Design memory]*        *Surface area of sphere equals 4 times 3.14 times 13.56$^2$*

| | |
|---|---|
| *[Design memory]* | *Surface area of sphere equals 2309* m$^2$ |
| *[Design memory]* | *Surface area of sphere is 2309* m$^2$ |

### The third step in the plan is initialized and mobilized: ###

| | |
|---|---|
| *[Design memory]* | *Should find volume of skin of sphere* |
| *[Design memory]* | *Find volume of skin of sphere!* |
| *[Design memory]* | *Volume of skin of sphere equals 2309* m$^2$ *times 0.016 m* |
| *[Design memory]* | *Volume of skin of sphere equals 37.0 m$^3$* |

### The fourth and last step in the plan is initialized and mobilized: ###

| | |
|---|---|
| *[Design memory]* | *Should find weight of skin of sphere* |
| *[Design memory]* | *Find weight of skin of sphere!* |
| *[Design memory]* | *Weight of skin of sphere equals 296 tons* |

### The plan is finished. Now remains resolution of the initial relationships: ###

| | |
|---|---|
| *[Design memory]* | *Minimum of weight of skin equals weight of skin of sphere* |

### Instantiating the relationship: ###

| | |
|---|---|
| *[Design memory]* | *Minimum of weight of skin equals 296 tons* |

### The system then instantiates and reports the 'range' relationship. The process is finished - all relevant concepts are sufficiently defined for this task! ###

| | |
|---|---|
| *[Design memory]* | *Range of weight of skin is larger than 296 tons and less than 3379 tons* |
| **[DPL System]** | **"Range of weight of skin is larger than 296 tons and less than 3379 tons"** |